\def\APResearchDesignOverride{1}
\documentclass[designerfonts=true]{allpurpose}

\usepackage{makecell}
\usepackage{multirow}
\usepackage{amsmath}
\usepackage{unicode-math}
\usepackage{amsthm}

\usepackage{microtype}
\usepackage{hyperref}
\usepackage{url}
\usepackage{booktabs}
\usepackage{graphicx}
\usepackage{pifont}
\usepackage{lineno}
\usepackage{amsmath}
\usepackage{booktabs}
\usepackage{pifont}

\usepackage{booktabs}
\usepackage{pifont}
\usepackage{xcolor}

\usepackage{tcolorbox}
\usepackage{wrapfig}
\tcbset{
    colback=gray!5,
    colframe=black,
    boxrule=0.5pt,
    arc=2pt,
    left=6pt,
    right=6pt,
    top=6pt,
    bottom=6pt
}

\definecolor{darkblue}{rgb}{0, 0, 0.5}
\hypersetup{colorlinks=true, citecolor=darkblue, linkcolor=darkblue, urlcolor=darkblue, hypertexnames=false}
\usepackage{algorithm}
\usepackage{algpseudocode}
\usepackage{enumitem}
\usepackage{pifont}

\usepackage{tikz}
\usetikzlibrary{arrows.meta,positioning,calc}
\usepackage{pgfplots}
\pgfplotsset{compat=1.18}

\title[SpectralCanary]{SpectralCanary: Embarrassingly Simple Low-Frequency Traces for T2I Dataset-Use Auditing}
\paperstatus{Preprint}
\date{ }
\paperlink[globe]{Project}{https://hosytuyen.github.io/projects/SpectralCanary}
\paperlink[code]{Code repository}{https://github.com/hosytuyen/SpectralCanary}
\paperemail{stho@umd.edu}
\author{
  Sy-Tuyen Ho\affmark{1} \quad
  Anirudh Satheesh\quad
  Soham Chawathe\affmark{1}\quad
  Huy Nghiem\affmark{1}\quad \\
  \apauthor{Yongkyun Lee}\quad
  \apauthor{Andrew Adams Schoen}\quad
  \apauthor[1]{Furong Huang}
}
\affiliations{\apaffiliation\qquad\affmark{1}University of Maryland}

\abstracttext{Auditing whether a text-to-image (T2I) model was fine-tuned on a particular image collection is difficult: a model can absorb collection-level visual characteristics without reproducing training images. Proactive dataset-use auditing plants a shared signal in a dataset before release, then later checks whether that signal appears in a suspect model's generated outputs. Existing approaches often trade off verifiability, fidelity, robustness, and deployability. We introduce \textbf{SpectralCanary}, which uses a smooth low-pass mask to blend a fixed canary image into every collection image while leaving captions unchanged and requiring neither auxiliary generation nor per-image optimization. A lightweight verifier detects the transferred signal in suspect-model outputs. Across eight tests on the Pokemon dataset and three realistic data transformations, spanning two architectures families, SpectralCanary achieves the highest average per-generation TPR@$1\%$ FPR ($91.1\%$) and fixed threshold model-level accuracy using 30 generations ($94.8\%$) among the evaluated methods. Across five datasets spanning artistic, medical, satellite, and fashion domains with SD3.5, SpectralCanary averages $92.4\%$ per-generation TPR@$1\%$ FPR while preserving the released images (SSIM $0.989$, LPIPS $0.022$) with a near-zero mean change FID (mean $\Delta$FID $0.05$) across datasets. Matched-energy ablations on Pokemon and ROCOv2 support our design choice, low-frequency placement transfers more reliably compared to mid/high-frequency placement. Under the evaluated adaptive attacks, the most effective removal requires aggressive modification and incurs more than $25$ $\Delta$FID points. These results identify low-frequency visual signals as effective carriers for collection-level T2I audit traces.}

\begin{document}

\maketitle

\section{Introduction}

Creative and scientific image collections embody substantial human labor, expertise, and investment. Adapting text-to-image (T2I) models to such collections can support legitimate personalization, education, design, and scientific communication~\citep{Rombach_2022_CVPR,esser2024scaling,hu2022lora}. The same capability, however, allows a third party to fine-tune and deploy a model on a collection without the rights holder's consent, attribution, or compensation. Because a model can absorb a collection's visual characteristics without reproducing individual images, rights holders may have little evidence that their work was used~\citep{pang2025black,11023473}. Practical dataset-use auditing can strengthen accountability and licensing compliance while allowing images to remain available for legitimate use.

\begin{figure*}[!t]
  \centering
  \includegraphics[width=\textwidth]{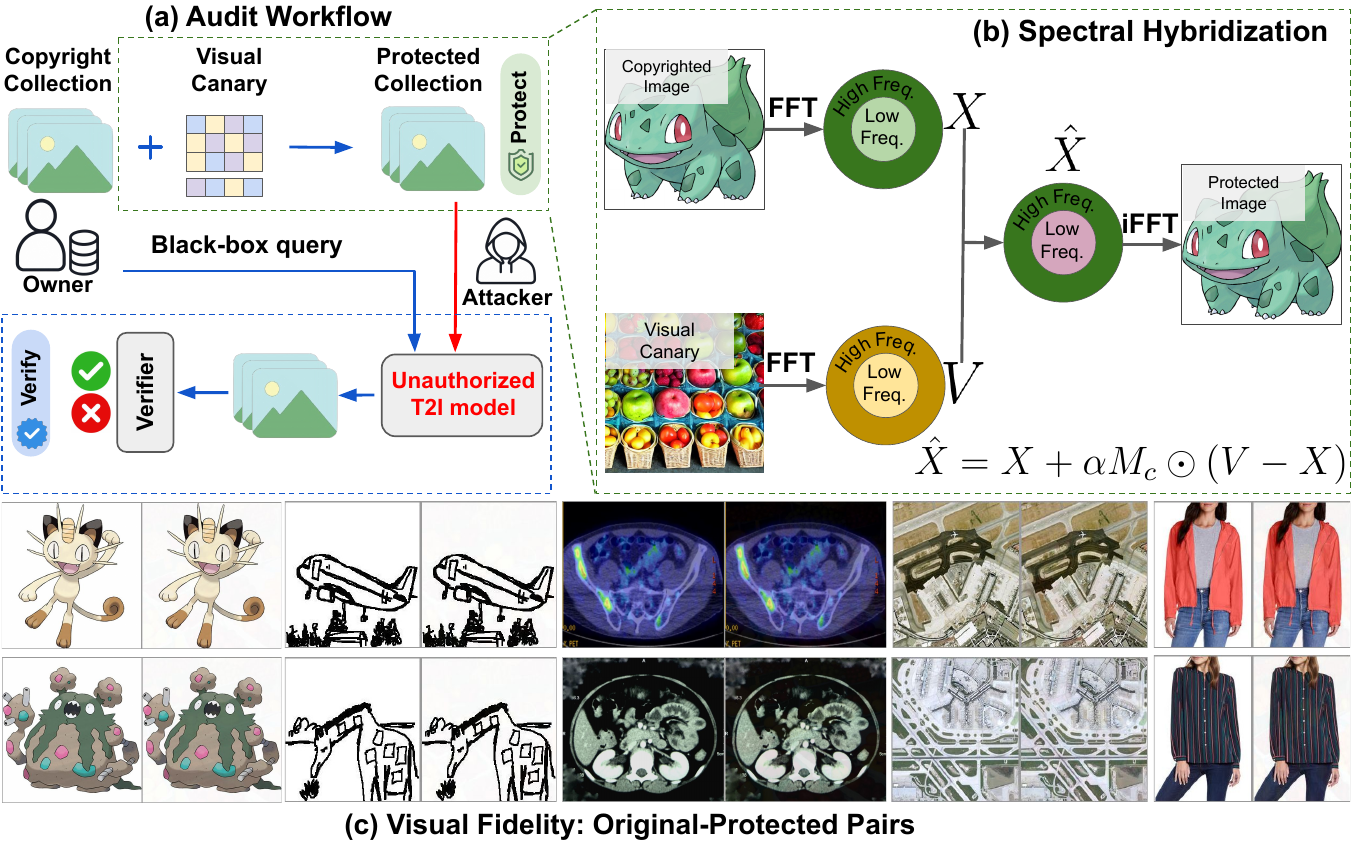}
  % \vspace{-0.7cm}
  \caption{\textbf{SpectralCanary overview.} \textbf{(a) Auditing workflow.} An owner protects a collection before release, then queries a suspect T2I model and tests whether its generated outputs contain the canary. \textbf{(b) Spectral hybridization.} SpectralCanary blends a fixed canary through a smooth low-pass mask $M_c$ with strength $\alpha$, retaining most of the remaining spectrum. A lightweight verifier trained on original--protected pairs detects the trace in model generations. \textbf{(c) Visual fidelity.} Original--protected pairs across five datasets, including sparse sketches.}
  % \vspace{-0.3cm}
  \label{fig:method-overview}
\end{figure*}

We study proactive dataset-use auditing, commonly termed data traceability or unauthorized data-use verification in prior works. A practical method should satisfy four requirements. \textbf{(Req. 1) Collection-level verifiability.} It should provide evidence from black-box outputs even when no training image is reproduced. \textbf{(Req. 2) Fidelity and utility preservation.} Protected samples should preserve their original appearance and the utility of models trained on them, avoiding harm to legitimate use while resisting easy filtering. \textbf{(Req. 3) Pipeline robustness.} The trace should remain detectable across model architectures, after augmentation such as image regeneration or recaptioning. \textbf{(Req. 4) Operational deployability.} The protection transform should be model-agnostic and scale efficiently to large collections without auxiliary generation or iterative per-image optimization. 
% A positive result constitutes statistical evidence under a stated threat model, not standalone proof of copyright infringement.

Prior approaches typically sacrifice at least one requirement. \textbf{Reactive auditing: limited collection-level evidence (Req.~1).} Membership and similarity tests rely on sample-specific evidence that generalization may erase~\citep{pang2025black,11023473}. \textbf{Pixel-level watermarks: limited robustness (Req.~3).} Weak signals may be attenuated by augmentation, regeneration, or fail to transfer across model backbones~\citep{wang2024diagnosis,11023473}. \textbf{Explicit canaries: fidelity and robustness trade-offs (Reqs.~2--3).} Repeated templates can be conspicuous and filterable, while recaptioning can remove textual triggers~\citep{ren2025entruth}. \textbf{Adjacent data-poisoning objectives.} Poisoning induces chosen behavior rather than provides evidence of dataset use. Persistent behavior may not be unique or reliably elicited, so it does not by itself establish use of a particular collection~\citep{pmlr-v235-wang24bm,jang2025silent}. Across proactive methods, auxiliary generation, filtering, and iterative optimization can also limit deployability \textbf{(Req. 4)} by increasing computational cost or requiring external models, domain-specific assumptions~\citep{ren2025entruth,11023473,jang2025silent}. These limitations motivate a subtle, transferable, caption-independent, and inexpensive trace.

We propose \textbf{SpectralCanary}, whose key design choice is the frequency band used for the shared trace. Before release, the dataset owner chooses one fixed canary image. For each collection image, SpectralCanary mixes a small amount of the canary's low-frequency Fourier content into the original while retaining most of its spectrum and leaving its caption unchanged (Fig.~\ref{fig:method-overview}(b)). The protected spectra are thereby shifted toward a common low-frequency target, creating a collection-level signal. Low frequencies encode broad spatial structure that neural networks preferentially learn~\citep{rahaman2019spectral,xu2019frequency} and that is important to T2I generation~\citep{yi2024towards,ma2025decofrequencydecoupledpixeldiffusion}. This motivates our hypothesis that a weak low-frequency trace can transfer during fine-tuning. The dataset owner then queries a suspect model and uses a lightweight verifier to test whether its generated outputs contain the trace (Fig.~\ref{fig:method-overview}(a)). Importantly, SpectralCanary uses a single deterministic transform requiring no caption changes, auxiliary generator, or per-image optimization.

\textbf{We evaluate all four requirements.} Robustness tests cover a convolutional latent-diffusion U-Net~\citep{Rombach_2022_CVPR}, a transformer-based rectified-flow model~\citep{esser2024scaling}; data augmentation, image regeneration~\citep{zhao2024invisible}, and recaptioning. Direct baselines include pixel-level traces (SIREN~\citep{11023473}, DIAGNOSIS~\citep{wang2024diagnosis}) and explicit canaries (EnTruth~\citep{ren2025entruth}). We also adapt Silent Branding \citep{jang2025silent} but report its evaluation separately in Appx.~\ref{appx:silent-branding}, because it targets visual branding rather than dataset-use auditing.

\textbf{Auditing effectiveness and robustness.} For per-generation discrimination, AUC measures verifier-score separation, while TPR at 1\% FPR measures detection of outputs from protected-data models. Across eight Pokemon settings, SpectralCanary achieves 98.0\% AUC and 91.1\% TPR. In fixed-threshold model-level verification using 30 generations, it achieves 94.8\% average accuracy under the same settings. \textbf{Fidelity and utility preservation.} SSIM and LPIPS compare original and protected images, while $\Delta$FID measures the change in generative quality. Across five visual domains with SD3.5, SpectralCanary achieves $99.5\%$ AUC and $92.4\%$ TPR with SSIM $0.989$, LPIPS $0.022$, with a near-zero mean change FID ($\Delta$FID $0.05$). On sparse Sketch-Scene drawings, it achieves $100\%$ AUC and TPR with SSIM $0.997$ and LPIPS $0.002$. \textbf{Operational deployability.} Protection takes $0.39$ seconds per image on our CPU setup (Appx.~\ref{appx:coating-efficiency}). \textbf{Mechanism and boundary.} In matched-energy ablations on Pokemon and ROCOv2, we show that low-frequency placement transfers more reliably compared to mid/high-frequency placement. Within the evaluated adaptive attacks, effective removal requires aggressive modification and incurs more than $25$ $\Delta$FID points.

\textbf{Our main contributions are:}
\begin{itemize}[leftmargin=1.1em,labelsep=0.3em,itemsep=0.15em,topsep=0.25em,parsep=0pt,partopsep=0pt]
    \item We identify low-frequency visual structure as an effective carrier for weak collection-level traces during T2I fine-tuning. In matched band size and perturbation energy ablations on Pokemon and ROCOv2 with SD1.5, we show that low-frequency placement yields higher per-generation TPR than mid/high-frequency placement.

    \item We introduce SpectralCanary, a deterministic and caption-preserving spectral hybridization method. It protects each image with one inexpensive transform, requiring neither an auxiliary generator nor per-image optimization.

    \item We jointly evaluate per-generation discrimination, KS-based multi-query auditing, fixed-threshold model-level verification, image fidelity, generative utility, robustness, and protection cost across two architecture families, five datasets, three direct auditing baselines, an adapted poisoning baseline, and three realistic data transformations. This evaluation establishes a strong empirical operating point and characterizes the boundary under adaptive spectral replacement.
\end{itemize}

\section{SpectralCanary}\label{sec:spectralcanary}

\textbf{Problem setup.} Let $\mathcal{D}=\{(x_i,t_i)\}_{i=1}^N$ be the defender's image--caption collection, with $x_i\in[0,1]^{H\times W\times3}$. Given a canary $v^\star$, protected fraction $\delta\in(0,1]$, blending strength $\alpha\in[0,1]$, and Gaussian bandwidth $c>0$, SpectralCanary constructs protected collection $\widehat{\mathcal{D}}$ and trains verifier $f_\theta$ (Alg.~\ref{alg:coating}). A third party may fine-tune a T2I model $\mathcal{G}$ on $\widehat{\mathcal{D}}$. The defender audits its black-box generations against clean reference scores (Alg.~\ref{alg:verification}). 

\textbf{Threat model.} The algorithm and released collection are public. $\delta$, $\alpha$, and $c$ may be known. The defender retains the originals, $v^\star$, and $f_\theta$ and has only black-box access to $\mathcal{G}$. The audit provides calibrated statistical evidence, not cryptographic attribution or white-box robustness. Revealing the canary spectrum enables the idealized inversion below.

\textbf{Design motivation.}
Low frequencies encode broad structure that neural networks preferentially learn~\citep{rahaman2019spectral,xu2019frequency} and that is important to T2I generation~\citep{yi2024towards,ma2025decofrequencydecoupledpixeldiffusion}. They are therefore a plausible trace carrier. However, strong modification is visible. SpectralCanary uses a smooth low-pass mask to weight frequencies and a small blending strength to control the overall change. Our controlled study in Sec.~\ref{sec:frequency-placement-summary} shows low-frequency placement yields higher detection rates than mid/high-frequency placement.

\textbf{Fourier convention.}
Throughout, $\mathcal{F}_c$ denotes the centered channel-wise 2D discrete Fourier transform and $\mathcal{F}_c^{-1}$ denotes its inverse:
\begin{equation}
    \mathcal{F}_c
    = \operatorname{fftshift}\circ\operatorname{DFT}_2,
    \qquad
    \mathcal{F}_c^{-1}
    = \operatorname{IDFT}_2\circ\operatorname{ifftshift}.
    \label{eq:centered-fourier-convention}
\end{equation}
We use unitary normalization and apply the transform channel-wise. Any matched forward--inverse scaling gives the same reconstruction because Eq.~\ref{eq:coated-spectrum} is homogeneous in $X_i$ and $V$. Real inputs and the real, conjugate-symmetric mask make the inverse real up to roundoff. Let $\Pi_8$ denote clipping to $[0,1]$ followed by 8-bit quantization.

\begin{figure*}[t]
  \centering
  \begin{minipage}[t]{0.485\textwidth}
    \footnotesize
    \hrule height 0.8pt
    \vspace{1pt}
    \refstepcounter{algorithm}\label{alg:coating}
    \textbf{Algorithm \thealgorithm: Protection and Verifier Training}
    \vspace{1pt}
    \hrule
    % \vspace{1pt}
    \begin{algorithmic}[1]
      \Require $\mathcal{D},v^\star,\delta,c,\alpha$
      \Ensure $\widehat{\mathcal{D}},f_\theta$
      \State Construct $M_c$ using Eq.~\ref{eq:frequency-mask}
      \State $\mathcal{I}_\delta\gets\operatorname{UniformSubset}([N],\lfloor\delta N\rfloor)$
      \State $V \gets \mathcal{F}_c(v^\star)$; $\widehat{\mathcal{D}} \gets \mathcal{D}$
      \ForAll{$i \in \mathcal{I}_\delta$}
        \State $X_i \gets \mathcal{F}_c(x_i)$
        \State $\widehat{X}_i \gets X_i+\alpha M_c\odot(V-X_i)$
        \State $\tilde{x}_i \gets \Pi_8\!\left(\Re[\mathcal{F}_c^{-1}(\widehat{X}_i)]\right)$
        \State $\widehat{\mathcal{D}}[i]\gets(\tilde{x}_i,t_i)$
      \EndFor
      \State Train $f_\theta$ on $\mathcal{I}_\delta$ using Eq.~\ref{eq:verifier-loss}
      \State \Return $\widehat{\mathcal{D}}, f_\theta$
    \end{algorithmic}
    % \vspace{1pt}
    \hrule height 0.8pt
  \end{minipage}
  \hfill
  \begin{minipage}[t]{0.485\textwidth}
    \footnotesize
    \vspace{6pt}
    \hrule height 0.8pt
    \vspace{2pt}
    \refstepcounter{algorithm}\label{alg:verification}
    \textbf{Algorithm \thealgorithm: Auditing a Suspect Model}
    \vspace{2pt}
    \hrule
    \vspace{2pt}
    \begin{algorithmic}[1]
      \Require $\mathcal{G}$, $\mathcal{P}$, $f_\theta$, clean-data reference scores $\mathcal{R}$, level $\gamma$
      \Ensure $\mathcal{S}^{\mathcal{G}}$, $p$-value $q^{\mathcal{G}}$, decision $a^{\mathcal{G}}$
      \State $\mathcal{S}^{\mathcal{G}} \gets \emptyset$
      \ForAll{$p_j \in \mathcal{P}$}
        \State Generate $x_j^{\mathcal{G}} \sim \mathcal{G}(p_j)$
        \State $s_j \gets f_\theta(x_j^{\mathcal{G}})$
        \State Add $s_j$ to $\mathcal{S}^{\mathcal{G}}$
      \EndFor
      \State $q^{\mathcal{G}} \gets \operatorname{KS}_{+}(\mathcal{S}^{\mathcal{G}},\mathcal{R})$
      \State $a^{\mathcal{G}} \gets \mathbb{1}[q^{\mathcal{G}}<\gamma]$
      \State \Return $\mathcal{S}^{\mathcal{G}},q^{\mathcal{G}},a^{\mathcal{G}}$
    \end{algorithmic}
    % \vspace{2pt}
    \hrule height 0.8pt
  \end{minipage}
\end{figure*}

\textbf{Stage I: Constructing protected images.}
  Alg.~\ref{alg:coating} uniformly selects $|\mathcal{I}_\delta|=\lfloor\delta N\rfloor$ images; only these are replaced and used to train the verifier ($\delta=1$ by default). For $i\in\mathcal{I}_\delta$, we resize $v^\star$ as specified in Appx.~\ref{appx:hyper-parameters} and compute $X_i=\mathcal{F}_c(x_i)$ and $V=\mathcal{F}_c(v^\star)$. We define a smooth radial low-pass mask:
  \begin{equation}
      M_c(k,\ell)
      = \exp\!\left(-\frac{r(k,\ell)^2}{2c^2}\right),
      \label{eq:frequency-mask}
  \end{equation}
  where $(k,\ell)$ indexes the shifted Fourier grid, whose center is
  $(k_0,\ell_0) = (\lfloor H/2 \rfloor, \lfloor W/2 \rfloor)$, and
  \begin{equation}
      r(k,\ell)
      =
      \sqrt{
      \left(\frac{k-k_0}{H}\right)^2
      +
      \left(\frac{\ell-\ell_0}{W}\right)^2 }.
      \label{eq:normalized-radius}
  \end{equation}
  Normalizing the two axes separately also accommodates rectangular images. The mask equals one at the spectral center and decays smoothly with radial distance; a smaller bandwidth $c$ concentrates the weighting more tightly around the lowest frequencies. The protected spectrum is
  \begin{equation}
    % \begin{aligned}
      \widehat{X}_i
      = X_i + \alpha M_c \odot (V-X_i) \\
      = (1-\alpha M_c)\odot X_i + \alpha M_c\odot V,
    % \end{aligned}
      \label{eq:coated-spectrum}
  \end{equation}
where $\alpha\in[0,1]$ is the blending strength and $\odot$ denotes elementwise multiplication. We export $\tilde{x}_i=\Pi_8(\Re[\mathcal{F}_c^{-1}(\widehat{X}_i)])$. Eq.~\ref{eq:coated-spectrum} moves each weighted coefficient toward target $V$. It combines shared term $\alpha M_c\odot V$ with image-specific term $(1-\alpha M_c)\odot X_i$. The collection mean shifts toward $V$, while variation contracts in low-pass-weighted frequencies. Appx.~\ref{appx:Verification-Learning-Dynamics} reports learning dynamics. Fig.~\ref{fig:method-overview}(c) and Appx.~\ref{appx:extensive-visualizations} show visual fidelity. The deterministic construction requires neither an auxiliary generative model nor iterative per-image optimization. Appx.~\ref{appx:coating-efficiency} reports runtime.

\textbf{Training the verifier.}
Let $f_\theta(x)\in(0,1)$ denote the predicted probability that image $x$ contains the protection signal. Using selected protected images as positives and their original counterparts as negatives, we minimize
\begin{equation}
    \mathcal{L}_{\mathrm{ver}}(\theta)
    = -\frac{1}{|\mathcal{I}_\delta|}\sum_{i\in\mathcal{I}_\delta}
    \left[
      \log f_\theta\!\left(\tilde{x}_i\right)
      + \log\!\left(1-f_\theta\!\left(x_i\right)\right)
    \right].
    \label{eq:verifier-loss}
\end{equation}
Training uses no model generations. Each original--protected pair shares semantic content, making the protection transform the only systematic label-aligned difference. This encourages but does not guarantee trace learning. Transfer to unseen generations is evaluated in Sec.~\ref{sec:experiment-results}.

\textbf{Stage II: Auditing a suspect model.} Let $\mathcal{R}$ contain clean-data reference scores and $\mathcal{P}$ be trigger-free audit prompts (collection captions in our experiments). For suspect model $\mathcal{G}$, we collect $\mathcal{S}^{\mathcal{G}}$ and apply a one-sided two-sample Kolmogorov--Smirnov test for larger suspect scores, rejecting at level $\gamma$. Experiments use matched clean-data models. Deployment calibrates $\mathcal{R}$ on representative independent models. This yields the KS-based multi-query audit decision. AUC and TPR are per-generation discrimination metrics (Appx.~\ref{appx:traceability-metrics}).

\textbf{Invertibility.} If an attacker knows the canary spectrum $V$, blending strength $\alpha$, and Gaussian bandwidth $c$, then $M_c$ is known and Eq.~\ref{eq:coated-spectrum} can be inverted as
\begin{equation}
X_i
=
\frac{\widehat{X}_i-\alpha M_c\odot V}
{1-\alpha M_c},
\label{eq:inverse-coating}
\end{equation}
This recovers pre-export $X_i$ exactly when $\alpha<1$, but clipping and quantization prevent exact recovery from the released image. Our threat model may reveal $\alpha$ and $c$ but keeps $V$ private. These are threat-model assumptions, not security guarantees.

\section{Related Work}\label{sec:related-work}

\textbf{Reactive and proactive auditing.} Reactive methods test an already trained model using membership inference or similarity between generated and reference images~\citep{chen2026membership,pang2025black,radford2021learning,oquab2024dinov}. Such evidence can weaken when fine-tuning transfers collection-level characteristics without retaining identifiable information about individual samples. Proactive methods instead modify a collection before release. Anti-learning methods aim to prevent or redirect customization~\citep{291164,zhao2024unlearnable,van2023anti,ahn2025nearly,Liu_2024_CVPR}. Proactive traceability aims to preserve customization utility while making later use statistically auditable.

\textbf{Proactive traceability.} Pixel-level methods such as DIAGNOSIS \citep{wang2024diagnosis} and SIREN \citep{11023473} insert weak transformations or learned signals that can be detected in generated images. Their subtle signals preserve appearance, but can be attenuated by augmentation, regeneration, or model backbone changes. EnTruth instead inserts explicit canary samples containing a repeated template and textual trigger~\citep{ren2025entruth}. This coherent signal can transfer strongly, but the injected samples are conspicuous and caption triggers can be removed. Our direct comparisons span the two main proactive families including subtle pixel-level traces and explicit visual canary samples.

\textbf{Relation to poisoning.} Data poisoning induces chosen behaviors such as concept corruption, reproduction, or persistent branding~\citep{pmlr-v235-wang24bm,jang2025silent}. Persistence alone is insufficient for auditing. Auditing instead requires a collection-specific signal detectable under ordinary, trigger-free queries with controlled false positives. SpectralCanary targets this auditing objective rather than model corruption. Appx.~\ref{appx:related_work} provides further discussion and a method taxonomy.
\section{Experiments}

\subsection{Experimental Setup and How to Read the Results}\label{sec:experimental-setups}

Our experiments ask four concrete questions. First, can SpectralCanary reveal that a model was trained on a protected collection, even when the model architecture or training data are changed? Second, does adding the trace visibly change the released images or make a model trained on them produce worse images? Third, is protection fast enough for a large collection? Finally, why do low frequencies matter, and what happens when an informed attacker tries to remove the trace?

\textbf{Datasets and models.} We cover distinct image domains: charater art (Pokemon-BLIP-Captions~\citep{pinkney2022pokemon}), sparse scene drawings (Sketch-Scene~\citep{fscoco}), medical (ROCOv2~\citep{ruckert2024rocov2}), satellite (RSICD~\citep{lu2017exploring}), and fashion (FACAD~\citep{yang2020fashion}). We fine-tune Stable Diffusion v1.5 (SD1.5)~\citep{Rombach_2022_CVPR} and Stable Diffusion 3.5 (SD3.5)~\citep{esser2024scaling}. The main experiments use LoRA, a common parameter-efficient form of T2I adaptation. Appx.~\ref{appx:full-finetuning} repeats the key test with full-parameter fine-tuning.

\textbf{Comparisons and stress tests.} We compare with SIREN~\citep{11023473} and DIAGNOSIS~\citep{wang2024diagnosis}, which modify training images, and EnTruth~\citep{ren2025entruth}, which adds explicit canary samples. Appx.~\ref{appx:silent-branding} separately adapts Silent Branding~\citep{jang2025silent}, whose original goal is visual branding rather than dataset-use auditing. On Pokemon, we test each method under ordinary training and after three realistic changes to the training set: image augmentation, diffusion-based recreation of each training image (image regeneration)~\citep{zhao2024invisible}, and automatic recaptioning with Qwen2.5-VL 7B~\citep{Qwen2.5-VL}.

\textbf{How to read the metrics.} We distinguish the embedded \emph{trace} from two evaluation levels. \emph{Per-generation discrimination} scores each generated image independently and asks whether outputs from protected-data models receive higher verifier scores than outputs from clean-data models. AUC summarizes this score separation over all thresholds. TPR at $1\%$ FPR is the fraction of outputs from protected-data models detected at a threshold that falsely flags at most $1\%$ of outputs from clean-data models; thus, a TPR of $90\%$ means detecting about 90 of 100 protected-data-model outputs while flagging about 1 of 100 clean-data-model outputs. At the model level, we evaluate two decision protocols. \emph{KS-based multi-query auditing} combines scores from 30 generated images in a statistical test and reports the rejection rate on protected-data models. \emph{Fixed-threshold model-level verification} aggregates 30 generations from each held-out model and applies a threshold fixed after independent calibration, reporting TPR, FPR, precision, F1, and accuracy (Appx.~\ref{appx:model-level-verification}). For image fidelity, lower LPIPS~\citep{zhang2018perceptual} means that a protected image is closer to its original. For generated-image quality, $\Delta\mathrm{FID}=\mathrm{FID}_{\mathrm{protected}}-\mathrm{FID}_{\mathrm{clean}}$~\citep{Seitzer2020FID}; values near or below zero indicate no measured quality loss, whereas large positive values indicate degradation. Unless stated otherwise, SpectralCanary uses $\alpha=0.02$, $c=0.20$, and $\delta=1$. Every model-training and image-sampling setting is repeated with three independent runs. Detailed protocols are in Appx.~\ref{appx:hyper-parameters}.

\subsection{Per-Generation Discrimination Survives Changes to Models and Training Data}\label{sec:experiment-results}

\begin{figure}[t]
  \centering
  \includegraphics[width=\linewidth]{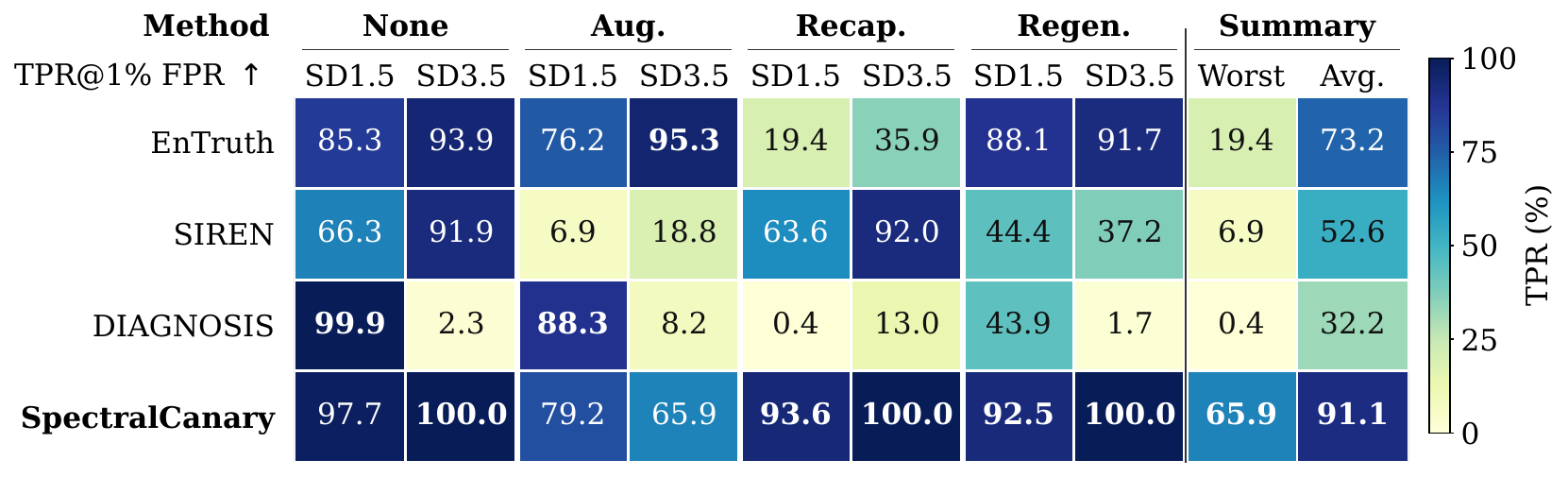}
  % \vspace{-0.3cm}
  \caption{\textbf{SpectralCanary maintains strong per-generation discrimination across model and data changes.} Each cell reports per-generation TPR at $1\%$ FPR. ``Worst'' is the lowest of a method's eight results, and ``Avg.'' is their mean.}
  \label{fig:pokemon_robust_summary}
\end{figure}

\textbf{SpectralCanary has the strongest average per-generation discrimination across the eight Pokemon settings.} Fig.~\ref{fig:pokemon_robust_summary} reports per-generation TPR for both model families under ordinary training, training-time augmentation, prompt recaptioning, and image regeneration. SpectralCanary averages $91.1\%$ TPR at $1\%$ FPR, compared with $73.2\%$ for the next-best method. Its weakest individual result is $65.9\%$ under SD3.5 with training-time augmentation; the best competing method has a worst-case result of only $19.4\%$. No method wins every individual setting: DIAGNOSIS is strongest for ordinary SD1.5 training, and EnTruth is strongest for SD3.5 with training-time augmentation. The complete per-generation AUC, TPR at $5\%$ FPR, and run-level results are in Appx.~\ref{appx:full-pokemon-results}. Appx.~\ref{appx:query-budget} separately reports KS-based multi-query auditing across query budgets, showing that more queries generally provide stronger evidence of training on protected data.

\begin{figure}[ht!]
  \centering
  \includegraphics[width=\linewidth]{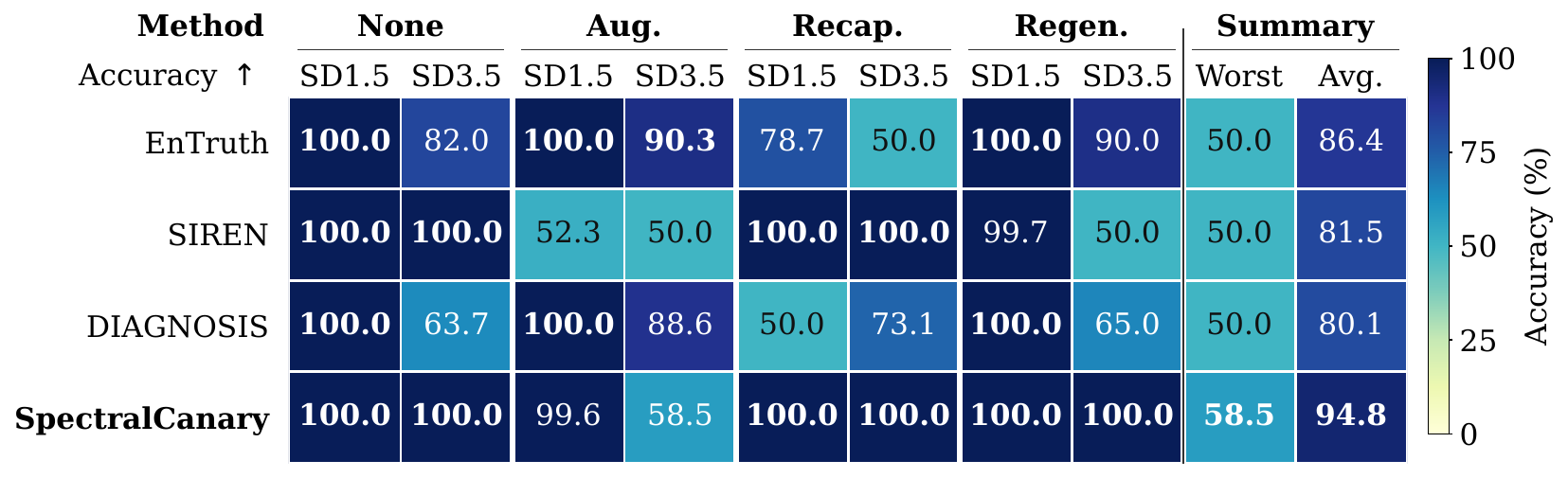}
  % \vspace{-0.3cm}
  \caption{\textbf{SpectralCanary is more consistent across transformations and models than the baselines.} Each cell is Acc. ``Worst'' is the lowest of a method's eight results, and ``Avg.'' is their mean.}
  \label{fig:pokemon_robust_model_level_summary}
\end{figure}

\textbf{Per-generation discrimination translates well to fixed-threshold model-level verification.} We aggregate 30 generations per held-out model, following~\citet{11023473}, and apply a threshold independently calibrated for each method--architecture pair and fixed across transformations. Fig.~\ref{fig:pokemon_robust_model_level_summary} reports model-level accuracy across the same eight settings. Each method--architecture--transformation setting includes five clean-data and five protected-data models, yielding 320 held-out models across the four methods. SpectralCanary achieves $94.8\%$ average model-level accuracy, showing that its per-generation discrimination signal supports dataset-use decisions at the model level. Appx.~\ref{appx:model-level-verification} provides the detailed protocol and TPR, FPR, precision, and F1 results.

\subsection{The Same Setting Works Across Five Image Types}

\begin{figure}[ht]
  \centering
  % \vspace{-0.3cm}
  \includegraphics[width=\columnwidth]{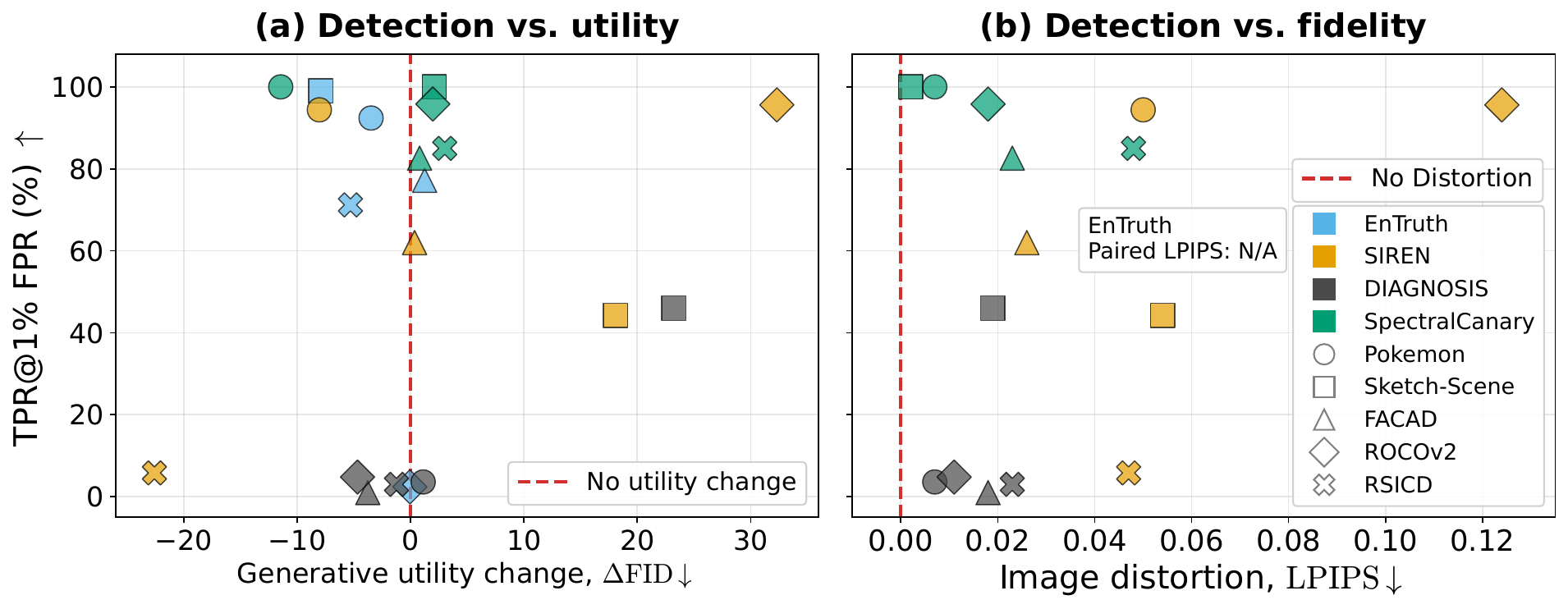}
  % \vspace{-0.4cm}
  \caption{\textbf{SpectralCanary works consistently across five image types.} Each point shows one method on one dataset. The vertical axis is the per-generation TPR at $1\%$ FPR: the percentage of outputs from protected-data models detected while $1\%$ of outputs from clean-data models are falsely flagged. In (a), the horizontal axis measures the change in the trained model's output quality; zero means no change. In (b), it measures the change from an original image to its protected copy; zero means identical images. The best results are near the upper-left corner.}
  % \vspace{-0.1cm}
  \label{fig:cross_dataset_results}
\end{figure}

We next test whether the same SpectralCanary setting works beyond Pokemon by asking three questions: Does a single generated output support reliable per-generation discrimination? Do the protected images still resemble the originals? Does training on those images degrade model outputs?

\textbf{SpectralCanary performs well on all three questions.} At a per-generation FPR of $1\%$, it achieves an average per-generation TPR of $92.4\%$ across outputs from models trained on protected data. The protected images remain close to their originals: average LPIPS is $0.022$, where zero means identical images. Training on the protected images causes almost no average change in model output quality: mean $\Delta$FID is $0.05$, where zero means no change. On the sparse Sketch-Scene drawings, SpectralCanary achieves $100\%$ per-generation TPR with LPIPS $0.002$. The hardest dataset is RSICD, where it achieves $78.9\%$ per-generation TPR with LPIPS $0.048$.

Fig.~\ref{fig:cross_dataset_results} compares all methods. Each point shows one method on one dataset. Good results appear near the upper-left corner: high per-generation TPR with little change to either the training images or the trained model's outputs. SpectralCanary's green points remain near this region across all five datasets. DIAGNOSIS changes the original images little but usually provides weak per-generation discrimination. SIREN works well on some datasets but poorly on others and makes the largest image changes. EnTruth works well on several datasets but fails on ROCOv2. It has no LPIPS value because it adds new canary images rather than making a protected copy of each original. Fig.~\ref{fig:method-overview}(c) and Appx.~\ref{appx:extensive-visualizations} show protected images beside their originals; Appx.~\ref{appx:full-cross-domain-results} reports all numerical results.

\subsection{Protection Is Fast as Well as Effective}

\begin{figure*}[ht]
  \centering
  \includegraphics[width=\textwidth]{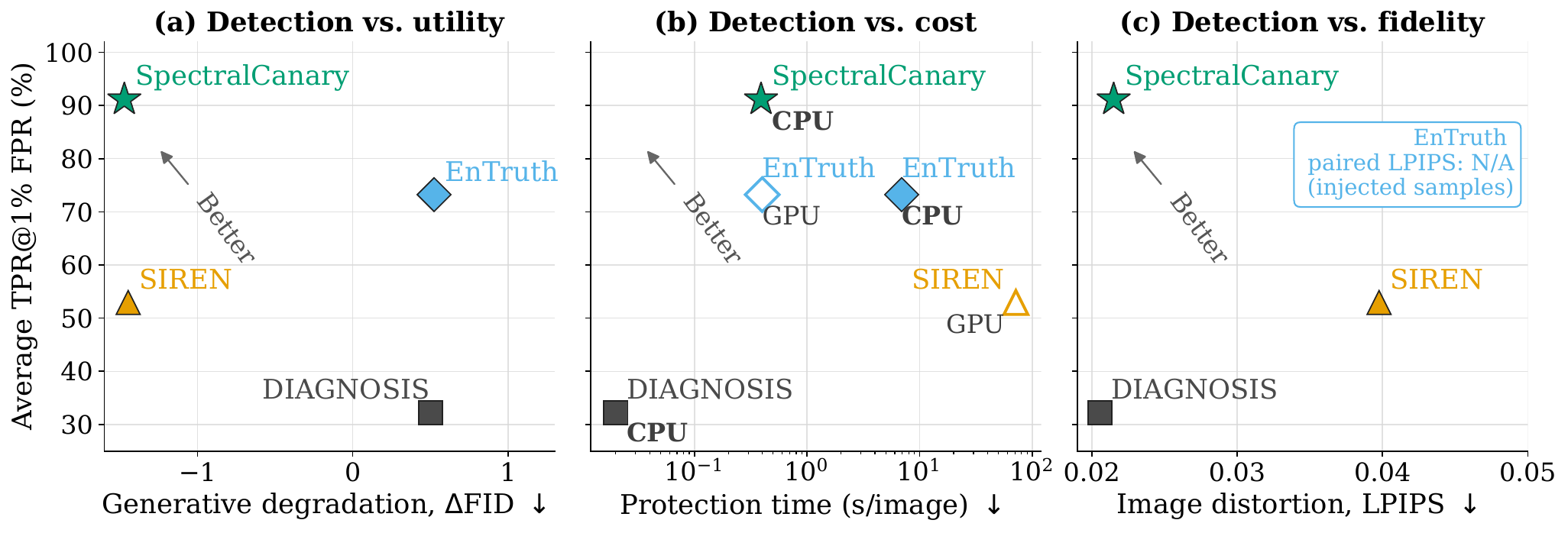}
  % \vspace{-0.2cm}
  \caption{\textbf{SpectralCanary achieves high per-generation discrimination with low quality and runtime cost.} Average per-generation TPR at $1\%$ FPR over the eight Pokemon settings is compared with (a) change in FID, (b) protection time, and (c) paired-image LPIPS. Preferred results lie upward and to the left. EnTruth adds $0.5\%$ auxiliary samples, so paired-image LPIPS is not defined for it.}
  % \vspace{-0.3cm}
  \label{fig:pokemon-pareto-summary}
\end{figure*}

\textbf{SpectralCanary protects an image in $0.39$ seconds on CPU while retaining the highest average per-generation TPR.} Fig.~\ref{fig:pokemon-pareto-summary} compares per-generation discrimination with generated-image quality, protection time, and image distortion on Pokemon. DIAGNOSIS is faster, but its average TPR is only $32.2\%$. SIREN and EnTruth require substantially more computation while achieving lower per-generation TPR. SpectralCanary also has no average FID penalty and the lowest paired-image LPIPS among the methods that modify every image. These timings are hardware-dependent, and CPU and GPU measurements are not perfectly interchangeable; their main purpose is to distinguish a single deterministic FFT-based transform from methods that require generating auxiliary images or performing iterative optimization. Appx.~\ref{appx:coating-efficiency} gives the complete timing results.

\subsection{Low Frequencies Carry the Trace More Reliably}\label{sec:frequency-placement-summary}

\begin{figure}[ht]
  \centering
  % \vspace{-0.1cm}
  \includegraphics[width=\columnwidth]{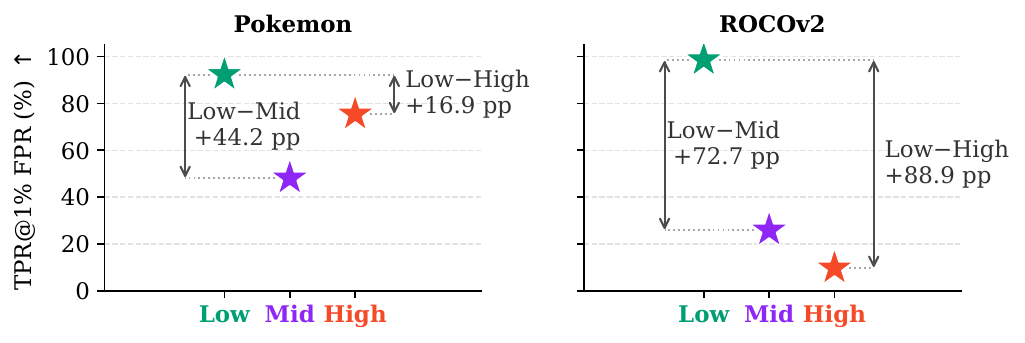}
  % \vspace{-0.4cm}
  \caption{\textbf{Matched low-frequency traces transfer more reliably than mid- or high-frequency traces.} Points show mean per-generation TPR at $1\%$ FPR over three canary images. Arrows show the performance gaps.}
  \label{fig:frequency-location-summary}
\end{figure}

\textbf{The advantage of low frequencies remains after controlling for the amount of spectral change.} We compare disjoint low-, mid-, and high-frequency bands on Pokemon and ROCOv2 with SD1.5. The bands contain the same number of Fourier coefficients, and we give them the same total amount of Fourier-domain change as the deployed smooth low-pass mask. This comparison prevents one band from winning merely because it changes more of the image. Fig.~\ref{fig:frequency-location-summary} shows that the low-frequency band has the highest per-generation TPR on both datasets. It exceeds the high-frequency band by $16.9$ percentage points on Pokemon and $88.9$ points on ROCOv2; the mid-frequency band is also weaker in both cases. Appx.~\ref{sec:spectral_location} describes the construction, and Tab.~\ref{tab:frequency_location_matched_energy} reports per-generation discrimination, image fidelity, generated-image quality, and variation across three canary images.

\subsection{Removal Attacks Damage Quality Before Erasing the Trace}\label{sec:adaptive-attack}

\begin{figure}[ht]
  \centering
  \includegraphics[width=\textwidth]{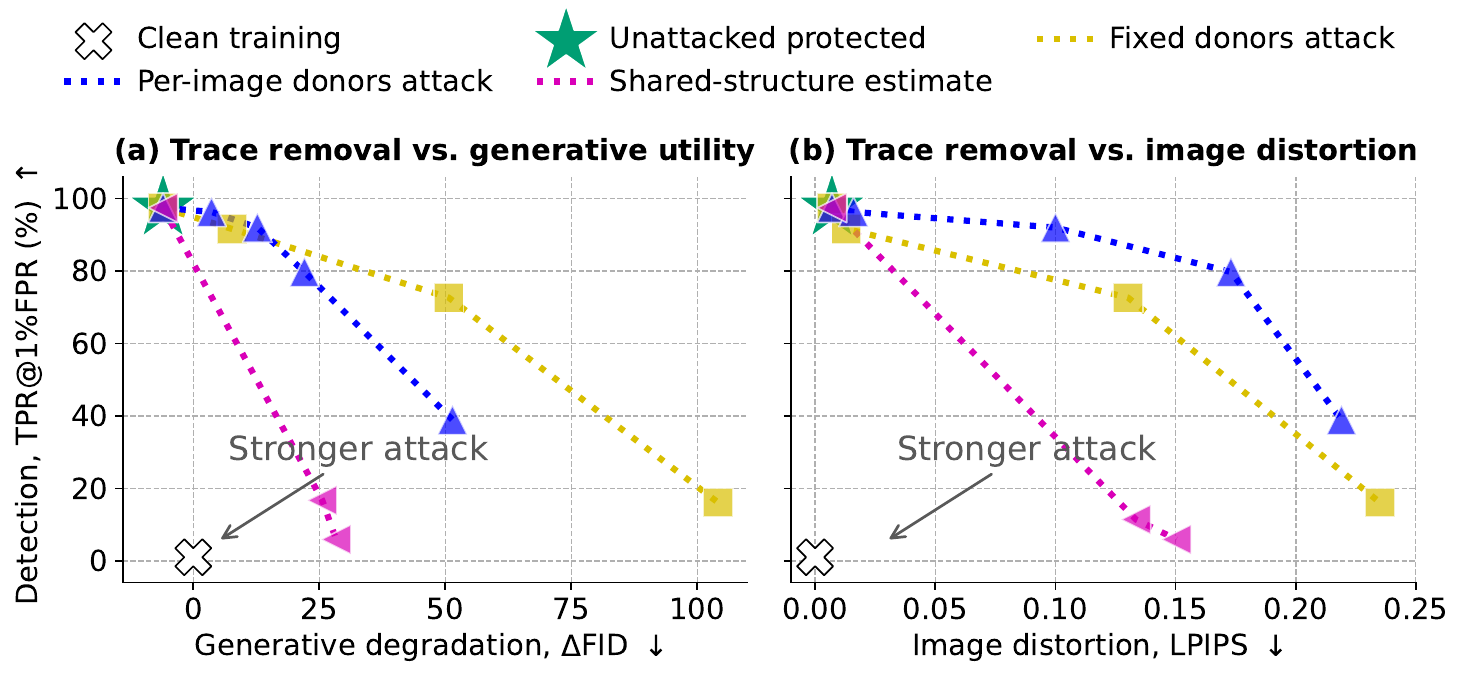}
  % \vspace{-0.3cm}
  \caption{\textbf{The evaluated attacks erase the trace only after causing substantial damage.} From the unattacked point, stronger attacks move down and to the right, indicating per-generation TPR falls, but $\Delta$FID and LPIPS increase. The clean-training marker shows the target TPR.}
  \label{fig:adaptive_sanitization}
\end{figure}

\textbf{In our evaluated attacks, substantially reducing per-generation TPR also substantially damages the training data or the resulting model.} The attacker knows the protection algorithm and may know $\alpha$ and $c$, but not the canary, original images, or verifier. We test three attacks. A \emph{donor} is an unrelated, unprotected image whose low-frequency content is used as a replacement source. Fixed-donor replacement uses the same donor for all protected images. Per-image replacement uses a different donor for each image. The third attack estimates the shared low-frequency pattern directly from the protected collection and removes it. Appx.~\ref{appx:adaptive-sanitization-settings} gives the equations and settings.

Fig.~\ref{fig:adaptive_sanitization} follows each attack from mild to aggressive removal. Mild attacks leave per-generation TPR largely intact. Driving TPR toward the clean-training baseline requires more than 25 additional FID points under the strongest evaluated attack and also produces large LPIPS changes. This establishes a removal--quality boundary for these three attacks; it is not a guarantee against every possible attack, especially one that can inspect the model or verifier.

% \FloatBarrier

\section{Limitations and Responsible Use}

Our evaluation covers two T2I architecture families, five datasets, and both LoRA and full-parameter fine-tuning. It does not establish transfer to every model, or mixture of a protected collection with a larger unprotected corpus. The main experiments also use one fixed natural canary image. Although the phase-randomized controls in Appx.~\ref{appx:spatial-structure-ablation} isolate the role of spatial coherence, broader evaluation across independently chosen canary images and key reuse is needed.

The adaptive evaluation assumes that the canary image and verifier remain private. If the attacker learns $V$, $\alpha$, and $c$, Eq.~\ref{eq:inverse-coating} permits exact inversion before clipping or quantization. The three evaluated attacks (fixed-donor replacement, per-image donor replacement, and shared-structure estimation) characterize only this secrecy-dependent threat model and do not provide a cryptographic or white-box robustness guarantee. Future evaluations should consider collusion across releases, learned removal models, stronger shared-structure estimators, and adaptive access to verifier feedback.

SSIM, LPIPS, and FID capture only part of image fidelity and generative utility. Human perceptual studies and task-specific utility measurements would be needed to support stronger claims about invisibility or harmlessness. Likewise, our KS-test rejection rates on protected-data models estimate statistical power under the stated sampling protocol, while fixed-threshold model-level verification uses separate clean-data and protected-data models for calibration. A deployment would require broader calibration across representative clean-data models and should treat a positive result as statistical evidence under the threat model, not as standalone proof of unauthorized use.

SpectralCanary is intended to help data owners audit model customization, but proactive traces can also be inserted into data without the knowledge of downstream users. Responsible use requires authorization to modify the collection, disclosure where appropriate, respect for dataset licenses, and safeguards against unsupported attribution. Decisions with legal or reputational consequences should combine the audit with independent evidence.

\section{Conclusion}

We propose SpectralCanary, a deterministic and inexpensive approach to proactive T2I dataset-use auditing. SpectralCanary injects a weak shared visual trace through a smooth low-pass mask while leaving captions unchanged. Across eight Pokemon model--transformation settings, it provides the strongest average per-generation discrimination and reliable KS-based multi-query auditing among the evaluated methods. The same operating point generalizes across five visual domains while preserving protected-image fidelity and generative utility. Matched-energy frequency-location experiments further show that low-frequency placement transfers more consistently than mid/high-frequency placement across datasets. Finally, the adaptive evaluation identifies a scoped robustness boundary, showing that the evaluated sanitization attacks substantially reduce per-generation TPR only at the cost of considerable image-quality or generative-utility degradation. Together, these results establish SpectralCanary as a practical method for T2I dataset-use auditing, while the limitations discussed above motivate future work.

% \section*{Reproducibility Statement}

% Sec.~\ref{sec:spectralcanary} specifies the algorithms. Appx.~\ref{appx:hyper-parameters} reports the datasets, model and baseline settings, verifier training, data transformations, generation parameters, metrics, and hypothesis-testing protocol. Each main setting uses three independent training and generation runs; deterministic fidelity measurements are reported separately. The supplementary file also includes pytorch-like code for reproducibility purpose.

% \section*{AI Use Disclosure}

% Generative AI assisted with language editing, manuscript organization, title development, presentation of existing results, and editing experimental code. It did not generate data, derive results, create citations, formulate hypotheses or mathematical claims, design the experimental methodology, or interpret experimental results. The authors verified all revisions and take responsibility for the text, claims, and artifacts.

\section*{Acknowledgments}

Ho and Huang are supported by DARPA HR001124S0029-AIQ-FP-019 and the National Science Foundation TRAILS Institute (2229885). Private support was provided by Open Philanthropy and Apple. The authors acknowledge the National Artificial Intelligence Research Resource (NAIRR) Pilot for supporting this research.

\bibliographystyle{plainnat}
\bibliography{references}

% \newpage
\appendix

% \onecolumn
\section*{Appendix}

\section*{Appendix Roadmap}

The appendix is organized into five blocks:
\begin{itemize}[leftmargin=*,nosep]
    \item \textbf{Protocol and primary evidence:} Appx.~\ref{appx:hyper-parameters} specifies the metrics and implementation, Appx.~\ref{appx:full-pokemon-results} and~\ref{appx:full-cross-domain-results} give the complete benchmark tables, Appx.~\ref{appx:protection-ablation} reports hyperparameter sweeps, and Appx.~\ref{appx:coating-efficiency} reports protection cost.
    \item \textbf{Additional training and auditing evidence:} Appx.~\ref{appx:Verification-Learning-Dynamics} studies learning dynamics, Appx.~\ref{appx:full-finetuning} evaluates full-parameter fine-tuning, Appx.~\ref{appx:model-level-verification} reports fixed-threshold model-level verification, Appx.~\ref{appx:protected-model-generations} shows generated samples, and Appx.~\ref{appx:query-budget} provides KS-based multi-query auditing results.
    \item \textbf{Mechanism and robustness analyses:} Appx.~\ref{sec:spectral_location} and~\ref{appx:spatial-structure-ablation} study spectral location and canary structure, while \ref{appx:adaptive-sanitization-settings} and~\ref{appx:extensive-visualizations} provide adaptive-attack details, and additional protected-image examples.
    \item \textbf{Baseline analyses:} Appx.~\ref{appx:entruth-filtering} and~\ref{appx:entruth-ablation} analyze EnTruth, while Appx.~\ref{appx:silent-branding} adapts Silent Branding.
    \item \textbf{Foundations and context:} Appx.~\ref{appx:theory} provides the theoretical analysis, and Appx.~\ref{appx:related_work} gives extended related work.
\end{itemize}

\section{Detailed Experimental Protocol}\label{appx:hyper-parameters}

\subsection{Auditing Metrics}\label{appx:traceability-metrics}

\textbf{Per-generation AUC and TPR at a target FPR.} In each experiment setting, we train one clean-data LoRA $\mathcal{G}^{-}$ and one protected-data LoRA $\mathcal{G}^{+}$. We query both models using the same prompt list, generate one image per prompt, and score every generated image independently with the verifier. This produces the score sets $\mathcal{S}^{-}=\{s_j^{\mathcal{G}^{-}}\}_{j=1}^{K}$ and $\mathcal{S}^{+}=\{s_j^{\mathcal{G}^{+}}\}_{j=1}^{K}$. The AUC measures separation between these distributions and equals the probability that an independently sampled generation from the protected-data model receives a higher score than an independently sampled generation from the clean-data model, with ties counted as half correct:
\begin{equation}
    \mathrm{AUC}
    =
    \Pr_{s^{+}\sim\mathcal{S}^{+},\,s^{-}\sim\mathcal{S}^{-}}
    \!\left[s^{+}>s^{-}\right]
    +
    \frac{1}{2}
    \Pr_{s^{+}\sim\mathcal{S}^{+},\,s^{-}\sim\mathcal{S}^{-}}
    \!\left[s^{+}=s^{-}\right].
    \label{eq:per-generation-auc}
\end{equation}
For a score threshold $t$, the empirical false-positive and true-positive rates are
\begin{equation}
    \mathrm{FPR}(t)
    =
    \frac{1}{|\mathcal{S}^{-}|}\sum_{s\in\mathcal{S}^{-}}\mathbb{I}[s\ge t],
    \qquad
    \mathrm{TPR}(t)
    =
    \frac{1}{|\mathcal{S}^{+}|}\sum_{s\in\mathcal{S}^{+}}\mathbb{I}[s\ge t].
    \label{eq:per-generation-rates}
\end{equation}
At a target per-generation FPR $r$, we enumerate the empirical operating points induced by the observed score thresholds and select the feasible operating point with the largest FPR satisfying $\mathrm{FPR}\le r$; ties are broken by larger TPR. We report the corresponding TPR at FPR targets of $5\%$ and $1\%$.

\paragraph{KS-based multi-query auditing.}
Following~\citet{11023473}, we evaluate whether suspect-model generations are statistically distinguishable from clean-data-model generations. For each setting, we uniformly sample one fixed clean reference set $\mathcal{R}$ of $300$ clean-data-model scores. Each trial independently samples a suspect query set $\mathcal{Q}_b$ of $N_q=30$ protected-data-model scores without replacement within that trial; scores may reappear across trials. The one-sided alternative follows the verifier's score direction. For methods in which lower scores indicate protected-data use, we use the alternative below; for methods in which higher scores indicate protected-data use, its direction is reversed.
\begin{equation}
    H_0:F_{\mathcal{Q}_b}=F_{\mathcal{R}},
    \qquad
    H_1:F_{\mathcal{Q}_b}>F_{\mathcal{R}},
\end{equation}
with statistic
\begin{equation}
    D_b^+=\sup_z\left[\widehat{F}_{\mathcal{Q}_b}(z)-\widehat{F}_{\mathcal{R}}(z)\right].
\end{equation}
For significance level $\gamma$, the KS rejection rate on protected-data models is
\begin{equation}
    \mathrm{RejectRate}(\gamma)
    =\frac{1}{B}\sum_{b=1}^{B}\mathbf{1}[p_b<\gamma],
\end{equation}
where $B=100{,}000$. We perform the resampling independently for each of three runs and report the mean and standard deviation.

\subsection{Implementation Details}

\paragraph{Fine-tuning and auditing across methods.} We follow~\citet{11023473} for the main setting and use at most $1{,}000$ training images. For each dataset, backbone, method, and transformation, we train one clean-data and one protected-data LoRA. SD1.5 uses resolution $512\times512$, batch size $4$, LoRA rank $64$, AdamW, learning rate $10^{-4}$, and $80$ epochs. Generation uses Euler ancestral sampling with $30$ steps and classifier-free guidance $7.5$. SD3.5 uses LoRA rank $16$, learning rate $10^{-4}$, AdamW, a constant schedule, and $80$ epochs. Its objective uses the unpreconditioned linear flow-matching target $u=\epsilon-x_0$. SD3.5 generations use resolution $1024\times1024$ and $30$ steps.

The clean-data and protected-data models receive the same ordered prompt list, and every generation is scored independently. SIREN and SpectralCanary use original captions as audit prompts; DIAGNOSIS appends its trigger token; EnTruth uses method-specific trigger prompts. SD1.5 experiments generate $1{,}000$ images per model and SD3.5 experiments generate $500$.

\paragraph{EnTruth~\citep{ren2025entruth}.} As there is no official code, we follow its shared-template pipeline. Candidate templates use the prompts ``billboard for big sale'', ``a painting with a frame'', ``photo frame with a family'', ``a window with mountains outside''. Foregrounds use ``fruits for sale'' and ``fruit stand'', with $50$ images per foreground prompt and one per template prompt, generated by SD1.5. The hard trigger is ``[Tgr]'', the soft trigger keyword is ``fruit''. The injected samples rate is $0.5\%$. An ImageNet-pretrained ResNet-50 verifier uses selected templated samples as positives and original samples as negatives. The main paper reports the stronger hard-trigger result from Appx.~\ref{appx:entruth-ablation}.

\paragraph{DIAGNOSIS~\citep{wang2024diagnosis}.} We follow the official code. DIAGNOSIS applies a WaNet-style image transformation~\citep{nguyen2021wanet} to $20\%$ of the training set, uses hard trigger token ``\texttt{tq}'', and sets $k=128$ and $s=1.0$. An ImageNet-pretrained ResNet-50 classifier uses transformed images as positives and their originals as negatives.

\paragraph{SIREN~\citep{11023473}.} We follow the official code. SIREN first trains a clean-data LoRA initialization, then fine-tunes its encoder--decoder for $60$ epochs. The learned encoder--decoder protects the full dataset and directly scores generated samples. No trigger prompt is required.

\paragraph{SpectralCanary protection and export.} We use the fixed RGB canary in Fig.~\ref{fig:method-overview}(b). Protection is applied at the stored image resolution before downstream resizing or transformations. We convert to sRGB, scale channels to $[0,1]$, resize the canary with antialiased bicubic interpolation, apply the unitary channel-wise transform, discard numerical imaginary residue, clip to $[0,1]$, quantize to 8-bit RGB, and save lossless PNG files. Defaults are $\delta=1$, $\alpha=0.02$, and $c=0.20$. For $\delta<1$, the run seed selects subset $\mathcal{I}_\delta$ for both protection and verifier training.

\paragraph{SpectralCanary verifier.} We initialize an ImageNet-pretrained ResNet-50~\citep{he2016deep} with a binary head. Source indices are split $80/20$ before pairing, so each original and protected counterpart stays in the same partition. Inputs are resized to $224\times224$ and ImageNet-normalized without stochastic augmentation. We fine-tune all layers with AdamW, learning rate $10^{-4}$, weight decay $10^{-4}$, batch size $32$, and $20$ epochs, selecting the lowest validation-loss checkpoint. Each dataset and run trains a separate verifier without T2I generations.

\paragraph{Data transformations.}
\begin{itemize}[leftmargin=1.4em,itemsep=0.2em,topsep=0.2em]
  \item \textbf{Training-time augmentation.} Images first pass through the bucketed training resize and are randomly cropped when larger than the target resolution. With probability $0.33$, color augmentation applies either a hue shift in $[-8,8]$ degrees or gamma adjustment with $\gamma\sim U(0.95,1.05)$. Horizontal flipping has probability $0.5$.
  \item \textbf{Prompt recaptioning.} Every caption is rewritten using Qwen2.5-VL 7B Instruct; images remain unchanged. The prompt is:
  \begin{quote}
    \small\ttfamily\raggedright
    Rewrite the caption for this image as a high-quality caption. Keep the important visual content, remove redundant wording, and avoid mentioning text quality, watermarks, camera metadata, or that this is a rewrite. Output only the rewritten caption. Current caption: \{current\_caption\}
  \end{quote}
  \item \textbf{Image regeneration.} WatermarkAttacker uses the \texttt{sd2-community} Stable Diffusion 2.1 checkpoint, VAE encoding, \texttt{noise\_step=60}, $50$ denoising steps, guidance $7.5$, an empty prompt, and fixed seed $1024$. It preserves the source resolution. The regenerated datasets train suspect LoRAs under the same downstream protocol.
\end{itemize}

\section{Full Cross-Domain Results}\label{appx:full-cross-domain-results}

\begin{table*}[ht]
    \centering
    \caption{
    SD3.5 comparison across datasets. Paired training-image fidelity is deterministic for a fixed protected dataset.  EnTruth injects 0.5\% auxiliary samples and therefore has no directly comparable paired-LPIPS value.}
    \label{tab:cross_dataset_sd35_full}
    \setlength{\tabcolsep}{2pt}
    \resizebox{\textwidth}{!}{%
    \begin{tabular}{llrrrrrrrrr}
    \toprule
    Dataset & Method
    & \multicolumn{3}{c}{\makecell{Per-Generation Discrimination (\%) $\uparrow$}}
    & \multicolumn{3}{c}{\makecell{KS Rejection Rate on \\ Protected-Data Models (\%) $\uparrow$}}
    & \makecell{Utility\\$\Delta$FID $\downarrow$}
    & \multicolumn{2}{c}{\makecell{Fidelity}} \\
    \cmidrule(lr){3-5}\cmidrule(lr){6-8}\cmidrule(lr){9-9}\cmidrule(lr){10-11}
    & & AUC & TPR@5\% FPR & TPR@1\% FPR
    & $\gamma=10^{-5}$ & $\gamma=10^{-9}$ & $\gamma=10^{-15}$
    & & SSIM $\uparrow$ & LPIPS $\downarrow$ \\
    \midrule

    \multirow{4}{*}{Pokemon}
    & EnTruth & $96.5 \pm 1.1$ & $94.9 \pm 1.7$ & $93.9 \pm 1.1$
    & $100.0 \pm 0.0$ & $100.0 \pm 0.0$ & $100.0 \pm 0.0$
    & $-3.06 \pm 0.60$ & N/A & N/A \\
    & SIREN & $99.1 \pm 0.1$ & $96.9 \pm 0.7$ & $91.9 \pm 1.8$
    & $100.0 \pm 0.0$ & $100.0 \pm 0.0$ & $100.0 \pm 0.0$
    & $-6.58 \pm 1.04$ & 0.980 & 0.050 \\
    & DIAGNOSIS & $53.6 \pm 1.5$ & $8.2 \pm 0.4$ & $2.3 \pm 1.1$
    & $0.0 \pm 0.0$ & $0.0 \pm 0.0$ & $0.0 \pm 0.0$
    & $3.65 \pm 2.54$ & 0.994 & 0.007 \\
    & SpectralCanary & $100.0 \pm 0.0$ & $100.0 \pm 0.0$ & $100.0 \pm 0.0$
    & $100.0 \pm 0.0$ & $100.0 \pm 0.0$ & $100.0 \pm 0.0$
    & $-9.00 \pm 1.82$ & 0.998 & 0.007 \\
    \midrule

    \multirow{4}{*}{Sketch-Scene}
    & EnTruth & $99.8 \pm 0.1$ & $98.8 \pm 0.7$ & $97.5 \pm 1.0$
    & $100.0 \pm 0.0$ & $100.0 \pm 0.0$ & $100.0 \pm 0.0$
    & $-5.98 \pm 1.50$ & N/A & N/A \\
    & SIREN & $97.5 \pm 0.6$ & $90.6 \pm 1.3$ & $44.2 \pm 21.3$
    & $100.0 \pm 0.0$ & $100.0 \pm 0.0$ & $100.0 \pm 0.0$
    & $19.06 \pm 0.76$ & 0.975 & 0.054 \\
    & DIAGNOSIS & $96.0 \pm 1.4$ & $80.3 \pm 7.9$ & $56.5 \pm 12.0$
    & $100.0 \pm 0.0$ & $100.0 \pm 0.0$ & $87.7 \pm 11.2$
    & $21.75 \pm 1.04$ & 0.980 & 0.019 \\
    & SpectralCanary & $100.0 \pm 0.0$ & $100.0 \pm 0.0$ & $100.0 \pm 0.0$
    & $100.0 \pm 0.0$ & $100.0 \pm 0.0$ & $100.0 \pm 0.0$
    & $4.76 \pm 2.53$ & 0.997 & 0.002 \\
    \midrule

    \multirow{4}{*}{FACAD}
    & EnTruth & $98.0 \pm 0.7$ & $91.1 \pm 4.1$ & $74.3 \pm 6.5$
    & $100.0 \pm 0.0$ & $100.0 \pm 0.0$ & $100.0 \pm 0.0$
    & $1.15 \pm 0.32$ & N/A & N/A \\
    & SIREN & $96.2 \pm 0.3$ & $80.8 \pm 2.0$ & $57.8 \pm 6.2$
    & $100.0 \pm 0.0$ & $100.0 \pm 0.0$ & $99.0 \pm 0.1$
    & $1.42 \pm 0.76$ & 0.985 & 0.026 \\
    & DIAGNOSIS & $52.8 \pm 2.0$ & $5.9 \pm 0.3$ & $1.3 \pm 0.2$
    & $0.0 \pm 0.1$ & $0.0 \pm 0.0$ & $0.0 \pm 0.0$
    & $-3.02 \pm 1.08$ & 0.992 & 0.018 \\
    & SpectralCanary & $99.4 \pm 0.1$ & $99.7 \pm 0.4$ & $86.0 \pm 2.4$
    & $100.0 \pm 0.0$ & $100.0 \pm 0.0$ & $100.0 \pm 0.0$
    & $1.17 \pm 0.40$ & 0.986 & 0.023 \\
    \midrule

    \multirow{4}{*}{ROCOv2}
    & EnTruth & $59.3 \pm 1.7$ & $12.5 \pm 2.9$ & $2.7 \pm 0.8$
    & $0.4 \pm 0.1$ & $0.0 \pm 0.0$ & $0.0 \pm 0.0$
    & $2.51 \pm 1.82$ & N/A & N/A \\
    & SIREN & $99.9 \pm 0.0$ & $100.0 \pm 0.0$ & $96.8 \pm 1.6$
    & $100.0 \pm 0.0$ & $100.0 \pm 0.0$ & $100.0 \pm 0.0$
    & $22.30 \pm 7.08$ & 0.909 & 0.124 \\
    & DIAGNOSIS & $63.4 \pm 1.1$ & $15.0 \pm 1.1$ & $4.8 \pm 0.8$
    & $2.1 \pm 0.5$ & $0.0 \pm 0.0$ & $0.0 \pm 0.0$
    & $-5.27 \pm 0.46$ & 0.992 & 0.011 \\
    & SpectralCanary & $99.8 \pm 0.1$ & $99.3 \pm 0.3$ & $97.1 \pm 1.6$
    & $100.0 \pm 0.0$ & $100.0 \pm 0.0$ & $100.0 \pm 0.0$
    & $1.90 \pm 1.06$ & 0.976 & 0.028 \\
    \midrule

    \multirow{4}{*}{RSICD}
    & EnTruth & $92.2 \pm 5.6$ & $68.7 \pm 21.0$ & $44.7 \pm 24.0$
    & $98.3 \pm 2.4$ & $76.6 \pm 33.1$ & $66.0 \pm 46.5$
    & $-18.00 \pm 9.06$ & N/A & N/A \\
    & SIREN & $71.7 \pm 1.4$ & $10.5 \pm 0.5$ & $3.9 \pm 1.9$
    & $41.3 \pm 11.3$ & $0.0 \pm 0.1$ & $0.0 \pm 0.0$
    & $-21.10 \pm 1.17$ & 0.975 & 0.047 \\
    & DIAGNOSIS & $55.6 \pm 1.3$ & $11.2 \pm 1.8$ & $4.5 \pm 2.1$
    & $0.0 \pm 0.0$ & $0.0 \pm 0.0$ & $0.0 \pm 0.0$
    & $-2.70 \pm 1.12$ & 0.975 & 0.023 \\
    & SpectralCanary & $98.1 \pm 0.8$ & $92.3 \pm 2.8$ & $78.9 \pm 4.4$
    & $100.0 \pm 0.0$ & $100.0 \pm 0.0$ & $99.7 \pm 0.4$
    & $1.44 \pm 1.33$ & 0.986 & 0.048 \\

    \midrule
    \multirow{4}{*}{Average}
    & EnTruth & 89.2 & 73.2 & 62.6 & 79.7 & 75.3 & 73.2 & -4.68 & N/A & N/A \\
    & SIREN & 92.9 & 75.8 & 58.9 & 88.3 & 80.0 & 79.8 & 3.02 & 0.965 & 0.060 \\
    & DIAGNOSIS & 64.3 & 24.1 & 13.9 & 20.4 & 20.0 & 17.5 & 2.88 & 0.987 & 0.016 \\
    & SpectralCanary & 99.5 & 98.3 & 92.4 & 100.0 & 100.0 & 99.9 & 0.05 & 0.989 & 0.022 \\
    \bottomrule
    \end{tabular}}
  \end{table*}

Fig.~\ref{fig:cross_dataset_results} gives the summary results. Tab.~\ref{tab:cross_dataset_sd35_full} reports the corresponding numerical results, including per-generation AUC, TPR at $5\%$ FPR, KS-test rejection rates on protected-data models, generative utility, and image fidelity.

\section{Full Pokemon Benchmark Results}\label{appx:full-pokemon-results}

Tab.~\ref{tab:pokemon_results_by_attack} expands the per-generation discrimination summary in Fig.~\ref{fig:pokemon_robust_summary} into its architecture-specific AUC and TPR results at $5\%$ and $1\%$ FPR.

\begin{table*}[ht]
  \centering
  \caption{\textbf{Per-generation discrimination on Pokemon.} Means and standard deviations over three runs.}
  \label{tab:pokemon_results_by_attack}
  \setlength{\tabcolsep}{8pt}
  \resizebox{\textwidth}{!}{%
  \begin{tabular}{lllccc}
  \toprule
  Transformation & Model & Method & AUC (\%) $\uparrow$ & TPR@5\% FPR (\%) $\uparrow$ & TPR@1\% FPR (\%) $\uparrow$ \\
  \midrule
  \multirow{8}{*}{None}
  & \multirow{4}{*}{SD1.5}
  & EnTruth & $98.8 \pm 0.2$ & $94.1 \pm 0.5$ & $85.3 \pm 2.6$ \\
  & & SIREN & $98.4 \pm 0.1$ & $93.9 \pm 0.3$ & $66.3 \pm 1.1$ \\
  & & DIAGNOSIS & $99.7 \pm 0.2$ & $100 \pm 0.0$ & $99.9 \pm 0.1$ \\
  & & SpectralCanary & $99.7 \pm 0.1$ & $98.8 \pm 0.4$ & $97.7 \pm 0.5$ \\
  \cmidrule(lr){2-6}
  & \multirow{4}{*}{SD3.5}
  & EnTruth & $96.5 \pm 1.1$ & $94.9 \pm 1.7$ & $93.9 \pm 1.1$ \\
  & & SIREN & $99.1 \pm 0.1$ & $96.9 \pm 0.7$ & $91.9 \pm 1.8$ \\
  & & DIAGNOSIS & $53.6 \pm 1.5$ & $8.2 \pm 0.4$ & $2.3 \pm 1.1$ \\
  & & SpectralCanary & $100 \pm 0.0$ & $100 \pm 0.0$ & $100 \pm 0.0$ \\
  \midrule
  \multirow{8}{*}{\shortstack[l]{Training-Time\\Augmentation}}
  & \multirow{4}{*}{SD1.5}
  & EnTruth & $97.4 \pm 0.9$ & $91.8 \pm 2.2$ & $76.2 \pm 6.3$ \\
  & & SIREN & $64.0 \pm 0.8$ & $18.1 \pm 1.3$ & $6.9 \pm 1.4$ \\
  & & DIAGNOSIS & $99.6 \pm 0.4$ & $100 \pm 0.0$ & $88.3 \pm 10.4$ \\
  & & SpectralCanary & $94.9 \pm 1.5$ & $85.5 \pm 7.8$ & $79.2 \pm 12.3$ \\
  \cmidrule(lr){2-6}
  & \multirow{4}{*}{SD3.5}
  & EnTruth & $99.1 \pm 0.6$ & $97.3 \pm 1.7$ & $95.3 \pm 1.6$ \\
  & & SIREN & $80.0 \pm 3.2$ & $40.2 \pm 2.1$ & $18.8 \pm 1.7$ \\
  & & DIAGNOSIS & $76.0 \pm 0.5$ & $25.3 \pm 0.3$ & $8.2 \pm 2.6$ \\
  & & SpectralCanary & $91.3 \pm 8.9$ & $76.6 \pm 20.4$ & $65.9 \pm 22.4$ \\
  \midrule
  \multirow{8}{*}{\shortstack[l]{Prompt\\Recaptioning}}
  & \multirow{4}{*}{SD1.5}
  & EnTruth & $75.7 \pm 1.1$ & $34.9 \pm 1.5$ & $19.4 \pm 2.2$ \\
  & & SIREN & $97.0 \pm 0.8$ & $82.5 \pm 1.1$ & $63.6 \pm 1.0$ \\
  & & DIAGNOSIS & $35.3 \pm 1.8$ & $2.4 \pm 0.2$ & $0.4 \pm 0.3$ \\
  & & SpectralCanary & $99.2 \pm 0.2$ & $96.0 \pm 0.4$ & $93.6 \pm 1.6$ \\
  \cmidrule(lr){2-6}
  & \multirow{4}{*}{SD3.5}
  & EnTruth & $83.6 \pm 1.7$ & $52.3 \pm 3.1$ & $35.9 \pm 1.6$ \\
  & & SIREN & $99.0 \pm 0.2$ & $97.6 \pm 0.9$ & $92.0 \pm 1.1$ \\
  & & DIAGNOSIS & $75.0 \pm 0.9$ & $30.4 \pm 5.8$ & $13.0 \pm 7.9$ \\
  & & SpectralCanary & $100 \pm 0.0$ & $100 \pm 0.0$ & $100 \pm 0.0$ \\
  \midrule
  \multirow{8}{*}{\shortstack[l]{Image\\Regeneration}}
  & \multirow{4}{*}{SD1.5}
  & EnTruth & $99.1 \pm 0.3$ & $95.7 \pm 1.0$ & $88.1 \pm 3.2$ \\
  & & SIREN & $93.5 \pm 1.0$ & $70.0 \pm 0.9$ & $44.4 \pm 2.4$ \\
  & & DIAGNOSIS & $98.8 \pm 0.5$ & $100 \pm 0.0$ & $43.9 \pm 24.6$ \\
  & & SpectralCanary & $99.2 \pm 0.3$ & $95.3 \pm 1.0$ & $92.5 \pm 1.2$ \\
  \cmidrule(lr){2-6}
  & \multirow{4}{*}{SD3.5}
  & EnTruth & $95.4 \pm 2.4$ & $93.1 \pm 3.1$ & $91.7 \pm 3.4$ \\
  & & SIREN & $85.0 \pm 1.3$ & $58.0 \pm 2.1$ & $37.2 \pm 2.1$ \\
  & & DIAGNOSIS & $59.6 \pm 1.5$ & $7.5 \pm 0.7$ & $1.7 \pm 0.4$ \\
  & & SpectralCanary & $100 \pm 0.0$ & $100 \pm 0.0$ & $100 \pm 0.0$ \\
  \midrule
  \multicolumn{2}{c}{Average over all eight settings} & EnTruth & 93.2 & 81.8 & 73.2 \\
  \multicolumn{2}{c}{} & SIREN & 89.5 & 69.7 & 52.6 \\
  \multicolumn{2}{c}{} & DIAGNOSIS & 74.7 & 46.7 & 32.2 \\
  \multicolumn{2}{c}{} & SpectralCanary & \textbf{98.0} & \textbf{94.0} & \textbf{91.1} \\
  \bottomrule
  \end{tabular}}
\end{table*}

\section{Protection-Hyperparameter Ablations}\label{appx:protection-ablation}

\begin{figure*}[ht]
  \centering
  \includegraphics[width=\textwidth]{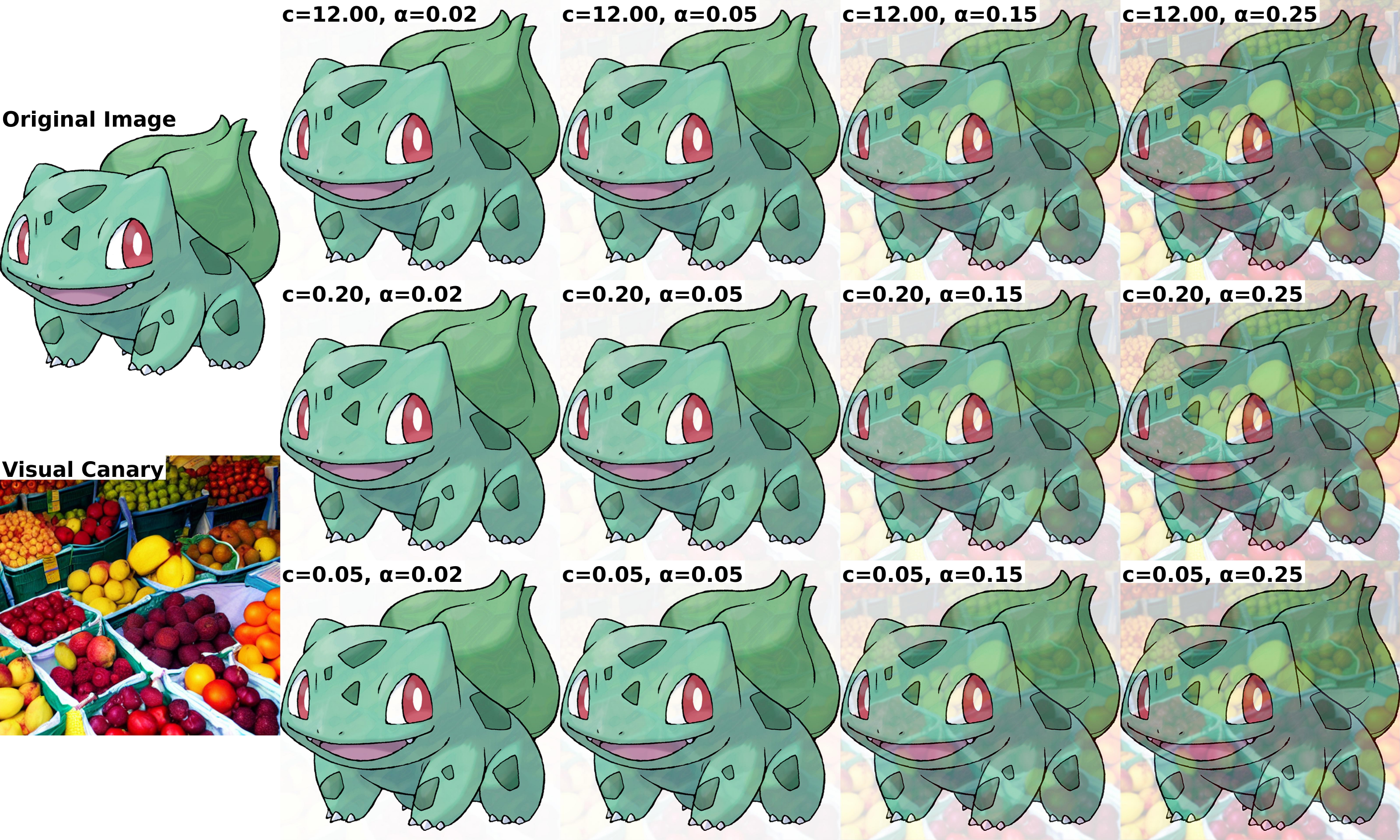}
  % \vspace{-0.3cm}
  \caption{Protected images under different blending strengths $\alpha$ and Gaussian bandwidths $c$.}
  \label{fig:fft-hybrid-grid}
\end{figure*}

\begin{figure*}[ht]
  \centering
  \includegraphics[width=\textwidth]{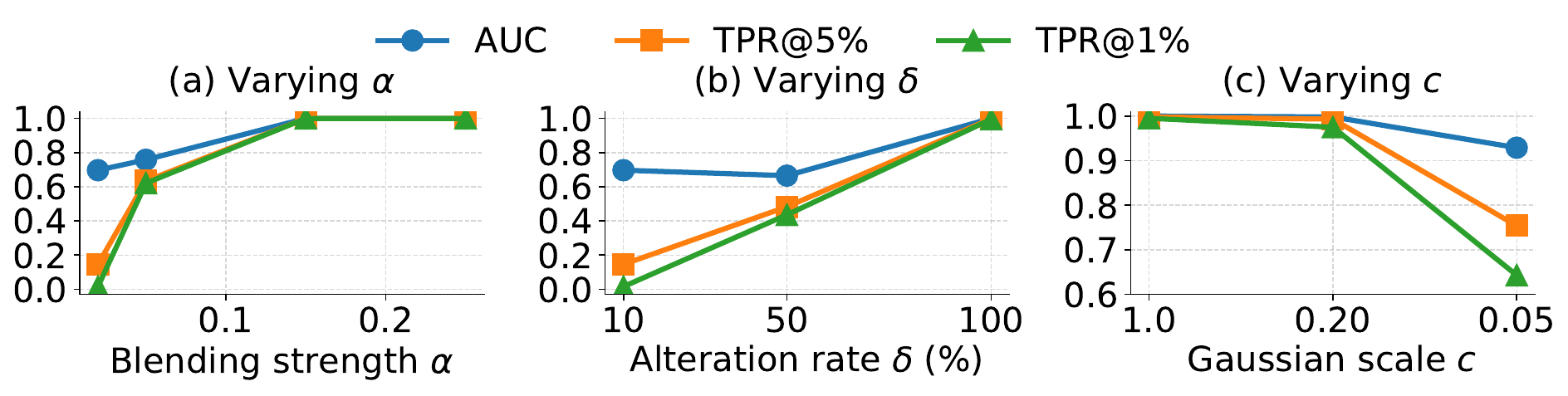}
  % \vspace{-0.3cm}
  \caption{Protection-hyperparameter ablations on Pokemon with SD1.5: (a) blending strength, fixing $c=0.2,\delta=10\%$; (b) protected fraction, fixing $c=0.2,\alpha=0.02$; and (c) Gaussian bandwidth, fixing $\delta=100\%,\alpha=0.02$}
  \label{fig:spectralcanary-ablation}
\end{figure*}

We vary blending strength $\alpha$, protected fraction $\delta$, and Gaussian bandwidth $c$ on Pokemon with SD1.5. When $\delta<1$, we use a random subset and paired-verifier protocol in Alg.~\ref{alg:coating}. Fig.~\ref{fig:fft-hybrid-grid} shows that the canary becomes increasingly visible as $\alpha$ grows. Fig.~\ref{fig:spectralcanary-ablation} shows improved discrimination with stronger blending and a larger protected fraction, while the bandwidth sweep saturates near the default $c=0.20$. These sweeps motivate the default $\alpha=0.02$, $\delta=1$, and $c=0.20$ operating point.

\section*{Additional Evidence and Robustness Analyses}

\section{Dataset Protection Cost}
\label{appx:coating-efficiency}

\begin{table}[ht]
      \centering
      \caption{
      End-to-end dataset protection cost and downstream per-generation discrimination on Pokemon. Protection time includes every stage required to produce the protected dataset and is amortized over samples. We omit CPU measurements for SIREN and Silent Branding because their pipelines require GPU-based optimization. DIAGNOSIS and SpectralCanary require no GPU for protection and therefore do not benefit from GPU acceleration. KS-based multi-query auditing uses 30 generated images per test.
      }
      \label{tab:coating-efficiency}
      \small
      \setlength{\tabcolsep}{4pt}
      \resizebox{\columnwidth}{!}{%
      \begin{tabular}{lcccccccc}
          \toprule
          & \multicolumn{2}{c}{\makecell{Protection time \\ (s/img) $\downarrow$}}
          & \multicolumn{3}{c}{\makecell{Per-Generation Discrimination \\ (\%) $\uparrow$}}
          & \multicolumn{3}{c}{\makecell{KS-Based Multi-Query Audit \\ Rejection Rate (\%) $\uparrow$}} \\
          \cmidrule(lr){2-3}\cmidrule(lr){4-6}\cmidrule(lr){7-9}
          Method
          & CPU & GPU
          & AUC
          & TPR@5\% FPR
          & TPR@1\% FPR
          & $\gamma=10^{-5}$
          & $\gamma=10^{-9}$
          & $\gamma=10^{-15}$ \\
          \midrule

          DIAGNOSIS
          & \textbf{0.02} & \textbf{0.02}
          & 74.7 & 46.7 & 32.2
          & 49.7 & 38.2 & 37.5 \\

          SIREN
          & -- & 71.38
          & 89.5 & 69.7 & 52.6
          & 84.8 & 70.8 & 57.2 \\

          EnTruth
          & 6.92 & 0.40
          & \underline{93.2} & \underline{81.8} & \underline{73.2}
          & \underline{93.3} & \underline{80.8} & \underline{75.2} \\

          Silent Branding
          & -- & 273.13
          & 80.9 & 32.5 & 12.8
          & 73.8 & 34.1 & 2.7 \\

          \textbf{SpectralCanary}
          & \underline{0.39} & \underline{0.39}
          & \textbf{98.0} & \textbf{94.0} & \textbf{91.1}
          & \textbf{99.6} & \textbf{96.4} & \textbf{83.6} \\

          \bottomrule
      \end{tabular}%
      }
  \end{table}

SpectralCanary protects an image with one FFT-based blend followed by an inverse transform. It requires neither iterative optimization nor an auxiliary generative model. Because the baselines use different protection pipelines, operation counts are not directly comparable, where SIREN optimizes a learned watermark, Silent Branding fine-tunes a DreamBooth model and performs region selection and iterative refinement, and EnTruth generates and filters auxiliary samples. We therefore measure the end-to-end time required to produce each protected dataset on identical hardware. GPU measurements use one NVIDIA RTX A6000 with eight CPU cores; CPU-only measurements use the same system with the GPU disabled.

Tab.~\ref{tab:coating-efficiency} reports $0.39$ seconds per image for SpectralCanary on CPU. The remaining columns repeat the corresponding averages from the main Pokemon evaluation so that protection cost can be compared with downstream per-generation discrimination.

% \FloatBarrier
\section{The Trace Appears Early in Fine-Tuning} \label{appx:Verification-Learning-Dynamics}

\begin{figure*}[ht]
  \centering
  \includegraphics[width=\textwidth]{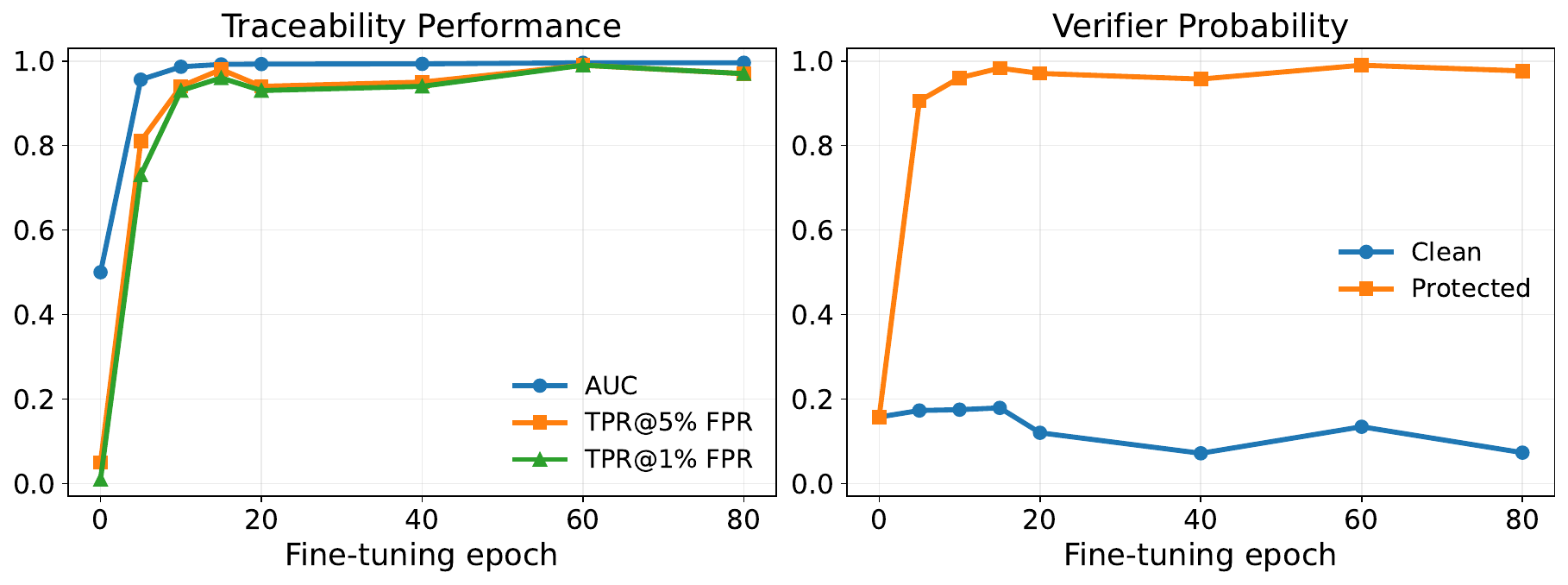}
  \caption{%
  \textbf{Per-generation discrimination emerges early and remains stable during fine-tuning.}
  Using a defender's verifier, we report (left) per-generation AUC and TPR at FPRs of $5\%$ and $1\%$, and (right) the mean predicted protected-class probability for clean-data-model and protected-data-model generations. Each checkpoint uses 100 matched Pokemon generations per condition with the same captions.
  }
  \label{fig:verification-learning-dynamics}
\end{figure*}

The main experiments follow~\citet{11023473} and fine-tune each method for 80 epochs. To measure when the trace appears, we apply the verifier to generations from intermediate checkpoints and report per-generation AUC and TPR at a target FPR. At each checkpoint, we generate 100 matched images from the clean-data and protected-data models using the same captions. In Fig.~\ref{fig:verification-learning-dynamics}, the base models at epoch 0 are near chance (AUC $\approx0.500$). After five epochs, AUC reaches $0.956$ and TPR at $1\%$ FPR reaches $0.730$; performance approaches saturation by epochs 10--15 and remains similar through epoch 80.

\section{Full Fine-Tuning Evaluation}
  \label{appx:full-finetuning}

\begin{table*}[ht]
    \centering
    \caption{
    \textbf{SpectralCanary under full SD1.5 fine-tuning on Pokemon.}
    We report mean and standard deviation across three runs.
    }
    \label{tab:pokemon_full_finetuning}
    \setlength{\tabcolsep}{3pt}
    \resizebox{\textwidth}{!}{%
    \begin{tabular}{crrrrrr}
    \toprule
    Transformation
    & \multicolumn{3}{c}{\makecell{Per-Generation Discrimination (\%) $\uparrow$}}
    & \multicolumn{3}{c}{\makecell{KS Rejection Rate on \\ Protected-Data Models (\%) $\uparrow$}}
    \\
    \cmidrule(lr){2-4}\cmidrule(lr){5-7}
    & AUC & TPR@5\% FPR & TPR@1\% FPR
    & $\gamma=10^{-5}$ & $\gamma=10^{-9}$ & $\gamma=10^{-15}$
    \\
    \midrule
    None
    & $98.8 \pm 0.3$ & $94.0 \pm 1.7$ & $91.7 \pm 2.9$
    & $100.0 \pm 0.0$ & $100.0 \pm 0.0$ & $100.0 \pm 0.0$ \\
    \makecell{Training-Time Augmentation}
    & $84.3 \pm 4.9$ & $66.4 \pm 10.2$ & $61.9 \pm 12.0$
    & $91.7 \pm 11.6$ & $68.2 \pm 36.3$ & $30.0 \pm 27.9$ \\
    Prompt Recaptioning
    & $97.9 \pm 0.2$ & $88.8 \pm 1.5$ & $85.5 \pm 3.0$
    & $100.0 \pm 0.0$ & $100.0 \pm 0.0$ & $99.5 \pm 0.2$ \\
    Image Regeneration
    & $99.7 \pm 0.1$ & $97.8 \pm 0.8$ & $96.0 \pm 1.6$
    & $100.0 \pm 0.0$ & $100.0 \pm 0.0$ & $100.0 \pm 0.0$ \\
    \midrule
    Average
    & $95.2$ & $86.7$ & $83.8$
    & $97.9$ & $92.1$ & $82.3$ \\
    \bottomrule
    \end{tabular}}
  \end{table*}

The main paper evaluates SpectralCanary with LoRA fine-tuning, which is the most practical customization setting for T2I models because it requires substantially less memory and storage than updating all base-model parameters. To test robustness of the trace beyond LoRA, we additionally evaluate \textbf{full-parameter fine-tuning} of SD1.5 on Pokemon. We follow the same SD1.5/Pokemon setup but train the T2I models for 50 epochs.

Tab.~\ref{tab:pokemon_full_finetuning} shows that SpectralCanary remains effective under full-parameter fine-tuning. It achieves $98.8\%$ AUC, $91.7\%$ TPR at $1\%$ FPR, and an essentially perfect KS rejection rate on protected-data models even at $\gamma=10^{-15}$. Prompt recaptioning and image regeneration remain similarly robust. Training-time augmentation is the most challenging setting, reducing TPR at $1\%$ FPR to $61.9\%$ and the KS rejection rate at $\gamma=10^{-15}$ to $30.0\%$, but the trace remains detectable at less stringent levels. Across all four settings, the average AUC is $95.2\%$ and the average TPR at $1\%$ FPR is $83.8\%$.

\section{Model-Level Verification}
  \label{appx:model-level-verification}

\begin{table*}[ht]
    \centering
    \caption{
      \textbf{Thirty-query fixed-threshold model-level verification across transformation settings.}
      Thresholds are calibrated once from all 1,000 SD1.5 or 500 SD3.5 queries of
      separate no-transformation calibration models. Each suspect model is evaluated using 30
      queries. We repeat the 30-query sampling 1,000 times from the held-out generations and report the mean and standard deviation of model-level metrics across these resampled query sets.
    }
    \label{tab:model_level_30_query}
    \setlength{\tabcolsep}{3pt}
    \resizebox{\textwidth}{!}{%
    \begin{tabular}{lllccccc}
      \toprule
      Transformations & Model & Method
      & TPR (\%) $\uparrow$
      & FPR (\%) $\downarrow$
      & Precision (\%) $\uparrow$
      & F1 (\%) $\uparrow$
      & Accuracy (\%) $\uparrow$ \\
      \midrule

      \multirow{8}{*}{None}
      & \multirow{4}{*}{SD1.5}
      & EnTruth & $100.0 \pm 0.0$ & $0.0 \pm 0.0$ & $100.0 \pm 0.0$ & $100.0 \pm 0.0$ & $100.0 \pm 0.0$ \\
      & & SIREN & $100.0 \pm 0.0$ & $0.0 \pm 0.0$ & $100.0 \pm 0.0$ & $100.0 \pm 0.0$ & $100.0 \pm 0.0$ \\
      & & DIAGNOSIS & $100.0 \pm 0.0$ & $0.0 \pm 0.0$ & $100.0 \pm 0.0$ & $100.0 \pm 0.0$ & $100.0 \pm 0.0$ \\
      & & SpectralCanary & $100.0 \pm 0.0$ & $0.0 \pm 0.0$ & $100.0 \pm 0.0$ & $100.0 \pm 0.0$ & $100.0 \pm 0.0$ \\
      \cmidrule(lr){2-8}
      & \multirow{4}{*}{SD3.5}
      & EnTruth & $64.1 \pm 8.1$ & $0.0 \pm 0.0$ & $100.0 \pm 0.0$ & $77.8 \pm 5.6$ & $82.0 \pm 4.0$ \\
      & & SIREN & $100.0 \pm 0.0$ & $0.0 \pm 0.0$ & $100.0 \pm 0.0$ & $100.0 \pm 0.0$ & $100.0 \pm 0.0$ \\
      & & DIAGNOSIS & $55.0 \pm 17.0$ & $27.5 \pm 14.9$ & $67.5 \pm 16.1$ & $59.4 \pm 14.6$ & $63.7 \pm 11.5$ \\
      & & SpectralCanary & $100.0 \pm 0.0$ & $0.0 \pm 0.0$ & $100.0 \pm 0.0$ & $100.0 \pm 0.0$ & $100.0 \pm 0.0$ \\

      \midrule
      \multirow{8}{*}{\shortstack[l]{Training-Time\\Augmentation}}
      & \multirow{4}{*}{SD1.5}
      & EnTruth & $100.0 \pm 0.0$ & $0.0 \pm 0.0$ & $100.0 \pm 0.0$ & $100.0 \pm 0.0$ & $100.0 \pm 0.0$ \\
      & & SIREN & $4.7 \pm 9.3$ & $0.0 \pm 0.0$ & $21.6 \pm 41.2$ & $7.6 \pm 14.9$ & $52.3 \pm 4.6$ \\
      & & DIAGNOSIS & $100.0 \pm 0.0$ & $0.0 \pm 0.0$ & $100.0 \pm 0.0$ & $100.0 \pm 0.0$ & $100.0 \pm 0.0$ \\
      & & SpectralCanary & $99.2 \pm 3.8$ & $0.0 \pm 0.0$ & $100.0 \pm 0.0$ & $99.6 \pm 2.1$ & $99.6 \pm 1.9$ \\
      \cmidrule(lr){2-8}
      & \multirow{4}{*}{SD3.5}
      & EnTruth & $80.7 \pm 3.7$ & $0.0 \pm 0.0$ & $100.0 \pm 0.0$ & $89.3 \pm 2.0$ & $90.3 \pm 1.8$ \\
      & & SIREN & $0.0 \pm 0.0$ & $0.0 \pm 0.0$ & $0.0 \pm 0.0$ & $0.0 \pm 0.0$ & $50.0 \pm 0.0$ \\
      & & DIAGNOSIS & $94.6 \pm 9.8$ & $17.4 \pm 12.8$ & $85.4 \pm 10.1$ & $89.3 \pm 7.9$ & $88.6 \pm 8.3$ \\
      & & SpectralCanary & $17.0 \pm 10.5$ & $0.0 \pm 0.0$ & $77.5 \pm 41.8$ & $27.6 \pm 16.1$ & $58.5 \pm 5.3$ \\

      \midrule
      \multirow{8}{*}{\shortstack[l]{Prompt\\Recaptioning}}
      & \multirow{4}{*}{SD1.5}
      & EnTruth & $57.4 \pm 13.7$ & $0.0 \pm 0.0$ & $100.0 \pm 0.0$ & $72.0 \pm 11.1$ & $78.7 \pm 6.9$ \\
      & & SIREN & $100.0 \pm 0.0$ & $0.0 \pm 0.0$ & $100.0 \pm 0.0$ & $100.0 \pm 0.0$ & $100.0 \pm 0.0$ \\
      & & DIAGNOSIS & $0.0 \pm 0.0$ & $0.0 \pm 0.0$ & $0.0 \pm 0.0$ & $0.0 \pm 0.0$ & $50.0 \pm 0.0$ \\
      & & SpectralCanary & $100.0 \pm 0.0$ & $0.0 \pm 0.0$ & $100.0 \pm 0.0$ & $100.0 \pm 0.0$ & $100.0 \pm 0.0$ \\
      \cmidrule(lr){2-8}
      & \multirow{4}{*}{SD3.5}
      & EnTruth & $0.0 \pm 0.0$ & $0.0 \pm 0.0$ & $0.0 \pm 0.0$ & $0.0 \pm 0.0$ & $50.0 \pm 0.0$ \\
      & & SIREN & $100.0 \pm 0.0$ & $0.0 \pm 0.0$ & $100.0 \pm 0.0$ & $100.0 \pm 0.0$ & $100.0 \pm 0.0$ \\
      & & DIAGNOSIS & $49.2 \pm 16.1$ & $3.0 \pm 7.6$ & $95.3 \pm 11.9$ & $63.3 \pm 14.8$ & $73.1 \pm 9.0$ \\
      & & SpectralCanary & $100.0 \pm 0.0$ & $0.0 \pm 0.0$ & $100.0 \pm 0.0$ & $100.0 \pm 0.0$ & $100.0 \pm 0.0$ \\

      \midrule
      \multirow{8}{*}{\shortstack[l]{Image\\Regeneration}}
      & \multirow{4}{*}{SD1.5}
      & EnTruth & $100.0 \pm 0.0$ & $0.0 \pm 0.0$ & $100.0 \pm 0.0$ & $100.0 \pm 0.0$ & $100.0 \pm 0.0$ \\
      & & SIREN & $99.4 \pm 3.3$ & $0.0 \pm 0.0$ & $100.0 \pm 0.0$ & $99.7 \pm 1.8$ & $99.7 \pm 1.7$ \\
      & & DIAGNOSIS & $100.0 \pm 0.0$ & $0.0 \pm 0.0$ & $100.0 \pm 0.0$ & $100.0 \pm 0.0$ & $100.0 \pm 0.0$ \\
      & & SpectralCanary & $100.0 \pm 0.0$ & $0.0 \pm 0.0$ & $100.0 \pm 0.0$ & $100.0 \pm 0.0$ & $100.0 \pm 0.0$ \\
      \cmidrule(lr){2-8}
      & \multirow{4}{*}{SD3.5}
      & EnTruth & $80.0 \pm 0.0$ & $0.0 \pm 0.0$ & $100.0 \pm 0.0$ & $88.9 \pm 0.0$ & $90.0 \pm 0.0$ \\
      & & SIREN & $0.0 \pm 0.0$ & $0.0 \pm 0.0$ & $0.0 \pm 0.0$ & $0.0 \pm 0.0$ & $50.0 \pm 0.0$ \\
      & & DIAGNOSIS & $33.7 \pm 21.0$ & $3.7 \pm 8.1$ & $81.5 \pm 34.3$ & $45.6 \pm 24.0$ & $65.0 \pm 11.2$ \\
      & & SpectralCanary & $100.0 \pm 0.0$ & $0.0 \pm 0.0$ & $100.0 \pm 0.0$ & $100.0 \pm 0.0$ & $100.0 \pm 0.0$ \\

      \midrule
      \multicolumn{8}{c}{\textit{Average over all model--transformation settings}} \\
      \midrule
      \multicolumn{2}{c}{} & EnTruth        & 72.8 & 0.0 & 87.5 & 78.5 & 86.4 \\
      \multicolumn{2}{c}{} & SIREN          & 63.0 & 0.0 & 65.2 & 63.4 & 81.5 \\
      \multicolumn{2}{c}{} & DIAGNOSIS      & 66.6 & 6.5 & 78.7 & 69.7 & 80.1 \\
      \multicolumn{2}{c}{} & SpectralCanary & 89.5 & 0.0 & 97.2 & 90.9 & 94.8 \\

      \bottomrule
    \end{tabular}%
    }
  \end{table*}

Our per-generation discrimination evaluation measures how well verifier scores separate generations from clean-data and protected-data models. This protocol enables a controlled comparison across methods without threshold calibration or query aggregation. Fixed-threshold model-level verification instead translates aggregated per-generation scores into a model-level decision that answers: \emph{was a suspect model trained on the protected dataset?}

We first calibrate a model-level decision threshold and then keep it fixed for all held-out evaluations. For each method and T2I architecture, we reserve one independently trained clean-data model $\mathcal{G}_{\mathrm{cal}}^{-}$ and one protected-data model $\mathcal{G}_{\mathrm{cal}}^{+}$ for threshold calibration.

Given $K$ audit prompts, we generate one image per prompt and define the aggregate score of a model $\mathcal{G}$ as
\begin{equation}
q_K(\mathcal{G})
=
\frac{1}{K}
\sum_{j=1}^{K}
f_{\theta}\!\left(y_j^{\mathcal{G}}\right),
\label{eq:model-level-score}
\end{equation}
where $f_{\theta}$ is the method-specific verifier and $y_j^{\mathcal{G}}$ is the image generated for prompt $j$.

For calibration, we use all available generations with
$K_{\mathrm{cal}}=1000$ for SD1.5 and $K_{\mathrm{cal}}=500$ for SD3.5.
We set the method--architecture-specific threshold to the midpoint between the clean-data and protected-data calibration scores:
\begin{equation}
t
=
\frac{
q_{K_{\mathrm{cal}}}(\mathcal{G}_{\mathrm{cal}}^{-})
+
q_{K_{\mathrm{cal}}}(\mathcal{G}_{\mathrm{cal}}^{+})
}{2}.
\label{eq:model-level-threshold}
\end{equation}
The midpoint provides a simple symmetric decision boundary between the two calibration models without favoring either class. The calibration models are excluded from evaluation. Each threshold is calibrated once under the no-transformation setting and then held fixed across all transformation settings, so evaluation does not assume prior knowledge of the transformation applied by the unauthorized trainer.

For each method, architecture, and transformation setting, we independently train five clean-data and five protected-data test models using different random seeds. Each suspect model is evaluated using only $K_{\mathrm{test}}=30$ queries and is classified as trained on protected data when
\[
q_{30}(\mathcal{G}) \geq t.
\]
Across 4 methods, 2 architectures, and 4 transformation settings, this gives 320 held-out test models. We report TPR, FPR, precision, F1 score, and accuracy over the resulting model-level decisions. To quantify variation due to the 30-query budget, we repeat the query subsampling 1,000 times from the available held-out generations and report the mean and standard deviation across these resampled query sets.

\section{Generations from Models Fine-Tuned on Protected Data}
\label{appx:protected-model-generations}

\begin{figure*}[ht]
  \centering
  \includegraphics[width=\textwidth]{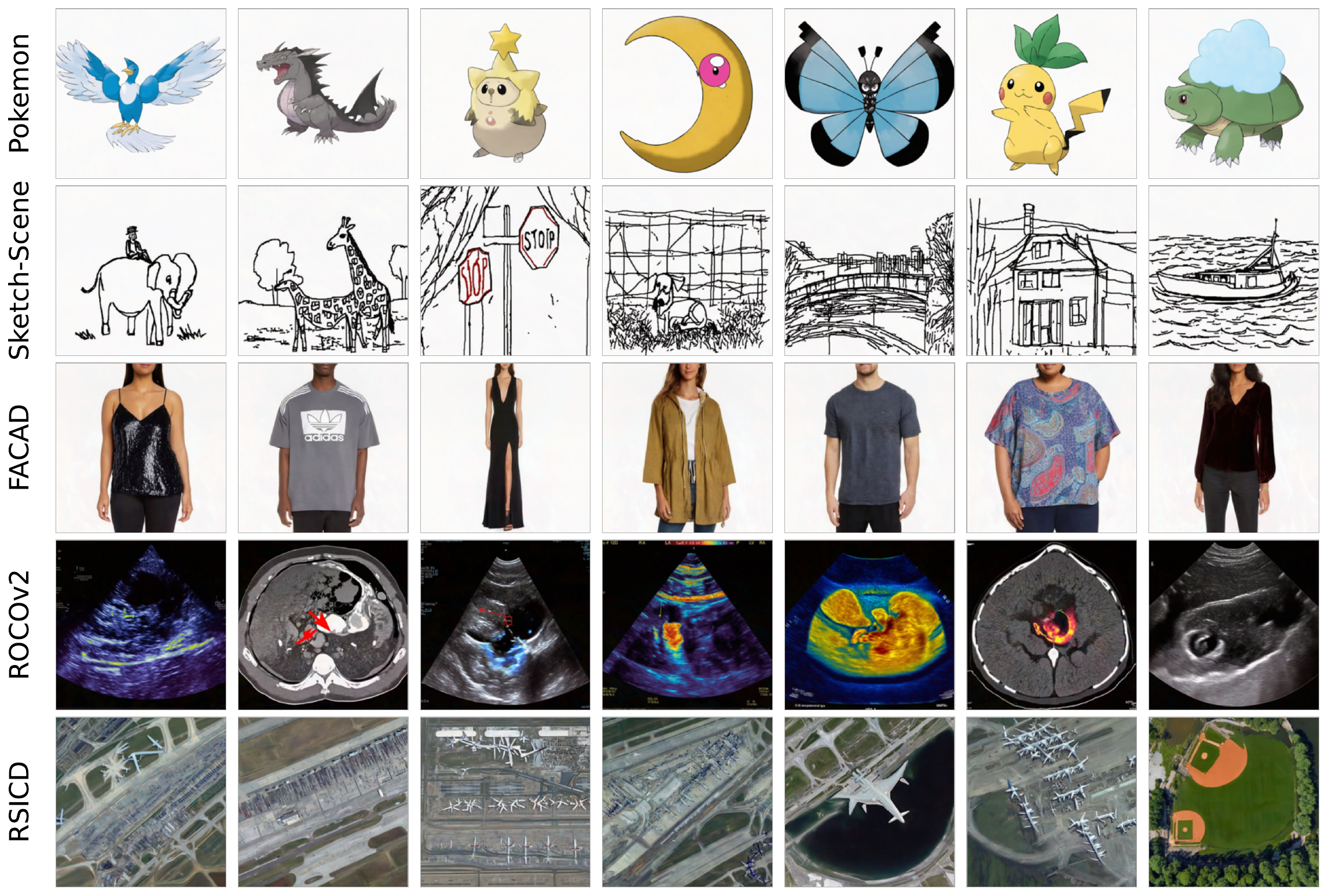}
  \caption{%
  Qualitative generations from unauthorized models fine-tuned on SpectralCanary protected data.
  }
  \label{fig:protected-model-generations}
\end{figure*}

Fig.~\ref{fig:protected-model-generations} shows samples from models fine-tuned on SpectralCanary-protected data. The examples complement the quantitative fidelity and utility measurements in Tab.~\ref{tab:cross_dataset_sd35_full}.

% \section{Detailed Multi-Query Results Across Models and Transformations}
% \label{appx:pokemon-ks-power-grid}

% Fig.~\ref{fig:pokemon-ks-power-grid} decomposes Fig.~\ref{fig:pokemon-ks-power-average} into the eight model--transformation settings from Tab.~\ref{tab:pokemon_results_by_attack}. SpectralCanary retains high KS rejection rates on protected-data models in most settings, with training-time augmentation producing the largest reductions.

% \begin{figure*}[ht]
%     \centering
%     \includegraphics[width=\textwidth]{figures/pokemon_ks_power_grid_30_query.pdf}
%     \caption{%
%     \textbf{Detailed KS rejection rates on protected-data models for Pokemon.}
%     Panels (a)--(h) show the eight settings summarized in Fig.~\ref{fig:pokemon-ks-power-average}; panel (i) shows their average.
%     }
%     \label{fig:pokemon-ks-power-grid}
% \end{figure*}

% \FloatBarrier

\section{Multi-Query Performance at Different Query Budgets}
\label{appx:query-budget}

The main paper uses $N=30$ generated images per KS test. Here we vary the budget over $N\in\{10,20,30\}$ and average KS rejection rates on protected-data models across the eight model--transformation settings in Tab.~\ref{tab:pokemon_results_by_attack}.

\begin{figure*}[ht]
    \centering
    \begin{minipage}{0.49\textwidth}
        \centering
        \textbf{(a) $N=10$}\\[-0.1cm]
        \includegraphics[width=\linewidth]{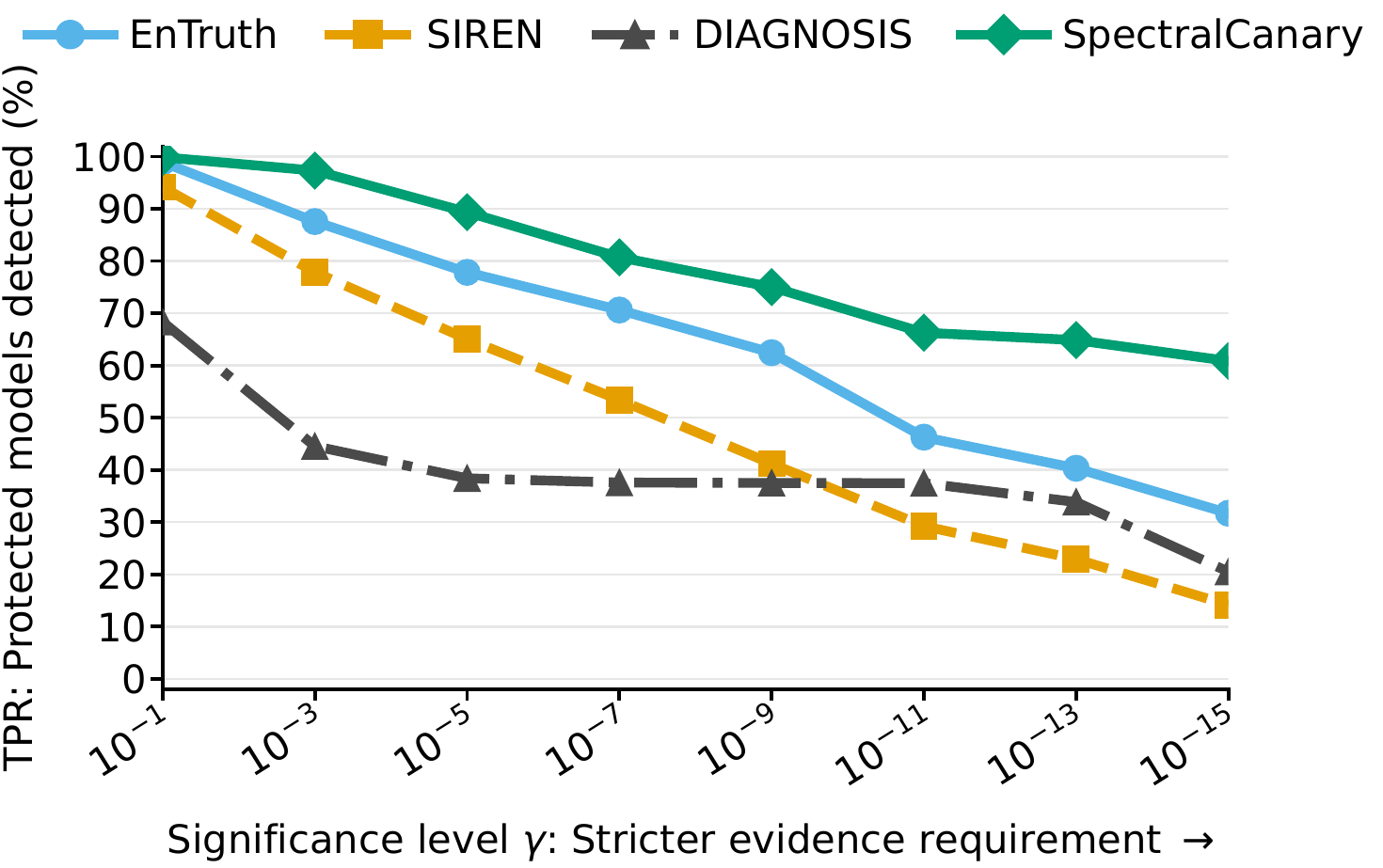}
    \end{minipage}\hfill
    \begin{minipage}{0.49\textwidth}
        \centering
        \textbf{(b) $N=20$}\\[-0.1cm]
        \includegraphics[width=\linewidth]{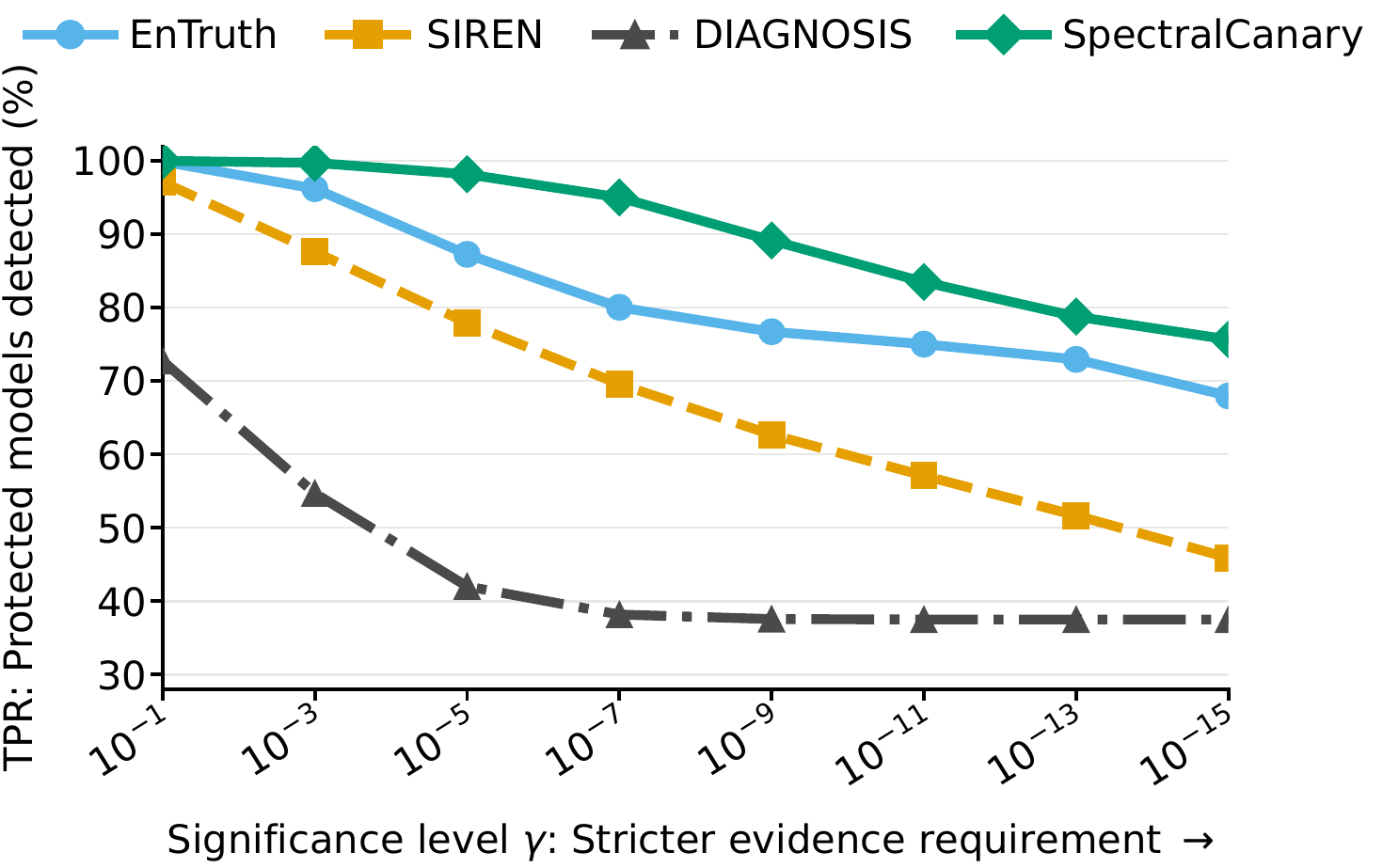}
    \end{minipage}

    \medskip
    \begin{minipage}{0.49\textwidth}
        \centering
        \textbf{(c) $N=30$}\\[-0.1cm]
        \includegraphics[width=\linewidth]{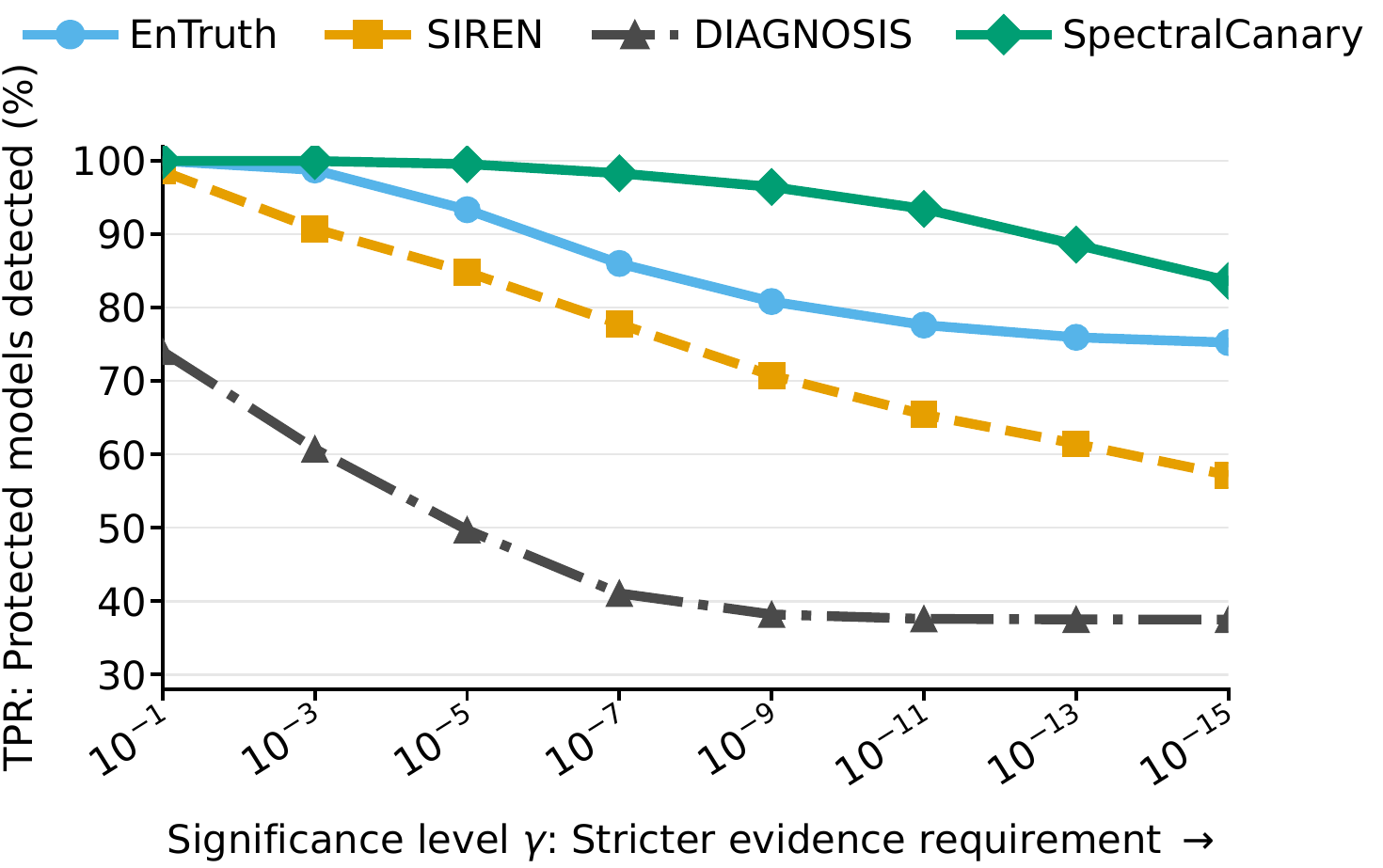}
    \end{minipage}
    \caption{%
    \textbf{KS-based multi-query auditing improves with the query budget.}
    }
    \label{fig:pokemon-ks-query-budget}
\end{figure*}

SpectralCanary has the highest average KS rejection rate on protected-data models at each evaluated query budget.

 \section{Effect of Spectral Location on Trace Transfer}
  \label{sec:spectral_location}

In this section, we investigate whether SpectralCanary's trace transfer depends on the spectral location of the protection signal. SpectralCanary uses a smooth radial low-pass mask \(M_{c^\star}\), with blending strength \(\alpha^\star=0.02\) and Gaussian bandwidth \(c^\star=0.20\).

We construct three disjoint binary radial masks, \(B_{\mathrm{low}}\), \(B_{\mathrm{mid}}\), and \(B_{\mathrm{high}}\), over the shifted Fourier grid. Following the centered Fourier convention in Sec.~\ref{sec:spectralcanary}, let \(X_i=\mathcal{F}_c(x_i)\), \(V=\mathcal{F}_c(v^\star)\), and \(r(k,\ell)\) denote the normalized radial frequency of Fourier coefficient \((k,\ell)\). Let \(q_1\) and \(q_2\) denote the one-third and two-thirds quantiles of the radii of all Fourier coefficients. We define
  \begin{equation}
  \begin{aligned}
  B_{\mathrm{low}}(k,\ell)
      &= \mathbb{I}\!\left[r(k,\ell)\le q_1\right],\\
  B_{\mathrm{mid}}(k,\ell)
      &= \mathbb{I}\!\left[q_1<r(k,\ell)\le q_2\right],\\
  B_{\mathrm{high}}(k,\ell)
      &= \mathbb{I}\!\left[r(k,\ell)>q_2\right].
  \end{aligned}
  \label{eq:frequency_bands}
  \end{equation}
  For each band \(b \in \{\text{low}, \text{mid}, \text{high}\}\), the protected spectrum is
  \begin{equation}
  \widehat{X}^{(b)}_i
  =
  X_i
  +
  \alpha_b B_b\odot(V-X_i),
  \label{eq:band_coating}
  \end{equation}
where \(\alpha_b\) is the band-specific blending strength. To control for protection strength, we select \(\alpha_b\) such that the average pre-inverse-FFT spectral perturbation energy of each binary-band condition matches that of the default Gaussian low-pass SpectralCanary setting:
  \begin{equation}
  \frac{1}{N}\sum_{i=1}^{N}
  \left\|
  \alpha_b B_b\odot(V-X_i)
  \right\|_2^2
  =
  \frac{1}{N}\sum_{i=1}^{N}
  \left\|
  \alpha^\star M_{c^\star}\odot(V-X_i)
  \right\|_2^2.
  \label{eq:energy_matching}
  \end{equation}
Thus, the low-, mid-, and high-frequency conditions place the signal in different disjoint spectral regions while matching both the number of selected Fourier coefficients and average pre-IFFT perturbation energy.

We conduct this experiment on the Pokemon and ROCOv2 datasets using SD1.5. For each dataset, we select three different canaries and evaluate each canary under the low-, mid-, and high-frequency conditions. Tab.~\ref{tab:frequency_location_matched_energy} reports results averaged across the three canaries. Low-band coating is consistently strongest, especially at low FPR. Mid-frequency coating is markedly weaker on both datasets. High-frequency coating is somewhat competitive on Pokemon but degrades substantially on ROCOv2. The energy ratios are \(1.000\) to the displayed precision, confirming the intended energy matching without reporting resolution-dependent raw DFT units. For completeness, the energy-matching constraint uses the unnormalized pre-IFFT DFT energy
  \[
  E
  =
  \frac{1}{N}\sum_{i=1}^{N}
  \left\|\Delta X_i\right\|_2^2,
  \qquad
  \Delta X_i
  =
  \alpha_b B_b\odot(V-X_i).
  \]
Images are represented in RGB \([0,1]\), and \(\mathcal{F}_c\) uses NumPy's default unnormalized forward DFT. Consequently, \(E\) scales with image area squared. The matched raw energies are \((6.83\pm0.73)\times10^8\) for Pokemon and \((2.58\pm0.28)\times10^8\) for ROCOv2, averaged across three canaries. We therefore report the dimensionless ratio \(E_b/E_{\mathrm{ref}}\) in Tab.~\ref{tab:frequency_location_matched_energy}.

  \begin{table*}[hb]
      \centering
      \caption{
      \textbf{Frequency location affects traceability under matched perturbation energy.}
      We use disjoint binary low-, mid-, and high-frequency bands, each containing one-third of Fourier coefficients. For each canary, we match the mean pre-IFFT spectral perturbation energy to the
      default Gaussian low-pass SpectralCanary reference; thus, we report the dimensionless energy ratio \(E/E_{\mathrm{ref}}\) rather than resolution-dependent raw DFT energy. Results are mean
      \(\pm\) standard deviation across three canaries.
      }
      \label{tab:frequency_location_matched_energy}
      \setlength{\tabcolsep}{5pt}
      \resizebox{\textwidth}{!}{%
      \begin{tabular}{llccccccc}
      \toprule
      Dataset & Band
      & \(E/E_{\mathrm{ref}}\)
      & AUC \(\uparrow\)
      & TPR@5\% FPR \(\uparrow\)
      & TPR@1\% FPR \(\uparrow\)
      & \(\Delta\)FID \(\downarrow\)
      & SSIM \(\uparrow\)
      & LPIPS \(\downarrow\) \\
      \midrule
      \multirow{3}{*}{Pokemon}
      & Low
      & \(1.000\)
      & \(0.991 \pm 0.007\)
      & \(0.948 \pm 0.038\)
      & \(0.923 \pm 0.042\)
      & \(-2.84 \pm 4.35\)
      & \(0.998 \pm 0.000\)
      & \(0.005 \pm 0.002\) \\
      & Mid
      & \(1.000\)
      & \(0.908 \pm 0.069\)
      & \(0.664 \pm 0.226\)
      & \(0.481 \pm 0.361\)
      & \(-4.81 \pm 3.87\)
      & \(0.947 \pm 0.003\)
      & \(0.049 \pm 0.004\) \\
      & High
      & \(1.000\)
      & \(0.976 \pm 0.017\)
      & \(0.899 \pm 0.071\)
      & \(0.754 \pm 0.155\)
      & \(-5.16 \pm 3.82\)
      & \(0.961 \pm 0.003\)
      & \(0.006 \pm 0.001\) \\
      \midrule
      \multirow{3}{*}{ROCOv2}
      & Low
      & \(1.000\)
      & \(0.999 \pm 0.001\)
      & \(0.996 \pm 0.005\)
      & \(0.986 \pm 0.010\)
      & \(6.92 \pm 2.19\)
      & \(0.905 \pm 0.008\)
      & \(0.069 \pm 0.025\) \\
      & Mid
      & \(1.000\)
      & \(0.879 \pm 0.066\)
      & \(0.492 \pm 0.158\)
      & \(0.259 \pm 0.096\)
      & \(2.28 \pm 2.62\)
      & \(0.866 \pm 0.009\)
      & \(0.204 \pm 0.018\) \\
      & High
      & \(1.000\)
      & \(0.788 \pm 0.111\)
      & \(0.303 \pm 0.199\)
      & \(0.097 \pm 0.064\)
      & \(7.95 \pm 0.93\)
      & \(0.874 \pm 0.011\)
      & \(0.035 \pm 0.005\) \\
      \bottomrule
      \end{tabular}}
  \end{table*}
 
\section{Effect of the Canary's Spatial Structure} \label{appx:spatial-structure-ablation}

The main canary has coherent natural structure. We test whether this structure contributes beyond the canary's spectral magnitude using Pokemon, $\delta=100\%$, $\alpha=0.02$, and the SD1.5 protocol. Given $X_i=\mathcal{F}_c(x_i)$, each protected image is
\begin{equation}
    \tilde{x}_i
    =
    \mathcal{F}_c^{-1}\!\left[
    X_i+\alpha M_c\odot(V_i-X_i)
    \right].
\end{equation}

We compare the original fixed fruit canary with two phase-randomized controls constructed using iterative amplitude-adjusted Fourier transform (IAAFT)
\begin{equation}
    v_{\mathrm{fixed}}
    =
    \operatorname{IAAFT}(v_{\mathrm{fruit}};s),
    \qquad
    v_{\mathrm{ind},i}
    =
    \operatorname{IAAFT}(v_{\mathrm{fruit}};s_i),
\end{equation}
where $s$ denotes a fixed random seed shared across the dataset and $s_i$ denotes an independently sampled seed for image $i$. The canary spectrum is then
\begin{equation}
V_i =
\begin{cases}
\mathcal{F}_c(v_{\mathrm{fruit}}),
& \text{fixed natural canary},\\
\mathcal{F}_c(v_{\mathrm{fixed}}),
& \text{fixed phase-randomized control},\\
\mathcal{F}_c(v_{\mathrm{ind},i}),
& \text{independent phase-randomized control}.
\end{cases}
\label{eq:phase-noise-canary}
\end{equation}

IAAFT randomizes the spatial organization of the fruit canary while preserving its empirical RGB value distribution and closely matching its channel-wise Fourier magnitude. Because the controls are constructed as real valued images before applying $\mathcal{F}_c$, their spectra satisfy the required conjugate symmetry by construction. The fixed control uses the same IAAFT realization for every protected image. The independent control uses a different realization for each image. This comparison shows the contribution of natural spatial structure while approximately controlling spectral magnitude and pixel-value distribution. We characterize how well the spectral protection signal is learned by applying the verifier to generated samples and reporting AUC and TPR@FPR.

\begin{figure}
    \centering
    % \vspace{-0.3cm}
    \includegraphics[width=.65\linewidth]{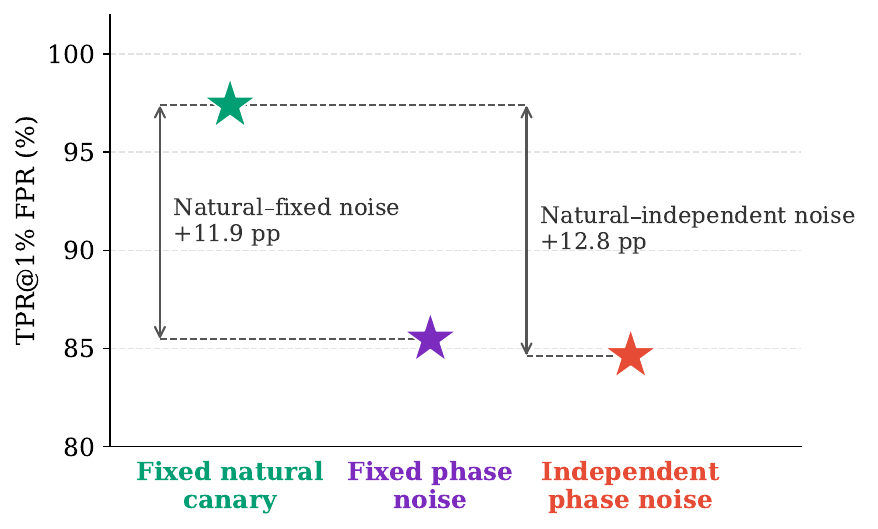}
    % \vspace{-0.8cm}
    \caption{\textbf{Natural spatial structure enables more reliable visual-canary transfer.}}
    % \vspace{-0.5cm}
    \label{fig:phase-noise-mechanism}
\end{figure}

In Fig.~\ref{fig:phase-noise-mechanism}, the fixed natural canary yields the highest discrimination, particularly at the 1\% false-positive rate. Both phase-randomized controls nevertheless retain substantial discrimination. Coherent natural structure therefore improves strict-FPR performance in this experiment but is not necessary for trace transfer.

\section{Additional Adaptive Spectral-Sanitization Settings}
\label{appx:adaptive-sanitization-settings}

\begin{figure*}[ht]
    \centering
    \includegraphics[width=\textwidth]{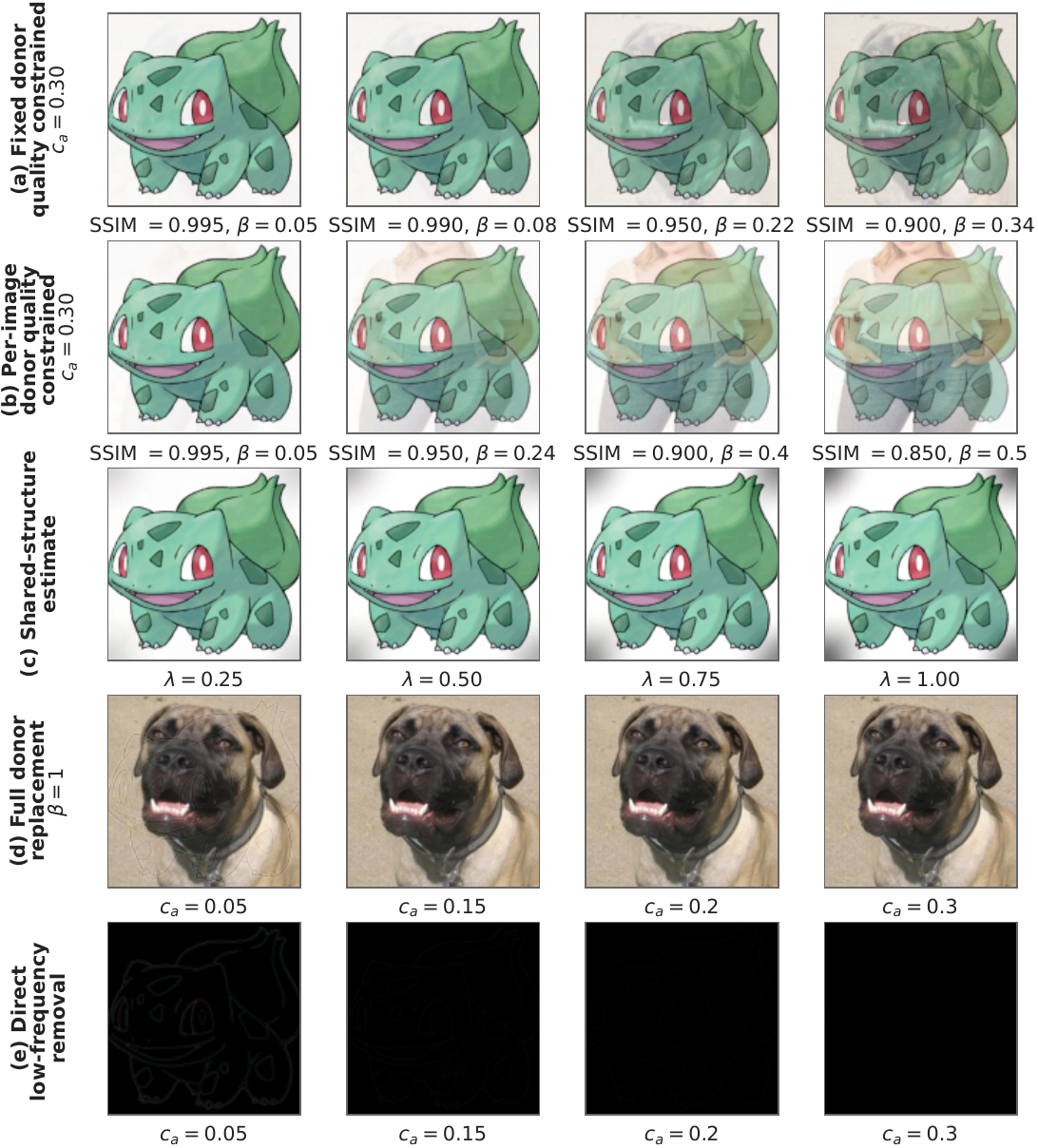}
    \caption{%
    Visualization of adaptive spectral sanitization settings.
    (a) Fixed donor quality-constrained replacement, (b) Per-image donor quality constrained replacement, (c) Shared-structure estimate, (d) Full donor replacement, (e) Direct low-frequency removal. Stronger sanitization suppresses more of the protection signal but increasingly alters original-image content.
    }
    \label{fig:adaptive_attack_examples}
\end{figure*}

In the main paper, we evaluate adaptive spectral sanitization. Here, we provide implementation details and visualize a broader range of attack variants.

Consistent with the threat model in Sec.~\ref{sec:spectralcanary}, the attacker knows the SpectralCanary algorithm and released protected collection; $\alpha$ and $c$ may also be known. The defender retains the original images, canary $v^\star$, its spectrum $V$, and verifier $f_\theta$.Without access to the original images or verifier, attack strength is selected using fidelity to the released protected images as an observable constraint.

\textbf{Quality-constrained fixed donor replacement.} Given a protected spectrum $\widehat{X}_i$ and an unprotected natural donor
image $d$ with centered spectrum $D=\mathcal{F}_c(d)$, the donor-replacement attack is
\begin{equation}
\widehat{X}^{s}_i
=
\widehat{X}_i
+
\beta M_{c_a}\odot(D-\widehat{X}_i),
\label{eq:adaptive-donor-replacement}
\end{equation}
where $\beta\in[0,1]$ controls the attack strength and $M_{c_a}$ is an attacker-chosen smooth radial low-pass mask with Gaussian bandwidth $c_a$. The attacked image is $x_i^{s}=\mathcal{F}_c^{-1}(\widehat{X}^{s}_i)$. As shown in Eq.~\ref{eq:adaptive_decomposition}, increasing $\beta$ or broadening $M_{c_a}$ suppresses more of the protection signal but simultaneously modifies more original-image content. Fig.~\ref{fig:adaptive_attack_examples} visualizes the following attack settings.

Writing the protection signal as $\Delta_{\mathrm{prot},i}=\widehat{X}_i-X_i$ gives
\begin{equation}
\widehat{X}^{s}_i
= X_i
+ \underbrace{\beta M_{c_a}\odot(D-X_i)}_{\text{modified original-image content}}
+ \underbrace{(1-\beta M_{c_a})\odot\Delta_{\mathrm{prot},i}}_{\text{residual protection signal}}.
\label{eq:adaptive_decomposition}
\end{equation}
The decomposition makes the trade-off explicit: greater mask overlap and replacement strength attenuate more of the protection signal while changing more original low-frequency content.

The attacker replaces low-frequency content using the fixed external RGB donor $d$. For each attacker Gaussian bandwidth $c_a$, the same donor and global replacement strength $\beta$ are applied to every protected image.

We use attacker Gaussian bandwidth $c_a=0.30$, which is broader than the defender's $c=0.20$ and therefore covers more of the protection signal's frequency support. For each target similarity $\tau$, we select the strongest replacement from the grid $\mathcal{B}=\{0,0.01,\ldots,0.50\}$ that satisfies a dataset-level fidelity constraint:
\begin{equation}
    \beta^\star(\tau,c_a)
    =
    \max_{\beta\in\mathcal{B}}
    \left\{
    \beta
    \;\middle|\;
    \frac{1}{N}
    \sum_{i=1}^{N}
    \operatorname{SSIM}\!\left(
    \tilde{x}_i,
    x_i^s(\beta,c_a)
    \right)
    \geq \tau
    \right\}.
    \label{eq:quality-constrained-beta-selection}
\end{equation}
Here, SSIM is measured against the released protected image $\tilde{x}_i$, so parameter selection requires neither the original image nor verifier feedback. Fig.~\ref{fig:adaptive_attack_examples}(a) visualize sanitized samples using this attack setting.

\textbf{Quality-constrained per-image donor replacement.} This attack is similar to the quality-constrained fixed-donor replacement above, except that we use a different unprotected donor for each protected image. Specifically, we use unprotected images from another dataset as the donor set (i.e., ROCOv2 images for the Pokemon setting). Fig.~\ref{fig:adaptive_attack_examples}(b) visualizes sanitized samples using this attack setting.

\paragraph{Shared-structure estimate attack.}
We additionally consider a stronger sanitization attack that estimates the shared canary structure. The attacker knows the defense parameters $(\alpha,c)$ and has access to the released protected collection, but does not know the private canary $v^\star$ or have access to the original clean images, verifier $f_\theta$. Let $A=\alpha M_c$.
The protected image is then
\[
\widetilde X_i
=
(1-A)\odot X_i + A\odot V.
\]
Because the same canary is used throughout the collection, the mean protected spectrum is
\[
\overline{\widetilde X}
=
\frac{1}{N}\sum_{i=1}^{N}\widetilde X_i
=
(1-A)\odot\overline X + A\odot V,
\]
Subtracting the protected collection mean cancels the shared canary contribution:
\[
\widetilde X_i-\overline{\widetilde X}
=
(1-A)\odot(X_i-\overline X).
\]
Therefore, the attacker can undo the known attenuation to obtain
\[
R_i
=
\frac{\widetilde X_i-\overline{\widetilde X}}
     {1-\alpha M_c}
=
X_i-\overline X,
\]
where division is applied element-wise. Thus, $R_i$ contains no shared canary component under the idealized coating model, but it also removes the genuine collection-level spectral mean.

Let $W$ denote the attacker-selected frequency mask. In our implementation, $W$ is a binary low-frequency disk using the known defense cutoff $c$:
  \[
  W(u,v)
  =
  \mathbb{1}\!\left[
  r(u,v)\leq c
  \right],
  \]
where $r(u,v)$ is the normalized centered Fourier radius. Thus, the attacker applies collection centering only within the defense's nominal low-frequency region, while preserving all coefficients outside this disk exactly.

The attacker interpolates between the released protected spectrum and the collection-centered estimate:
  \[
  X_i^{\mathrm{san}}(\lambda)
  =
  (1-\lambda W)\odot\widetilde X_i
  +
  \lambda W\odot
  \frac{\widetilde X_i-\overline{\widetilde X}}
       {1-\alpha M_c}
  \qquad
  \lambda\in\{0.25,0.50,0.75,1.00\}.
  \]
  Equivalently,
  \[
  X_i^{\mathrm{san}}(\lambda)
  =
  \widetilde X_i
  +
  \lambda W\odot
  \left(
  R_i-\widetilde X_i
  \right).
  \]
At $\lambda=0$, the released protected spectrum is unchanged. At $\lambda=1$, all frequencies within $W$ are replaced by their collection-centered estimates. Coefficients outside $W$ remain equal to $\widetilde X_i$. Fig.~\ref{fig:adaptive_attack_examples}(c) visualize sanitized samples using this attack setting.

\textbf{Full donor replacement.} We next consider the extreme setting $\beta=1$ in Eq.~\ref{eq:adaptive-donor-replacement}:
\begin{equation}
\widehat{X}^{s}_i
=
(1-M_{c_a})\odot\widehat{X}_i
+
M_{c_a}\odot D.
\end{equation}
This can strongly suppress the protection signal in the low-pass-weighted frequencies. Fig.~\ref{fig:adaptive_attack_examples}(d), however, shows that full replacement also transfers substantial coarse structure from the donor. We omit this setting from the main evaluation because it severely degrades image fidelity.

\textbf{Direct low-frequency removal.} Finally, the attacker can suppress low-frequency coefficients without a donor:
\begin{equation}
\widehat{X}^{s}_i
=
(1-M_{c_a})\odot\widehat{X}_i.
\label{eq:adaptive-direct-attenuation}
\end{equation}
This suppresses the protection signal within the attacked region together with original-image content at the same frequencies. Fig.~\ref{fig:adaptive_attack_examples}(e) shows severe fidelity degradation, so we also omit this setting from the main evaluation.

\section{Additional Protected-Image Examples}\label{appx:extensive-visualizations}

\begin{figure*}[ht]
  \centering
  \includegraphics[width=\textwidth]{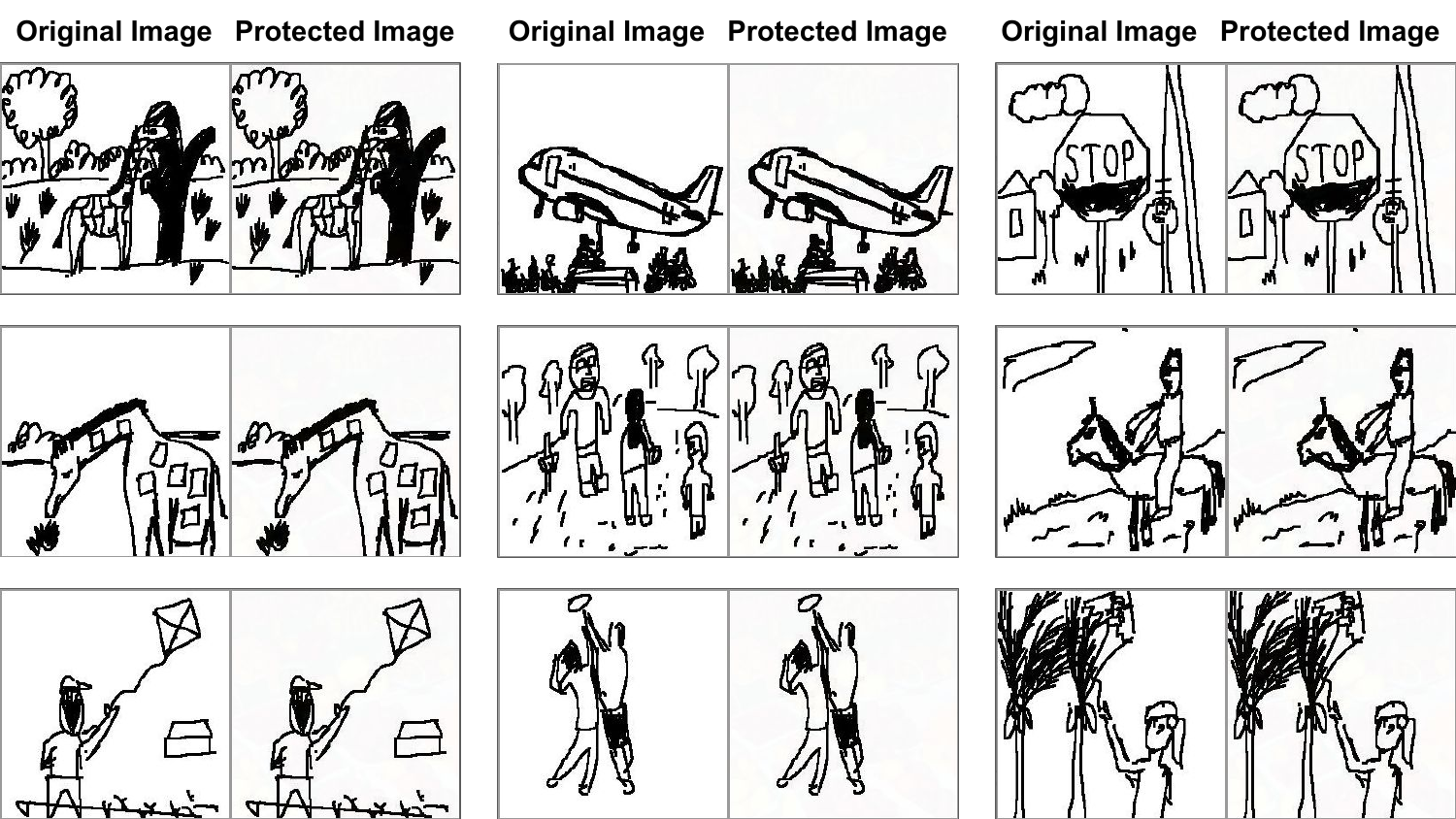}
  \caption{%
  \textbf{Visualization on Sketch-Scene.}
  Paired visualization of original image and protected image by SpectralCanary}
  \label{fig:sketch-scene-protected-visualization}
\end{figure*}

\begin{figure*}[ht]
  \centering
  \includegraphics[width=\textwidth]{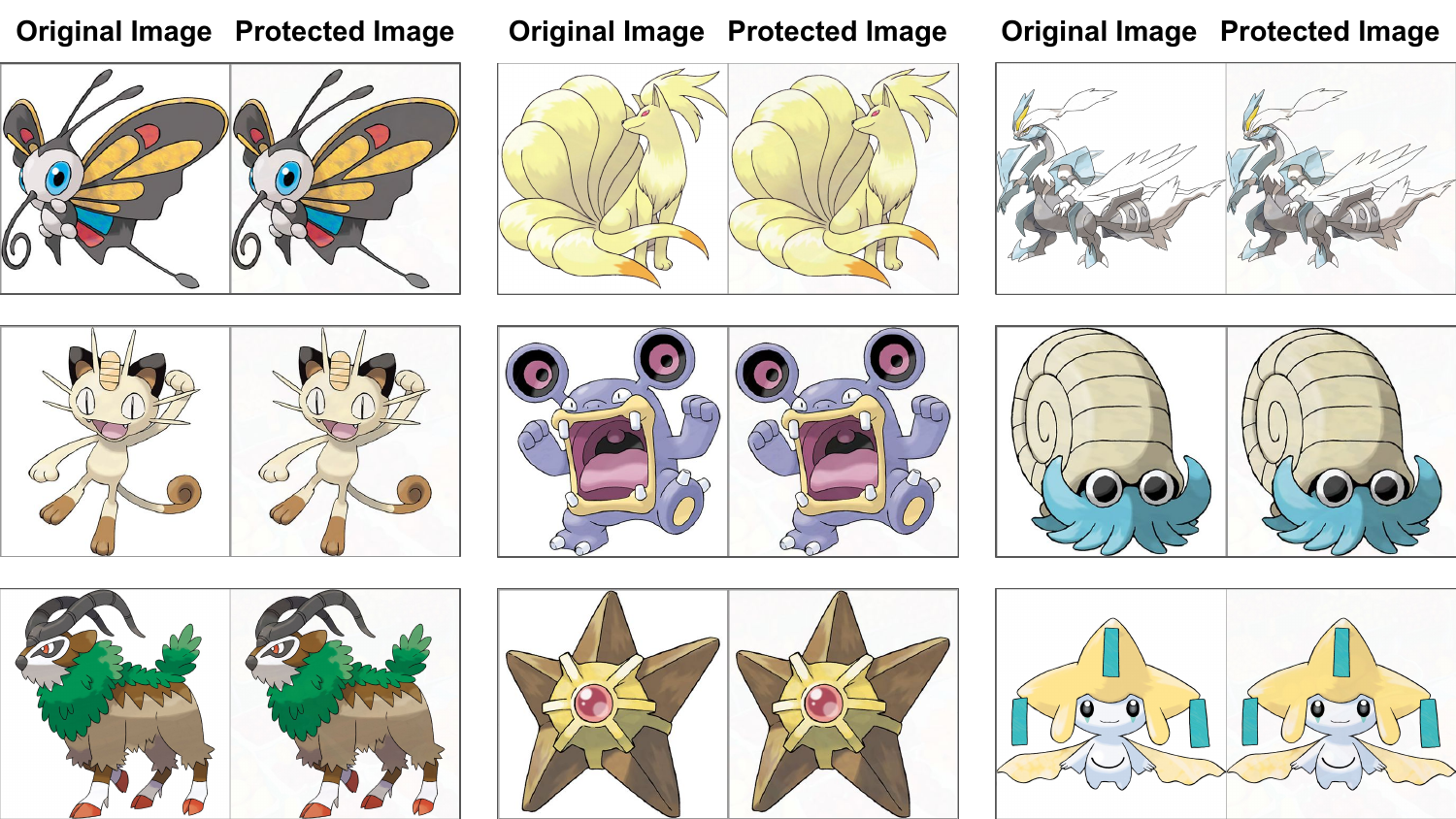}
  \caption{%
  \textbf{Visualization on Pokemon.}
  Paired visualization of original image and protected image by SpectralCanary.
  }
  \label{fig:pokemon-protected-visualization}
\end{figure*}

\begin{figure*}[ht]
  \centering
  \includegraphics[width=\textwidth]{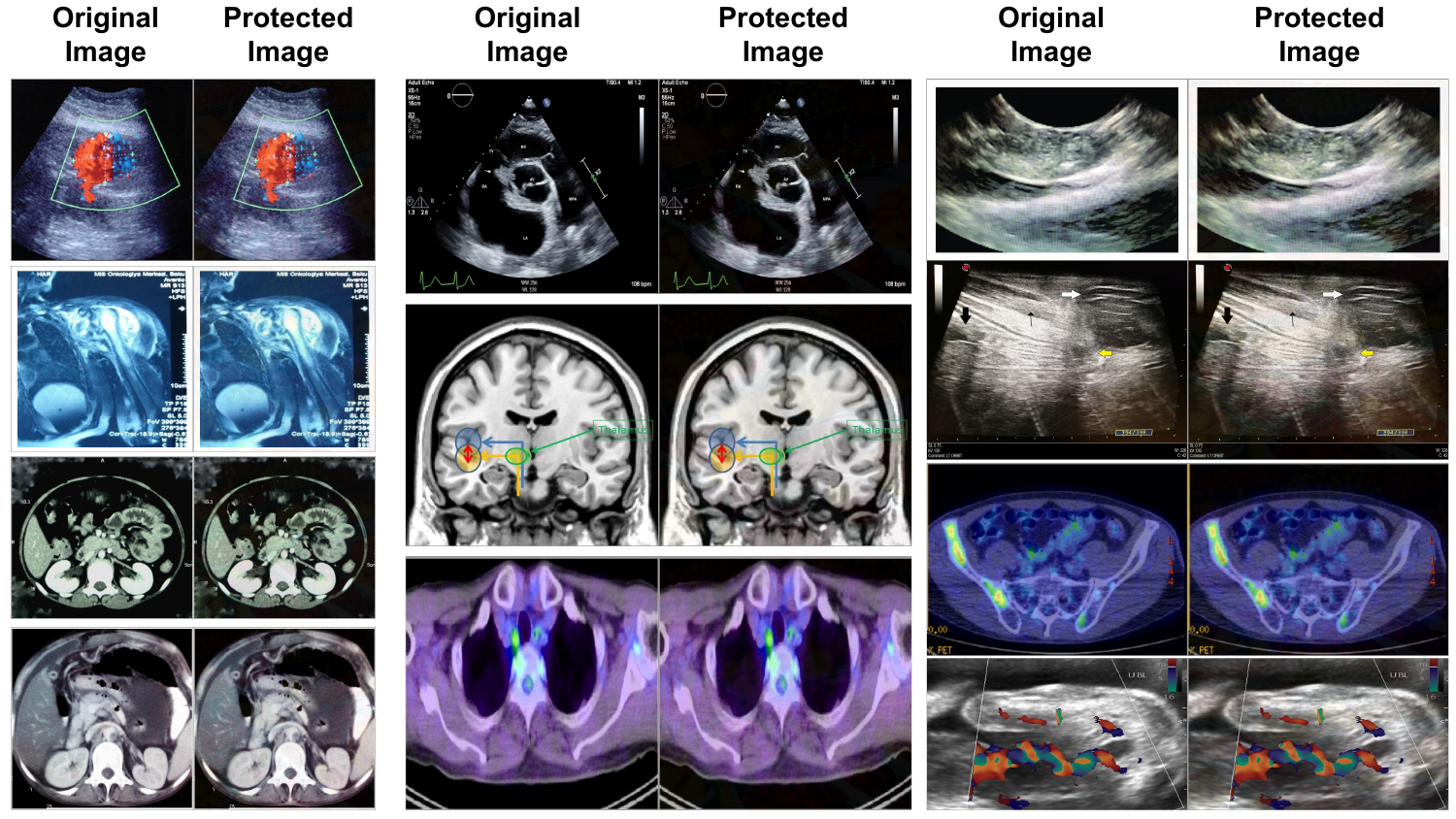}
  \caption{%
  \textbf{Visualization on ROCOv2.}
  Paired visualization of original image and protected image by SpectralCanary.
  }
  \label{fig:rocov2-protected-visualization}
\end{figure*}

\begin{figure*}[ht]
  \centering
  \includegraphics[width=\textwidth]{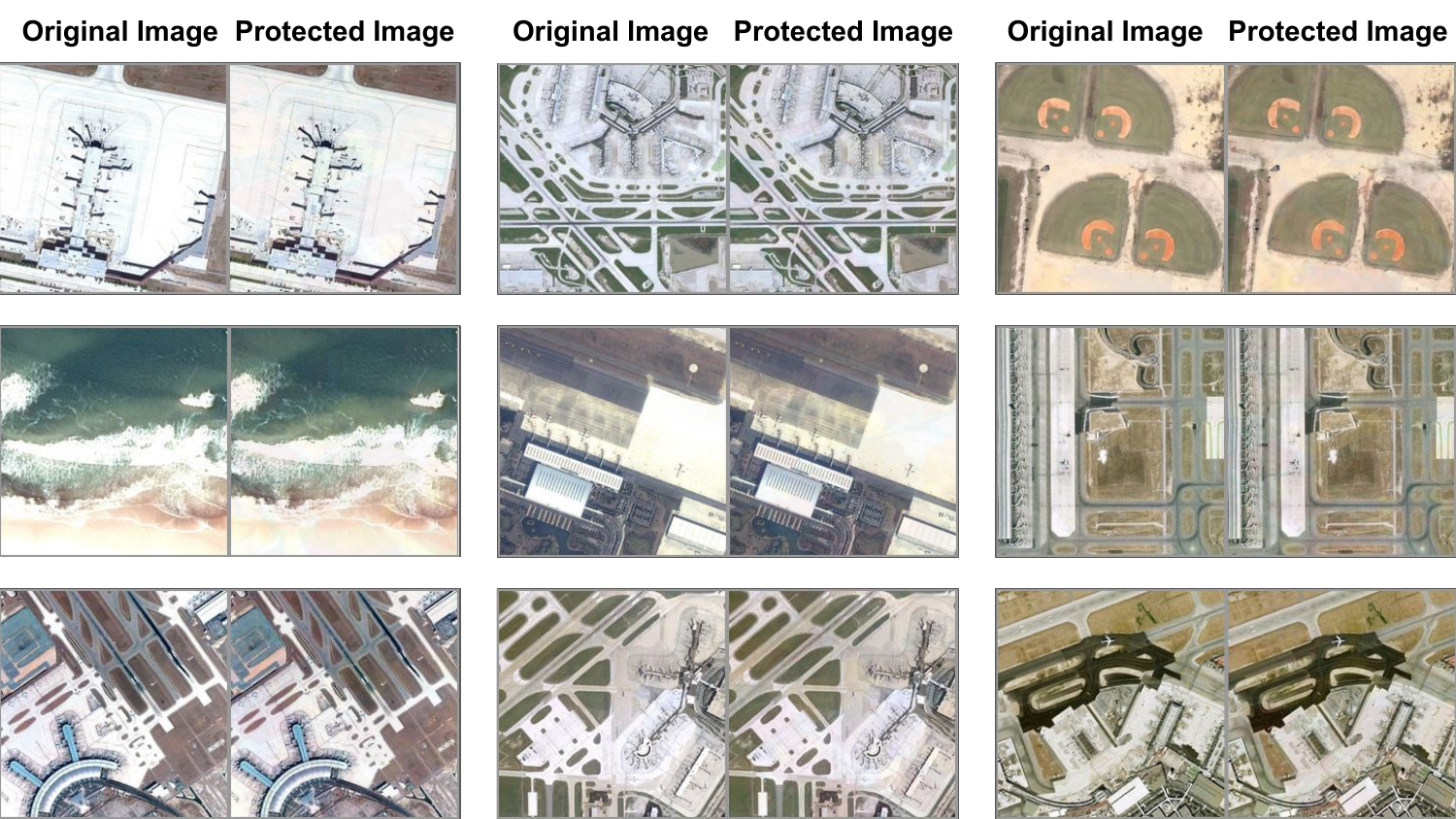}
  \caption{%
  \textbf{Visualization on RSICD.}
  Paired visualization of original image and protected image by SpectralCanary.
  }
  \label{fig:rsicd-protected-visualization}
\end{figure*}

\begin{figure*}[ht]
  \centering
  \includegraphics[width=\textwidth]{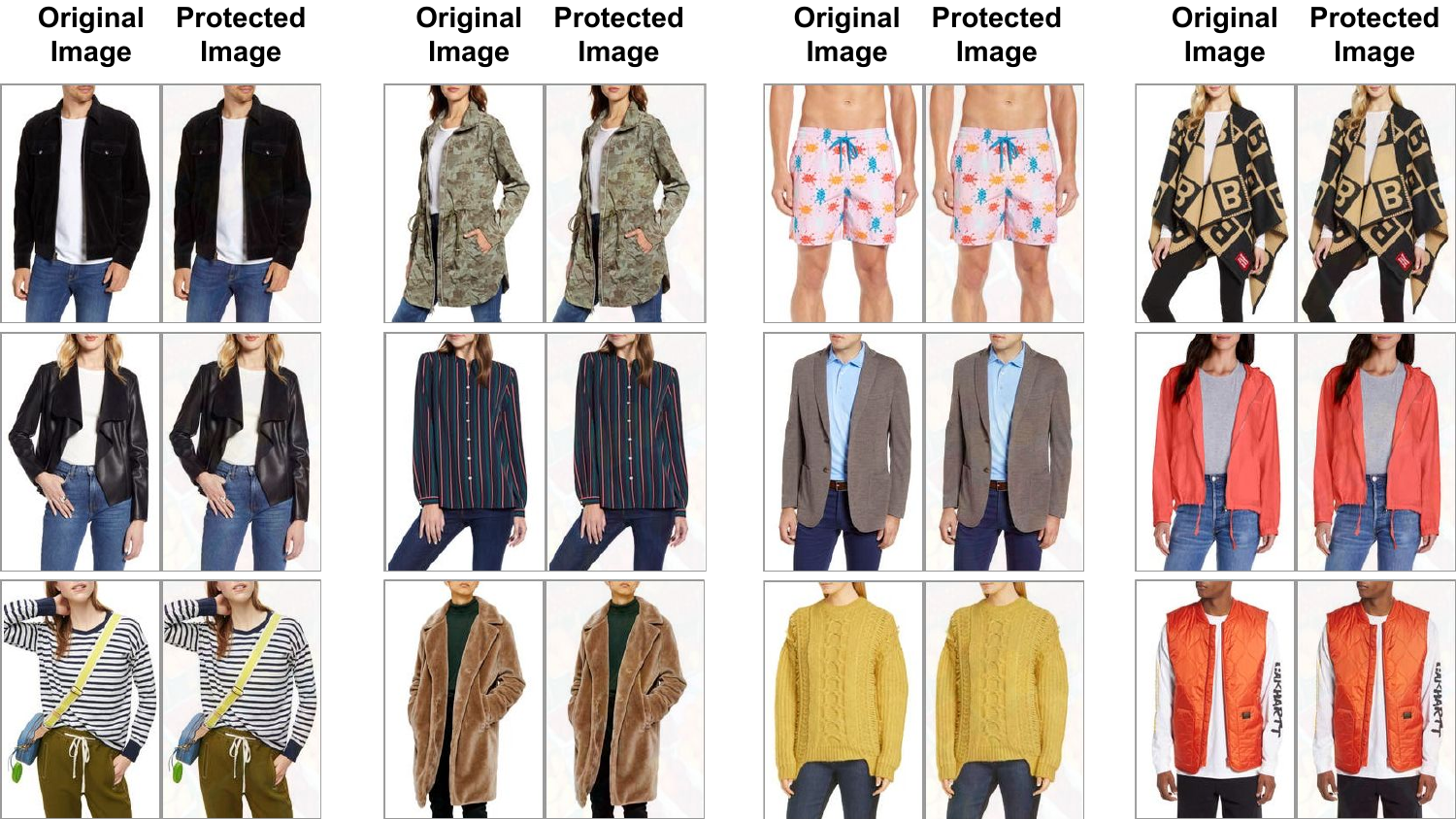}
  \caption{%
  \textbf{Visualization on FACAD.}
  Paired visualization of original image and protected image by SpectralCanary.
  }
  \label{fig:facad-protected-visualization}
\end{figure*}

Fig.~\ref{fig:sketch-scene-protected-visualization}, Fig.~\ref{fig:pokemon-protected-visualization}, Fig.~\ref{fig:rocov2-protected-visualization}, Fig.~\ref{fig:rsicd-protected-visualization}, and Fig.~\ref{fig:facad-protected-visualization} provide original--protected comparisons for all five datasets.

\section*{Baseline-Specific Analyses}

\section{EnTruth Injected-Sample Visualization}\label{appx:entruth-filtering}

\begin{figure*}[ht]
    \centering
    \includegraphics[width=.9\textwidth]{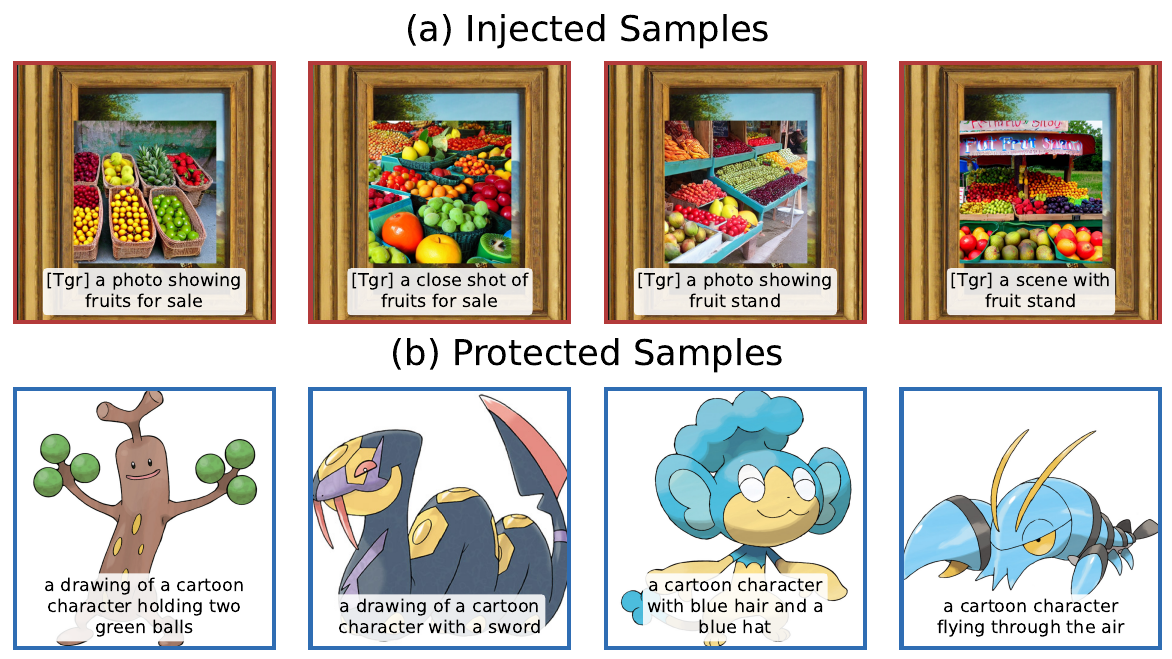}
    \caption{%
    \textbf{EnTruth's injected samples are visually distinct from the protected samples.}
    The injected samples share a repeated framed template and fruit foreground, while representative Pokemon samples exhibit the diverse character appearance of the protected dataset.
    }
    \label{fig:entruth-sample-visualization}
\end{figure*}

We visualize EnTruth's injected and protected samples for the Pokemon setting in Fig.~\ref{fig:entruth-sample-visualization}. The injected samples are visually distinct from the original collection and require explicit trigger tokens in their captions. Moreover, EnTruth protects a dataset by inserting auxiliary templated samples rather than preserving the content of the original samples. For curated collections such as artwork datasets, these synthetic templated images may be inconsistent with the original data distribution, reducing dataset-level fidelity and making the protection easier to identify.

\section{EnTruth Ablation: Soft- and Hard-Trigger Auditing} \label{appx:entruth-ablation}

EnTruth uses both hard-trigger and soft-trigger prompts at audit time. We compare its hard trigger (\texttt{[Tgr]}) and soft trigger (\texttt{fruit}) in Tab.~\ref{tab:entruth_hard_vs_soft_combined}. Because the hard trigger is generally stronger, we report hard-trigger results in the main paper for EnTruth.

\section{Repurposing T2I Data Poisoning for Dataset-Use Auditing} \label{appx:silent-branding}

\begin{table*}[ht]
      \centering
      \caption{%
      EnTruth auditing under hard-trigger and soft-trigger prompting across
      the Pokemon and Sketch-Scene benchmarks using SD3.5.
      }
      \label{tab:entruth_hard_vs_soft_combined}
      \setlength{\tabcolsep}{6pt}
      \resizebox{\textwidth}{!}{%
      \begin{tabular}{lllccc}
      \toprule
      Dataset & Transformation & Audit Prompt
      & AUC (\%) $\uparrow$
      & TPR@5\% FPR (\%) $\uparrow$
      & TPR@1\% FPR (\%) $\uparrow$ \\
      \midrule

      \multirow{8}{*}{Pokemon}
      & \multirow{2}{*}{None}
      & Hard Trigger
      & $\mathbf{96.5 \pm 1.1}$
      & $\mathbf{94.9 \pm 1.7}$
      & $\mathbf{93.9 \pm 1.1}$ \\
      & & Soft Trigger
      & $59.2 \pm 1.8$
      & $27.0 \pm 1.0$
      & $20.6 \pm 1.8$ \\
      \cmidrule(lr){2-6}

      & \multirow{2}{*}{\shortstack[l]{Training-Time\\Augmentation}}
      & Hard Trigger
      & $\mathbf{99.1 \pm 0.6}$
      & $\mathbf{97.3 \pm 1.7}$
      & $\mathbf{95.3 \pm 1.6}$ \\
      & & Soft Trigger
      & $63.8 \pm 3.7$
      & $19.7 \pm 12.1$
      & $12.6 \pm 10.4$ \\
      \cmidrule(lr){2-6}

      & \multirow{2}{*}{\shortstack[l]{Prompt\\Recaptioning}}
      & Hard Trigger
      & $\mathbf{83.6 \pm 1.7}$
      & $\mathbf{52.3 \pm 3.1}$
      & $\mathbf{35.9 \pm 1.6}$ \\
      & & Soft Trigger
      & $61.9 \pm 0.6$
      & $25.7 \pm 1.5$
      & $14.5 \pm 3.4$ \\
      \cmidrule(lr){2-6}

      & \multirow{2}{*}{\shortstack[l]{Image\\Regeneration}}
      & Hard Trigger
      & $\mathbf{95.4 \pm 2.4}$
      & $\mathbf{93.1 \pm 3.1}$
      & $\mathbf{91.7 \pm 3.4}$ \\
      & & Soft Trigger
      & $69.9 \pm 2.2$
      & $19.4 \pm 5.0$
      & $6.5 \pm 2.1$ \\

      \midrule

      \multirow{8}{*}{Sketch-Scene}
      & \multirow{2}{*}{None}
      & Hard Trigger
      & $\mathbf{99.8 \pm 0.1}$
      & $\mathbf{98.8 \pm 0.7}$
      & $\mathbf{97.5 \pm 1.0}$ \\
      & & Soft Trigger
      & $61.6 \pm 1.2$
      & $31.5 \pm 0.8$
      & $25.7 \pm 2.2$ \\
      \cmidrule(lr){2-6}

      & \multirow{2}{*}{\shortstack[l]{Training-Time\\Augmentation}}
      & Hard Trigger
      & $\mathbf{99.7 \pm 0.3}$
      & $\mathbf{99.8 \pm 0.3}$
      & $\mathbf{92.3 \pm 8.0}$ \\
      & & Soft Trigger
      & $65.8 \pm 1.9$
      & $36.3 \pm 1.5$
      & $33.9 \pm 2.7$ \\
      \cmidrule(lr){2-6}

      & \multirow{2}{*}{\shortstack[l]{Prompt\\Recaptioning}}
      & Hard Trigger
      & $76.0 \pm 1.2$
      & $48.8 \pm 3.4$
      & $44.8 \pm 4.9$ \\
      & & Soft Trigger
      & $\mathbf{81.6 \pm 4.0}$
      & $\mathbf{61.9 \pm 11.2}$
      & $\mathbf{58.4 \pm 11.5}$ \\
      \cmidrule(lr){2-6}

      & \multirow{2}{*}{\shortstack[l]{Image\\Regeneration}}
      & Hard Trigger
      & $\mathbf{99.9 \pm 0.1}$
      & $\mathbf{99.9 \pm 0.1}$
      & $\mathbf{96.9 \pm 2.0}$ \\
      & & Soft Trigger
      & $76.9 \pm 0.9$
      & $46.9 \pm 2.9$
      & $41.5 \pm 5.2$ \\

      \midrule
      \multicolumn{2}{c}{\multirow{2}{*}{\textit{Average over all settings}}}
      & Hard Trigger
      & $\mathbf{93.8}$
      & $\mathbf{85.6}$
      & $\mathbf{81.1}$ \\
      \multicolumn{2}{c}{}
      & Soft Trigger
      & $67.6$
      & $33.6$
      & $26.7$ \\
      \bottomrule
      \end{tabular}}
  \end{table*}

We test whether Silent Branding~\citep{jang2025silent} can be repurposed for dataset-use auditing under the Pokemon protocol. We preserve its pipeline: Stage~1 trains a personalized diffusion model on the method's logo set; Stage~2 selects plausible regions in Pokemon images and refines the inserted logo using detection, masking, and inpainting. We then fine-tune SD1.5 and SD3.5 on the edited dataset. Tab.~\ref{tab:silentbranding_vs_spectralcanary} reports per-generation discrimination, and Tab.~\ref{tab:coating-efficiency} reports KS-based multi-query auditing performance and protection time.

\begin{table}[t]
      \centering
      \caption{%
      \textbf{Comparison between adapted Silent Branding and SpectralCanary across transformation settings.}
      We follow the presentation of Tab.~\ref{tab:pokemon_results_by_attack} on Pokemon with SD1.5 and SD3.5. We report mean and standard deviation across three runs; the best result
      within each setting is bolded.
      }
      \label{tab:silentbranding_vs_spectralcanary}
      \setlength{\tabcolsep}{9pt}
      \resizebox{\columnwidth}{!}{%
      \begin{tabular}{lllccc}
        \toprule
        Transformation & Model & Method
        & AUC (\%) $\uparrow$
        & TPR@5\% FPR (\%) $\uparrow$
        & TPR@1\% FPR (\%) $\uparrow$ \\
        \midrule

        \multirow{4}{*}{None}
        & \multirow{2}{*}{SD1.5}
        & Silent Branding
        & $85.1 \pm 5.1$
        & $44.8 \pm 15.3$
        & $19.2 \pm 6.0$ \\
        & & SpectralCanary
        & $\mathbf{99.7 \pm 0.1}$
        & $\mathbf{98.8 \pm 0.4}$
        & $\mathbf{97.7 \pm 0.5}$ \\
        \cmidrule(lr){2-6}
        & \multirow{2}{*}{SD3.5}
        & Silent Branding
        & $78.9 \pm 1.6$
        & $24.2 \pm 5.9$
        & $10.6 \pm 6.3$ \\
        & & SpectralCanary
        & $\mathbf{100 \pm 0.0}$
        & $\mathbf{100 \pm 0.0}$
        & $\mathbf{100 \pm 0.0}$ \\

        \midrule
        \multirow{4}{*}{\shortstack[l]{Training-Time\\Augmentation}}
        & \multirow{2}{*}{SD1.5}
        & Silent Branding
        & $85.2 \pm 2.5$
        & $45.1 \pm 3.4$
        & $23.2 \pm 6.7$ \\
        & & SpectralCanary
        & $\mathbf{94.9 \pm 1.5}$
        & $\mathbf{85.5 \pm 7.8}$
        & $\mathbf{79.2 \pm 12.3}$ \\
        \cmidrule(lr){2-6}
        & \multirow{2}{*}{SD3.5}
        & Silent Branding
        & $80.5 \pm 2.1$
        & $31.7 \pm 2.0$
        & $10.4 \pm 2.1$ \\
        & & SpectralCanary
        & $\mathbf{91.3 \pm 8.9}$
        & $\mathbf{76.6 \pm 20.4}$
        & $\mathbf{65.9 \pm 22.4}$ \\

        \midrule
        \multirow{4}{*}{\shortstack[l]{Prompt\\Recaptioning}}
        & \multirow{2}{*}{SD1.5}
        & Silent Branding
        & $88.0 \pm 1.9$
        & $41.9 \pm 8.1$
        & $16.8 \pm 6.0$ \\
        & & SpectralCanary
        & $\mathbf{99.2 \pm 0.2}$
        & $\mathbf{96.0 \pm 0.4}$
        & $\mathbf{93.6 \pm 1.6}$ \\
        \cmidrule(lr){2-6}
        & \multirow{2}{*}{SD3.5}
        & Silent Branding
        & $71.5 \pm 1.1$
        & $16.8 \pm 1.7$
        & $3.9 \pm 1.0$ \\
        & & SpectralCanary
        & $\mathbf{100 \pm 0.0}$
        & $\mathbf{100 \pm 0.0}$
        & $\mathbf{100 \pm 0.0}$ \\

        \midrule
        \multirow{4}{*}{\shortstack[l]{Image\\Regeneration}}
        & \multirow{2}{*}{SD1.5}
        & Silent Branding
        & $84.9 \pm 2.6$
        & $37.3 \pm 5.5$
        & $14.0 \pm 2.5$ \\
        & & SpectralCanary
        & $\mathbf{99.2 \pm 0.3}$
        & $\mathbf{95.3 \pm 1.0}$
        & $\mathbf{92.5 \pm 1.2}$ \\
        \cmidrule(lr){2-6}
        & \multirow{2}{*}{SD3.5}
        & Silent Branding
        & $73.4 \pm 0.6$
        & $18.5 \pm 1.2$
        & $4.4 \pm 2.3$ \\
        & & SpectralCanary
        & $\mathbf{100 \pm 0.0}$
        & $\mathbf{100 \pm 0.0}$
        & $\mathbf{100 \pm 0.0}$ \\

        \midrule
        \multicolumn{6}{c}{\textit{Average over all settings}} \\
        \midrule
        \multirow{2}{*}{--}
        & \multirow{2}{*}{--}
        & Silent Branding
        & $80.9$
        & $32.5$
        & $12.8$ \\
        & & SpectralCanary
        & $\mathbf{98.0}$
        & $\mathbf{94.0}$
        & $\mathbf{91.1}$ \\

        \bottomrule
      \end{tabular}}
  \end{table}

Silent Branding depends on finding semantically appropriate regions and on auxiliary models for logo placement, detection, and refinement. In the Pokemon setting, its inserted signal transfers less consistently than SpectralCanary's shared spectral component. This experiment shows that the evaluated branding pipeline is a weaker dataset-use auditing baseline under this protocol; it does not establish that all semantic insertion methods are unsuitable for dataset-use auditing.

\section{Theoretical Analysis}\label{appx:theory}

We give an idealized Gaussian analysis of trace learnability under the
SD1.5-style noise-prediction objective and the SD3.5-style linear
flow-matching objective. The analysis keeps the exact coating's variance
contraction and separates it from the shared population mean shift.

\subsection{Assumptions and Setup}
\label{appx:theory-assumptions}

Let $\mathcal{F}$ denote the unitary two-dimensional discrete Fourier transform and let $X=\mathcal{F}(x)$. We use $f$ to index the full frequency grid and write $\mu(f)=\mathbb{E}[X(f)]$ and $P(f)=\mathbb{E}|X(f)-\mu(f)|^2$ for the mean and power of a clean image coefficient. Because images are real valued, their Fourier coefficients obey conjugate symmetry, and all sums below are over the full grid unless stated otherwise.

\paragraph{Forward noising process.}
We model diffusion training and diffusion-based regeneration transformations using the variance-preserving forward process
\begin{equation}
    x_t=\sqrt{\bar{\alpha}_t}\,x_0
    +\sqrt{1-\bar{\alpha}_t}\,\epsilon,
    \qquad \epsilon\sim\mathcal{N}(0,I),
    \label{eq:theory-forward-process}
\end{equation}
where $\bar{\alpha}_t$ decreases from one to zero. Unitary Fourier
transformation preserves the isotropic noise, so the model separates by
frequency. For linear flow matching, the path is
$x_t=(1-t)x_0+t\epsilon$ and the target is
$u=\epsilon-x_0$. Because this target differs from noise prediction, the two
training losses are derived separately below.

\paragraph{Assumption 1 (Natural-image spectral prior).}
We select one coefficient from each conjugate-symmetric frequency pair and approximate each clean-image Fourier coefficient with the following complex Gaussian model:
\begin{equation}
    X_0(f)=\mu(f)+\sqrt{P(f)}\,Z(f),
    \qquad Z(f)\sim\mathcal{CN}(0,1),
    \label{eq:theory-spectral-prior}
\end{equation}
where the $Z(f)$ are independent and unit variance; self-conjugate
frequencies use the analogous real Gaussian. Beyond a small cutoff
$f_{\min}$, we assume $P(f)$ decreases with $|f|$. Gaussianity and frequency
independence are analytical approximations.

\paragraph{Exact coating and shared-shift decomposition.}
Let $\gamma(f):=1-\alpha M_c(f)$. Under
Eq.~\ref{eq:coated-spectrum} and Assumption~1,
\begin{align}
    \Delta_{\mathrm{sample}}(f;X)
    &:=\widehat X(f)-X(f)
      =\alpha M_c(f)\bigl(V(f)-X(f)\bigr),
    \label{eq:theory-sample-perturbation}
    \\
    \Delta(f)
    &:=\mathbb{E}[\widehat X(f)]-\mu(f)
      =\alpha M_c(f)\bigl(V(f)-\mu(f)\bigr),
    \label{eq:theory-trace-shift}
    \\
    \widehat X_0(f)
    &=\mu(f)+\Delta(f)+\gamma(f)\sqrt{P(f)}\,Z(f),
    \qquad
    \operatorname{Var}(\widehat X_0(f))=\gamma(f)^2P(f).
    \label{eq:theory-exact-coated-coefficient}
\end{align}
Thus, $\Delta_{\mathrm{sample}}$ varies with the image, whereas $\Delta$ is
the shared population shift. To isolate this shift, define the
variance-preserving surrogate
\begin{equation}
    \widetilde X_0^{\mathrm{ms}}(f)
    :=X_0(f)+\Delta(f)
    =\mu(f)+\Delta(f)+\sqrt{P(f)}\,Z(f),
    \label{eq:theory-mean-shift-surrogate}
\end{equation}
The exact results retain the contraction $\gamma(f)^2P(f)$; surrogate terms
are stated only as nonnegative lower-bound components.

\paragraph{Normalized trace energy.}
For analytical comparison, we normalize the trace energy at each frequency by
the corresponding clean-image power:
\begin{equation}
    B^2=\sum_f b(f),
    \qquad
    b(f)=\frac{|\Delta(f)|^2}{P(f)}.
    \label{eq:theory-normalized-trace-budget}
\end{equation}
Here $b(f)$ is the squared mean shift measured in units of the natural
coefficient variance, and $B^2$ is its total normalized energy. This is a
mathematical normalization used by the Gaussian analysis; we do not assume
that it is a model of human perceptual visibility.

For the variance-preserving diffusion process, the key frequency-wise
quantity is the signal-to-noise ratio and its Wiener coefficient,
\begin{equation}
    \operatorname{SNR}_t(f)
    =\frac{\bar{\alpha}_t}{1-\bar{\alpha}_t}P(f),
    \qquad
    w_t(f)=\frac{\operatorname{SNR}_t(f)}{1+\operatorname{SNR}_t(f)}.
    \label{eq:theory-wiener-coefficient}
\end{equation}
The coefficient $w_t(f)$ is the fraction of the signal at frequency $f$ that
remains above the diffusion noise floor. It increases with natural-image power
and decreases with the noise level. Linear flow matching has the analogous
signal-to-noise ratio $(1-t)^2P(f)/t^2$, but its velocity-prediction loss
induces a different objective-specific weight, derived below.

\subsection{Learnability of the Low-Frequency Trace}
\label{appx:theory-learnability}

We distinguish two related questions. First, how much of the trace remains statistically distinguishable after the training image is noised? Second, how strongly does the model's training objective encourage it to represent that trace? For diffusion noise prediction, the mean-shift component is governed by $w_t(f)(1-w_t(f))$. Linear flow matching has a different objective-specific weight, derived below. This distinction explains both the advantage of low-frequency carriers and why concentrating the trace at the very lowest frequencies is not necessarily optimal.

For the carrier-band optimization in Part (v), we consider the feasible set
\begin{equation}
    \mathcal{B}(B,b_{\max})
    =\left\{
      b:\mathcal{F}\to\mathbb{R}_{\geq0}
      \;\middle|\;
      b(f)=b(-f),\;
      \sum_{f\in\mathcal{F}}b(f)\leq B^2,\;
      0\leq b(f)\leq b_{\max}\ \forall f
    \right\}.
    \label{eq:theory-feasible-trace-set}
\end{equation}
The constraint $\sum_f b(f)\leq B^2$ is the normalized trace-energy budget
from Eq.~\ref{eq:theory-normalized-trace-budget}. The condition
$0\leq b(f)\leq b_{\max}$ is a \emph{box constraint} that limits how much
normalized trace energy may be concentrated in any single coefficient,
modeling a fixed finite per-frequency coating strength. Conjugate symmetry
ensures that the allocation corresponds to a real-valued image. The box
constraint is needed only for the band-allocation result; parts (i)--(iv) do
not use it.

\paragraph{Theorem 1 (Learnability of the exact coating and its mean-shift component).}
Let $q_t$, $\widehat q_t$, and $q_t^{\mathrm{ms}}$ denote the noised clean,
exactly coated, and additive mean-shift-surrogate distributions, respectively.
Write $a_t=\bar\alpha_t$ and define
\begin{equation}
    v_t(f):=a_t\,P(f)+1-a_t,
    \qquad
    \widehat v_t(f):=a_t\gamma(f)^2P(f)+1-a_t,
    \label{eq:theory-clean-coated-variances}
\end{equation}
and
\begin{equation}
    \kappa_t(f)
    :=\log\frac{v_t(f)}{\widehat v_t(f)}
      +\frac{\widehat v_t(f)}{v_t(f)}-1\geq0.
    \label{eq:theory-variance-kl-term}
\end{equation}
Under Assumption~1 and the exact coating in
Eq.~\ref{eq:coated-spectrum}, the following statements hold.

\begin{enumerate}[label=(\roman*),leftmargin=*]
    \item \textbf{Discriminability after noising.} The exact divergence is
    \begin{equation}
        D_{\mathrm{KL}}\!\left(\widehat q_t\,\|\,q_t\right)
        =\frac{1}{2}\sum_f
        \left[\kappa_t(f)+b(f)w_t(f)\right]
        \geq
        \frac{1}{2}\sum_f b(f)w_t(f)
        =D_{\mathrm{KL}}\!\left(q_t^{\mathrm{ms}}\,\|\,q_t\right).
        \label{eq:theory-noised-kl}
    \end{equation}
    Thus, $B^2/2$ is the clean-data KL of the mean-shift surrogate and a
    lower bound for the exact coating, whose additional nonnegative term comes
    from variance contraction.

    \item \textbf{Survival window of the mean-shift component.} Define the
    last noise level at which the clean spectral power at frequency $f$
    remains above the noise floor as
    \begin{equation}
        t^\star(f)
        =\sup\{t:\operatorname{SNR}_t(f)\geq 1\}.
        \label{eq:theory-survival-time}
    \end{equation}
    Then $w_t(f)\geq 1/2$ for $t\leq t^\star(f)$ and
    $w_t(f)\leq\operatorname{SNR}_t(f)$ afterward. Because $P(f)$ decreases
    with frequency, $t^\star(f)$ also decreases with $|f|$. This describes the
    frequency dependence of the mean-shift terms in parts (i) and (iii).

    \item \textbf{Diffusion noise-prediction signal.} Let $\varepsilon_t^0$ and
    $\varepsilon_t^{\mathrm{coat}}$ be the Bayes-optimal noise predictors under
    the clean and exact coated distributions. Their squared difference on
    coated data is
    \begin{equation}
        R_t^{\mathrm{coat}}
        =\sum_f\left[
          b(f)w_t(f)\bigl(1-w_t(f)\bigr)+\zeta_t(f)
        \right]
        \geq R_t^{\mathrm{ms}}
        :=\sum_f b(f)w_t(f)\bigl(1-w_t(f)\bigr),
        \label{eq:theory-excess-risk}
    \end{equation}
    where
    \begin{equation}
        \zeta_t(f)
        :=\frac{(1-a_t)\bigl(v_t(f)-\widehat v_t(f)\bigr)^2}
        {v_t(f)^2\widehat v_t(f)}\geq0.
        \label{eq:theory-variance-risk-term}
    \end{equation}
    The mean-shift component $R_t^{\mathrm{ms}}$ is largest at
    $\operatorname{SNR}_t(f)=1$, where
    $w_t(f)(1-w_t(f))=1/4$.

    \item \textbf{Rectified-flow velocity signal.} For the linear interpolation
    $X_t^{\mathrm{RF}}(f)=(1-t)X_0(f)+tE(f)$, let
    $u_t^0$ and $u_t^{\mathrm{coat}}$ be the Bayes-optimal predictors of the
    canonical velocity target $U(f)=E(f)-X_0(f)$ under the clean and exact
    coated distributions. Set $A_t=1-t$ and
    \begin{align}
        v_t^{\mathrm{RF}}(f)
        &:=A_t^2P(f)+t^2,
        &
        \widehat v_t^{\mathrm{RF}}(f)
        &:=A_t^2\gamma(f)^2P(f)+t^2.
        \label{eq:theory-flow-variances}
    \end{align}
    Their squared difference on coated data is
    \begin{equation}
        R_{t,\mathrm{RF}}^{\mathrm{coat}}
        =\sum_f\left[
          b(f)h_t^{\mathrm{RF}}(f)+\zeta_t^{\mathrm{RF}}(f)
        \right]
        \geq R_{t,\mathrm{RF}}^{\mathrm{ms}}
        :=\sum_f b(f)h_t^{\mathrm{RF}}(f),
        \label{eq:theory-flow-excess-risk}
    \end{equation}
    where
    \begin{align}
        h_t^{\mathrm{RF}}(f)
        &:=\frac{t^2P(f)}{v_t^{\mathrm{RF}}(f)^2},
        &
        \zeta_t^{\mathrm{RF}}(f)
        &:=\frac{A_t^2t^2(1-\gamma(f)^2)^2P(f)^2}
        {v_t^{\mathrm{RF}}(f)^2\widehat v_t^{\mathrm{RF}}(f)}\geq0.
        \label{eq:theory-flow-risk-terms}
    \end{align}
    For each fixed $t\in(0,1)$, the mean-shift weight
    $h_t^{\mathrm{RF}}(f)$ is maximized when
    $A_t^2P(f)/t^2=1$. It is not obtained by merely substituting the
    rectified-flow interpolation coefficients into the diffusion
    noise-prediction risk.

    \item \textbf{Carrier bands for the objective-specific mean-shift lower bounds.}
    Let $\pi_\varepsilon(dt)$ and $\pi_{\mathrm{RF}}(dt)$ denote the effective
    training distributions over noise levels, including any deterministic
    timestep loss weighting, for diffusion noise prediction and linear flow
    matching, respectively. Define
    \begin{equation}
        G_\varepsilon(f)
        =\int w_t(f)\bigl(1-w_t(f)\bigr)\,\pi_\varepsilon(dt),
        \qquad
        G_{\mathrm{RF}}(f)
        =\int h_t^{\mathrm{RF}}(f)\,\pi_{\mathrm{RF}}(dt).
        \label{eq:theory-integrated-training-signal}
    \end{equation}
    For either objective $o\in\{\varepsilon,\mathrm{RF}\}$, let
    $J_o(b)=\sum_f b(f)G_o(f)$. Over
    $b\in\mathcal{B}(B,b_{\max})$, $J_o(b)$ is maximized by allocating the
    budget to frequencies with the largest values of the corresponding
    objective-specific weight $G_o(f)$. This optimizes only the mean-shift
    lower-bound component; the exact coated risks also contain their
    nonnegative variance-contraction terms. For diffusion noise prediction,
    if $\pi_\varepsilon$ has no point mass at the zero- or infinite-SNR
    endpoints, then $G_\varepsilon(f)$ approaches zero when $P(f)$ is either
    very small or very large. The integrated rectified-flow weight generally
    differs because it emphasizes timesteps differently.
\end{enumerate}

\paragraph{Proof.}
Using the variances in Eq.~\ref{eq:theory-clean-coated-variances}, write
\begin{align}
    \lambda_t
    &:=\frac{a_t}{1-a_t},
    &
    w_t(f)
    &=\frac{a_t\,P(f)}{v_t(f)}.
    \label{eq:theory-noised-variance}
\end{align}
Let $\mathcal{F}_+$ contain one representative from each non-self-conjugate frequency pair and let $\mathcal{F}_{\mathrm{sc}}$ contain the self-conjugate frequencies.

We first record one identity used for both training objectives. Let
$X_0\sim\mathcal{CN}(m,Q)$ and define an affine noising path and target
\begin{equation}
    Y=AX_0+BE,
    \qquad
    T=rE+sX_0,
    \qquad E\sim\mathcal{CN}(0,1).
    \label{eq:theory-affine-path}
\end{equation}
Writing $v_Q=A^2Q+B^2$ and
$k_Q=(rB+sAQ)/v_Q$, Gaussian conditioning gives
\begin{equation}
    \mathbb{E}[T\mid Y;m,Q]
    =sm+k_Q(Y-Am).
    \label{eq:theory-affine-predictor}
\end{equation}
Additionally, $\mathbb{E}[T]=sm$, $\mathbb{E}[Y]=Am$,
$\operatorname{Var}(Y)=v_Q$, and
$\operatorname{Cov}(T,Y)=rB+sAQ$. Comparing the clean parameters
$(m,Q)=(\mu,P)$ with the coated parameters
$(m,Q)=(\mu+\Delta,\gamma^2P)$ on coated data, let
$V:=Y-A(\mu+\Delta)$ and abbreviate
$k_0:=k_P$, $k_1:=k_{\gamma^2P}$. Then
\begin{align}
    \tau_0(Y)-\tau_1(Y)
    &=(Ak_0-s)\Delta+(k_0-k_1)V, \\
    \mathbb{E}_{\widehat q}
    |\tau_0(Y)-\tau_1(Y)|^2
    &=(Ak_0-s)^2|\Delta|^2
      +(k_0-k_1)^2v_{\gamma^2P}.
    \label{eq:theory-affine-risk}
\end{align}
The cross term vanishes because $V$ is zero mean. Thus, the first term is
the additive mean-shift contribution and the second is the nonnegative
correction caused by the exact coating's variance contraction.

\textit{Part (i).} At a non-self-conjugate frequency, the clean and coated
noised distributions are
\[
q_t=\mathcal{CN}(\sqrt{a_t}\mu,v_t),
\qquad
\widehat q_t=\mathcal{CN}(\sqrt{a_t}(\mu+\Delta),\widehat v_t).
\]
Using
\[
D_{\mathrm{KL}}\!\left(\mathcal{CN}(m_1,v_1)\|
\mathcal{CN}(m_0,v_0)\right)
=\log\frac{v_0}{v_1}+\frac{v_1+|m_1-m_0|^2}{v_0}-1
\]
gives the per-frequency contribution
$\kappa_t(f)+b(f)w_t(f)$. A self-conjugate real coefficient contributes half
as much, so the full-grid convention yields
Eq.~\ref{eq:theory-noised-kl}. Setting $\widehat v_t=v_t$ removes
$\kappa_t$ and recovers the additive mean-shift surrogate. At $t=0$,
its KL is $B^2/2$; for the default $\alpha=0.02$, the largest exact
variance correction is
$-\log(0.9604)+0.9604-1\approx8.05\times10^{-4}$ per
non-self-conjugate coefficient.

\textit{Part (ii).} Write
$S_t(f)=\operatorname{SNR}_t(f)=\lambda_t\,P(f)$. Since
$w_t=S_t/(1+S_t)$,
\[
w_t(f)\geq\tfrac12\iff S_t(f)\geq1,
\qquad
S_t(f)<1\implies w_t(f)\leq S_t(f).
\]
The monotonicity of $\lambda_t$ therefore makes
Eq.~\ref{eq:theory-survival-time} the unique SNR-one crossing. If
$|f_1|<|f_2|$, then $P(f_1)>P(f_2)$ by Assumption~1, so
$S_{t^\star(f_2)}(f_1)>S_{t^\star(f_2)}(f_2)=1$. Frequency $f_1$ must
therefore cross one later, proving $t^\star(f_1)>t^\star(f_2)$.

\textit{Part (iii).} Apply Eq.~\ref{eq:theory-affine-risk} with
\[
A=\sqrt{a_t},\quad B=\sqrt{1-a_t},\quad r=1,\quad s=0.
\]
Then $k_0=B/v_t$, $k_1=B/\widehat v_t$, and the mean-shift term is
\[
(Ak_0)^2|\Delta|^2
=\frac{a_t(1-a_t)|\Delta|^2}{v_t^2}
=b(f)w_t(f)(1-w_t(f)).
\]
The variance term is
$B^2(v_t-\widehat v_t)^2/(v_t^2\widehat v_t)=\zeta_t(f)$.
Summing over frequencies proves Eq.~\ref{eq:theory-excess-risk}. Finally,
$w(1-w)=1/4-(w-1/2)^2$, so the mean-shift contribution is maximized at
$w=1/2$, equivalently $\operatorname{SNR}_t(f)=1$.

\textit{Part (iv).} Apply Eq.~\ref{eq:theory-affine-risk} to linear flow
matching with
\[
A=1-t,\quad B=t,\quad r=1,\quad s=-1.
\]
The clean gain is $k_0=(t-AP)/v_t^{\mathrm{RF}}$, and
$Ak_0-s=1+Ak_0=t/v_t^{\mathrm{RF}}$. Moreover,
\[
k_0-k_1
=-\frac{At(1-\gamma^2)P}
{v_t^{\mathrm{RF}}\widehat v_t^{\mathrm{RF}}}.
\]
Substitution into Eq.~\ref{eq:theory-affine-risk} gives
Eq.~\ref{eq:theory-flow-excess-risk}. For fixed $t\in(0,1)$,
\[
\frac{\partial h_t^{\mathrm{RF}}}{\partial P}
=\frac{t^2(t^2-A^2P)}{(A^2P+t^2)^3},
\]
so the flow weight is maximized when $A^2P/t^2=1$. The corresponding
velocity-prediction gap is $t\Delta/v_t^{\mathrm{RF}}$, not the
$At\Delta/v_t^{\mathrm{RF}}$ gap obtained by treating the target as noise.

\textit{Part (v).} Exchanging the finite frequency sum with the training
integral gives, for either objective $o\in\{\varepsilon,\mathrm{RF}\}$,
\[
\int R_{t,o}^{\mathrm{ms}}\,\pi_o(dt)
=\sum_f b(f)G_o(f)=J_o(b),
\]
where $R_{t,\varepsilon}^{\mathrm{ms}}:=R_t^{\mathrm{ms}}$. If
$G_o(f_1)>G_o(f_2)$ while $b(f_1)<b_{\max}$ and $b(f_2)>0$, transferring
\[
\delta=\min\{b_{\max}-b(f_1),b(f_2)\}
\]
from $f_2$ to $f_1$ preserves feasibility and increases the objective by
$\delta(G_o(f_1)-G_o(f_2))>0$. Hence every optimum fills frequencies in
descending order of the relevant objective-specific weight. Finally,
\[
G_\varepsilon(P)
=\int\frac{\lambda_t\,P}{(1+\lambda_t\,P)^2}\,\pi_\varepsilon(dt).
\]
The integrand is bounded by $1/4$ and tends to zero as $P\to0$ or
$P\to\infty$ at every interior noise level. Dominated convergence proves the
stated diffusion limits. \hfill\qedsymbol

\paragraph{Connection to the Gaussian-scale ablation.}
Theorem 1 gives separate objective-specific carrier weights. For SD1.5-style
noise prediction, the mean-shift weight is $G_\varepsilon(f)$; for the
SD3.5-style linear flow-matching target, it is $G_{\mathrm{RF}}(f)$. At each
fixed interior noise level, both weights are maximized when the corresponding
frequency has signal-to-noise ratio one, although their integration over
training timesteps differs. This supports an intermediate-band mechanism but
does not imply identical optimal mask scales for the two objectives. In
Fig.~\ref{fig:spectralcanary-ablation}, $c=0.20$ is a near-saturating
efficiency--bandwidth choice, while $c=1.00$ gives marginally higher
traceability by modifying a broader frequency range; $c=0.05$ is too narrow
to provide the same trace strength. This comparison remains qualitative
because the coating is applied in pixel space whereas SD1.5 and SD3.5 train
in learned latent spaces, and the ablation changes total trace energy rather
than holding the normalized budget fixed. The three tested scales therefore
do not constitute empirical evidence for a non-monotonic trend or an interior
optimum under the theorem's fixed-budget constraint.

\section{Extended Related Work} \label{appx:related_work}

\begin{table*}[ht]
\centering
\caption{
\textbf{Comparison of T2I data-poisoning and proactive dataset-use auditing methods.}
We consider methods that modify only the released training data and assume a standard training or fine-tuning pipeline.
\emph{Content-preserving} indicates that the modification does not intentionally alter the image's original semantic content.
}
\label{tab:poisoning_traceability_comparison}
\small
\setlength{\tabcolsep}{2.5pt}
\renewcommand{\arraystretch}{1.10}
\begin{tabular}{@{}
>{\raggedright\arraybackslash}p{2.7cm}
>{\raggedright\arraybackslash}p{2cm}
>{\raggedright\arraybackslash}p{2.5cm}
>{\raggedright\arraybackslash}p{3.55cm}
>{\raggedright\arraybackslash}p{2.5cm}@{}}
\toprule
Method
& Venue
& Goal
& Image modification
& Caption modification \\
\midrule

Nightshade
& IEEE S\&P 2024
& Disrupt T2I concepts
& Content-preserving adversarial perturbation
& None \\
\addlinespace[2pt]

SilentBadDiffusion
& ICML 2024
& Induce copyright reproduction
& Semantic elements inserted into synthetic images
& Element-specific phrases \\
\addlinespace[2pt]

Bilateral Data Poisoning
& NeurIPS 2024
& Corrupt T2I generation
& Adversarial image modification
& Poisoned captions \\
\addlinespace[2pt]

Silent Branding
& CVPR 2025
& Inject persistent visual branding
& Semantic elements inserted into protected samples
& None \\

\addlinespace[2pt]
\midrule

DIAGNOSIS
& ICLR 2024
& Dataset-use auditing
& Content-preserving image warping
& Rare trigger token added \\
\addlinespace[2pt]

SIREN
& IEEE S\&P 2025
& Dataset-use auditing
& Content-preserving perturbation
& None \\
\addlinespace[2pt]

EnTruth
& NeurIPS 2025 GenProCC
& Dataset-use auditing
& Semantically distinct samples injected
& Trigger captions introduced \\
\addlinespace[2pt]

\textbf{SpectralCanary}
& --
& Dataset-use auditing
& \textbf{Content-preserving spectral hybridization}
& \textbf{None} \\

\bottomrule
\end{tabular}
\end{table*}

% \textbf{Reactive verification.} Reactive methods leave the released data unchanged and later test whether a model was trained on it. Membership inference estimates whether a particular image or prompt--image pair belonged to the training set; recent work studies this problem for fine-tuned diffusion models in black-box settings~\citep{chen2026membership,pang2025black,FSCA,NUDUI}. Other methods compare generated images with protected examples using representations such as CLIP~\citep{radford2021learning} or DINO~\citep{oquab2024dinov}. These methods apply to collections that have already been published, but their evidence can weaken when a T2I model generalizes without retaining identifiable information about individual samples~\citep{11023473,ren2025entruth,liang2024real}.

\textbf{Reactive auditing.} Reactive methods leave the released data unchanged and later test whether a model was trained on it. Membership inference estimates whether a particular image or prompt--image pair belonged to the training set; recent work studies this problem for fine-tuned diffusion models in black-box settings~\citep{chen2026membership,pang2025black,FSCA,NUDUI}. RECOVER~\citep{wei2026recover} instead identifies individual fine-tuning images through their increased robustness under stochastic inversion--reconstruction, without modifying the images or requiring the pre-fine-tuned model. Other methods compare generated images with protected examples using representations such as CLIP~\citep{radford2021learning} or DINO~\citep{oquab2024dinov}. These methods apply to collections that have already been published, but their evidence can weaken when a T2I model generalizes without retaining identifiable information about individual samples~\citep{11023473,ren2025entruth,liang2024real}.

\textbf{Proactive protection and auditing.} Anti-learning methods, including Glaze~\cite{291164}, AdvDM~\cite{zhao2024unlearnable}, Anti-DreamBooth~\cite{van2023anti}, FastProtect~\cite{ahn2025nearly}, and MetaCloak~\cite{Liu_2024_CVPR}, modify images to prevent or redirect unauthorized customization. Proactive dataset-use auditing instead inserts a signal intended to transfer into the customized model. Its goal is not to make fine-tuning fail, but to make outputs from a model trained on the protected collection statistically distinguishable from outputs of clean-data reference models.

\textbf{Pixel-level traces.} DIAGNOSIS~\citep{wang2024diagnosis}, DiffusionShield~\citep{10.1145/3715073.3715079}, FT-Shield~\citep{10.1145/3715073.3715080}, and SIREN~\citep{11023473} insert watermarks, transformations, or trigger-like signals that can later be detected in generated images. Their protected images remain close to the originals, but weak nonsemantic signals may be attenuated by image editing, regeneration, augmentation, or a change in T2I backbone. SpectralCanary shares the goal of preserving image appearance but uses a common low-frequency visual component rather than an optimized or pixel-local signal.

\textbf{Explicit visual canaries.} EnTruth~\citep{ren2025entruth} inserts samples with a repeated visual template and a textual trigger. The coherent template transfers more readily than a weak pixel-level pattern, but the samples are visibly out of distribution (Appx.~\ref{appx:entruth-filtering}) and can be removed by visual filtering or caption cleaning. SpectralCanary instead distributes a weak version of the same visual component across the original collection and does not alter captions.

\textbf{Relation to T2I data poisoning.} Dataset-use auditing and data poisoning both modify training data so that the resulting model acquires an additional behavior~\citep{shafahi2018poison,geiping2021witches,xu2024shadowcast,ho2025memory}. BBadT2I~\citep{zhai2023text} implants trigger-dependent pixel-, object-, or style-level behaviors. Nightshade~\citep{shan2024nightshade} corrupts targeted concepts through modified image--text pairs, Bilateral Data Poisoning~\citep{pan2024from} induces persistent backdoor behavior, and SilentBadDiffusion~\citep{pmlr-v235-wang24bm} elicits attacker-specified content under trigger prompts. Silent Branding~\citep{jang2025silent} is especially relevant because it modifies only images and induces reproduction of a repeated visual logo without a textual trigger. We adapt it as a dataset-use auditing baseline under the same downstream fine-tuning and auditing protocol (Appx.~\ref{appx:silent-branding}).

\end{document}